\documentclass[twoside,11pt]{article}
\usepackage{jair}
\usepackage{theapa}
\usepackage{rawfonts}

\usepackage{url}
\usepackage{latexsym}
\usepackage{amssymb}
\usepackage{amsmath}
\usepackage{graphicx}
\usepackage{xspace}
\usepackage{tikz-qtree}
\usepackage[all]{xy}
\usepackage[utf8x]{inputenc}
\usepackage[leftmargin=2mm,rightmargin=2mm]{quoting}
\usepackage[margin=1in]{geometry}

\usepackage{tikz}

\usetikzlibrary{intersections,positioning}

\newcommand{\pmi}{\textsc{pmi}\xspace}
\newcommand{\avgpmi}{\textsc{AvgTokPmi}\xspace}

\newcommand{\nlg}{\textsc{nlg}\xspace}
\newcommand{\np}{\textsc{np}\xspace}
\newcommand{\nps}{\textsc{np}s\xspace}
\newcommand{\nl}{\textsc{nl}\xspace}

\newcommand{\nlowl}{Natural\textsc{owl}\xspace}
\newcommand{\owl}{\textsc{owl}\xspace}
\newcommand{\owltwo}{\textsc{owl2}\xspace}
\newcommand{\rdf}{\textsc{rdf}\xspace}
\newcommand{\pos}{\textsc{pos}\xspace}
\newcommand{\nonl}{\textsc{no-nln}\xspace}
\newcommand{\authnl}{\textsc{manual-nln}\xspace}
\newcommand{\autonl}{\textsc{auto-nln}\xspace}
\newcommand{\semiautonl}{\textsc{semi-auto-nln}\xspace}

\newcommand{\spg}{\textsc{sp}\xspace}
\newcommand{\spgrerank}{\textsc{sp*}\xspace}

\newcommand{\verbalizer}{\textsc{verbalizer}\xspace}
\newcommand{\jointauto}{\textsc{auto}\xspace}
\newcommand{\jointsemiauto}{\textsc{semi-auto}\xspace}
\newcommand{\manual}{\textsc{manual}\xspace}
\newcommand{\nosp}{\textsc{no-sp}\xspace}
\newcommand{\judgetwo}{$J_2$\xspace}
\newcommand{\judgeone}{$J_1$\xspace}
\newcommand{\authsp}{\textsc{manual-sp}\xspace}

\newcommand{\autospgrerank}{\textsc{auto-sp*}\xspace}
\newcommand{\semiautospgrerank}{\textsc{semi-auto-sp*}\xspace}
\newcommand{\autobootstrap}{\textsc{auto-boot}\xspace}
\newcommand{\semiautobootstrap}{\textsc{semi-auto-boot}\xspace}
\newcommand{\bootstrap}{\textsc{boot}\xspace}

\newcommand{\etalt}{et al.\xspace}
\newcommand{\ontosum}{\textsc{ontosum}\xspace}

\newcommand{\protege}{Prot\'eg\'e\xspace}
\newcommand{\rdfs}{\textsc{rdf schema}\xspace}
\newcommand{\swat}{\textsc{swat}\xspace}

\newcommand{\code}[1]{{\small \texttt{#1}}}
\newcommand{\mpiro}{\textsc{m-piro}\xspace}
\newcommand{\itab}[1]{\hspace{0em}\rlap{#1}}
\newcommand{\tab}[1]{\hspace{.3\textwidth}\rlap{#1}}

\newcommand{\mystrut}{\rule[-.3\baselineskip]{0pt}{\baselineskip}}

\ShortHeadings{Extracting Linguistic Resources from the Web for Concept-to-Text Generation}
{Lampouras, \& Androutsopoulos}

\begin{document}

\title{Extracting Linguistic Resources from the Web for Concept-to-Text Generation}

\author{\name Gerasimos Lampouras \email lampouras06@aueb.gr \\
       \addr Department of Informatics,  Athens University of Economics and Business, Greece \\
       \AND
       \name Ion Androutsopoulos \email ion@aueb.gr \\
       \addr Department of Informatics, Athens University of Economics and Business, Greece 
}

\maketitle

\begin{abstract}
Many concept-to-text generation systems require domain-specific linguistic resources to produce high quality texts, but manually constructing these resources can be tedious and costly. Focusing on 
\nlowl, a publicly available state of the art natural language generator for \owl ontologies, we
propose methods to 
extract from the Web sentence plans and natural language names, 
two of the most important types of domain-specific linguistic resources used by the generator.
Experiments show that texts generated using 
linguistic resources extracted by our methods in a semi-automatic manner, 
with minimal human involvement, are perceived as being 
almost as good as texts generated using 
manually authored linguistic resources, and much better than texts produced by using linguistic resources extracted from the relation and entity identifiers of the ontology.
\end{abstract}

\section{Introduction} \label{Introduction}

The Semantic Web \cite{BernersLee2001,Shadbolt2006} and the growing popularity of Linked Data (data 
published using Semantic Web technologies) have renewed interest in concept-to-text generation \cite{ReiterDale2000}, especially text generation from ontologies \cite{Bontcheva2005,Mellish2006,Galanis2007,Mellish2008,Schwitter2008,Schwitter2010,Liang2011b,Williams2011,Androutsopoulos2013}. An ontology provides a conceptualization of a knowledge domain 
(e.g., wines, consumer electronics) by defining the classes and subclasses of the individuals (entities) 
in the domain, the possible relations between them etc. The current standard to specify Semantic Web ontologies is \owl \cite{Horrocks2003}, which is based on description logics \cite{Baader2002}, \rdf, and \rdfs \cite{Antoniou2008}.
\owltwo is the latest version of \owl \cite{Grau2008}.\footnote{Most Linked Data currently use only \rdf and \rdfs, but \owl is in effect a superset of \rdfs and, hence, methods to produce texts from \owl also apply to Linked Data. Consult also \url{http://linkeddata.org/}.}  Given an \owl ontology for a knowledge domain, one can publish on the Web machine-readable data pertaining to that domain (e.g., catalogues of products, their features etc.), with the data having formally defined
semantics based on the ontology. 

Several equivalent \owl syntaxes have been developed, but people unfamiliar with formal knowledge representation 
have difficulties understanding them \cite{Rector2004}. For example, the following statement defines the class of St.\ Emilion wines, using the functional-style syntax of \owl, one of the easiest to understand.\footnote{Consult \texttt{http://www.w3.org/TR/owl2-primer/} for an introduction to the functional-style syntax of \owl.}

\label{stemilion}
{\footnotesize
\begin{verbatim}
   SubClassOf(:StEmilion 
     ObjectIntersectionOf(:Bordeaux                     
       ObjectHasValue(:locatedIn :stEmilionRegion) 
       ObjectHasValue(:hasColor :red)
       ObjectHasValue(:hasFlavor :strong)          
       ObjectHasValue(:madeFrom :cabernetSauvignonGrape) 
       ObjectMaxCardinality(1 :madeFrom)))
\end{verbatim}
}

To make ontologies easier to understand, several \emph{ontology verbalizers} have been developed \cite{Cregan2007,Kaljurand2007,Schwitter2008,HalaschekWiener2008,Schutte2009,Power2010b,Power2010,Schwitter2010,Stevens2011,Liang2011b}. Although verbalizers can be viewed as performing a kind of light natural language generation (\nlg), they usually translate the axioms (in our case, \owl statements) of the ontology one by one to controlled, often not entirely fluent English statements, typically without considering the coherence of the resulting texts. By contrast, more elaborate \nlg systems \cite{Bontcheva2005,Androutsopoulos2007,Androutsopoulos2013} can produce more fluent and coherent multi-sentence texts, but they need domain-specific linguistic resources. For example, \nlowl \cite{Androutsopoulos2013}, a publicly available \nlg system for \owl ontologies, produces the following description of St.\ Emilion wines from the \owl statement above. It needs, however: a \emph{sentence plan} for each relation (e.g., \code{:locatedIn}) of the ontology, i.e., a linguistically annotated template showing how to express the relation; a \emph{natural language name} for each class and individual, i.e., a linguistically annotated noun phrase to be used as the name of the class or individual; 
a text plan specifying the order in which relations should be expressed etc. Similar domain-specific linguistic resources are used in most concept-to-text systems \cite{ReiterDale2000}. Manually constructing resources of this kind, however, can be tedious and costly. 

\begin{quoting}
{\small 
\noindent St.\ Emilion is a kind of red, strong Bordeaux from the St.\ Emilion region. It is made from exactly one grape variety: Cabernet Sauvignon grapes.
}
\end{quoting}

Instead of requiring domain-specific linguistic resources, simpler verbalizers use the \owl identifiers of classes and individuals (e.g., \code{:cabernetSauvignonGrape}) typically split into tokens as their natural language names, they express relations using phrases obtained by tokenizing the \owl identifiers of the relations (e.g., 
\code{:hasColor}), 
they order the resulting sentences following the ordering of the corresponding \owl statements etc. Without domain-specific linguistic resources, \nlowl behaves like a simple verbalizer, producing the following lower quality text from the \owl statement above. A further limitation of using tokenized \owl identifiers is that non-English texts cannot be generated, since \owl identifiers are usually English-like. 

\begin{quoting}
{\small
\noindent St Emilion is Bordeaux. St Emilion located in St Emilion Region. St Emilion has color Red. St Emilion has flavor Strong. St Emilion made from grape exactly 1: Cabernet Sauvignon Grape. 
}
\end{quoting}

\noindent Previous experiments \cite{Androutsopoulos2013} indicate that the texts that \nlowl generates with domain-specific linguistic resources are perceived as significantly better than 
(i) those of \swat, one of the 
best available \owl verbalizers \cite{Stevens2011,Williams2011}, and 
(ii) those of \nlowl without domain-specific linguistic resources, with little or no difference between 
(i) and (ii). 
The largest difference in the perceived quality of the texts was reported to be due to the sentence plans, natural language names, and
(to a lesser extent) text plans. 

In this paper, we present 
methods to automatically or semi-automatically extract from the Web the natural language names and sentence plans required by \nlowl for a given ontology. 
We do not examine how other types of domain-specific linguistic resources (e.g., text plans) 
can be 
generated, leaving them for future work. We base our work on \nlowl, because it 
appears to be the only open-source \nlg system for \owl that implements all the processing stages of a typical \nlg pipeline \cite{ReiterDale2000}, it supports \owltwo, it is extensively documented, and has been tested with several ontologies.
The processing stages and linguistic resources of 
\nlowl, however, are typical of \nlg systems \cite{Mellish2006c}. Hence, we believe that our work is also applicable, at least in principle, to other \nlg systems. Our methods may also be useful in simpler verbalizers,  where the main concern seems to be to avoid manually authoring domain-specific linguistic resources. Experiments show that texts generated 
using linguistic resources extracted by our methods
with minimal human involvement are perceived as being almost as good as texts generated using manually authored linguistic resources, and much better than texts produced by using tokenized \owl identifiers. 

Section~\ref{NaturalOWL} below provides background information about \nlowl, especially its natural language names and sentence plans. Sections~\ref{OurMethodNLN} and \ref{OurMethodSP} then describe our methods to extract natural language names and sentence plans, respectively, from the Web. Section~\ref{Experiments} presents our experimental results. Section~\ref{RelatedWork} discusses related work. Section~\ref{Conclusion} concludes and suggests future work.

\section{Background information about NaturalOWL} \label{NaturalOWL}

Given an \owl ontology and a particular target individual or class to describe, \nlowl first scans the 
ontology to select statements
relevant to the target. It then converts each relevant 
statement into (possibly multiple) \emph{message triples} of the form $\left<S, R, O\right>$, where $S$ is an individual or class, $O$ is another individual, class, or datatype value, and $R$ is a relation (property) that connects $S$ to $O$. For example, the \code{ObjectHasValue(:madeFrom :cabernetSauvignonGrape)} part of the \owl statement above is converted to the message triple $<$\code{:StEmilion}, \code{:madeFrom}, \code{:cabernetSauvignonGrape}$>$. Message triples are similar to \rdf triples, but they are easier to express as 
sentences. Unlike \rdf triples, the relations ($R$) of the message triples may include \emph{relation modifiers}. For example, the \code{ObjectMaxCardinality(1 :madeFrom)} part of the \owl statement 
above is turned into the message triple $<$\code{:StEmilion}, \code{maxCardinality(:madeFrom)}, 1$>$, where \code{maxCardinality} is a relation modifier. In this paper, we  
consider only sentence plans for message triples without relation modifiers, because \nlowl already automatically constructs sentence plans for triples with  
relation modifiers from sentence plans for triples without them.

Having produced the message triples, \nlowl consults a user model to select the most interesting ones that have not been expressed already,
and orders 
the selected triples according to manually authored text plans. Later processing stages convert each message triple to an abstract  sentence representation, aggregate sentences to produce longer ones, and produce appropriate referring expressions (e.g., pronouns). The latter three stages require a sentence plan for each relation ($R$),
while the last stage also requires natural language names for each individual or class
($S$ or $O$). 

\subsection{The natural language names of NaturalOWL} \label{NLnamesBackground}

In \nlowl, a natural language (\nl) name is a sequence of slots. The contents of the slots are concatenated to produce a noun phrase to be used as the name of a class or individual. Each slot is accompanied by annotations specifying how to fill it in; the annotations may also provide linguistic information about the contents of the slot. For example, we may specify that the English \nl name of the class \code{:TraditionalWinePiemonte} is the following.\footnote{The \nl names and sentence plans of \nlowl are actually represented in \owl, as instances of an ontology that describes the domain-dependent linguistic resources of the system.}

\begin{center}
[\,]$^{1}_{\textit{article}, \, \textit{indef}, \, \textit{agr}=3}$ 
[traditional]$^{2}_{\textit{adj}}$ 
[wine]$^{3}_{\textit{headnoun}, \, \textit{sing}, \, \textit{neut}}$
[from]$^{4}_{\textit{prep}}$ 
[\,]$^{5}_{\textit{article}, \, \textit{def}}$ 
[Piemonte]$^{6}_{\textit{noun}, \, \textit{sing}, \, \textit{neut}}$ 
[region]$^{7}_{\textit{noun}, \, \textit{sing}, \, \textit{neut}}$ 
\end{center}

\noindent The first slot is to be filled in with an indefinite article, whose number should agree with the third slot.  The second slot is to be filled in with the adjective `traditional'. The third slot with the neuter noun `wine', which will also be the head (central) noun of the noun phrase, in singular number, and similarly for the other slots. \nlowl makes no distinctions between common and proper nouns, but it can be instructed to capitalize particular nouns (e.g., `Piemonte'). In the case of the message triple  $<$\code{:wine32}, \code{instanceOf}, \code{:TraditionalWinePiemonte}$>$, the \nl name above would allow a sentence like ``This is \emph{a traditional wine from the Piemonte region}'' to be produced. 

The 
slot annotations allow \nlowl to automatically adjust the \nl names.
For example, the system also generates comparisons to previously encountered individuals or classes, as in ``Unlike the previous products that you have seen, which were all \emph{traditional wines from the Piemonte region}, this is a French wine''. In this particular example, the head noun (`wine') had to be turned into plural. Due to number agreement, its article also had to be turned into plural; in English, the plural indefinite article is void, hence the article of the head noun was omitted. 

As a further example, we may specify that the \nl name of the class \code{FamousWine} is the following.

\begin{center}
[\,]$^{1}_{\textit{article}, \, \textit{indef}, \, \textit{agr}=3}$ 
[famous]$^{2}_{\textit{adj}}$ 
[wine]$^{3}_{\textit{headnoun}, \, \textit{sing}, \, \textit{neut}}$
\end{center}

\noindent 
If $<$\code{:wine32}, \code{instanceOf}, \code{:TraditionalWinePiemonte}$>$ and $<$\code{:wine32}, \code{instanceOf}, \code{:FamousWine}$>$ were 
both to be expressed, \nlowl would then produce the single, aggregated sentence ``This is a famous traditional wine from the Piemonte region'', instead of two separate sentences ``This is a traditional wine from the Piemonte region'' and ``This is a famous wine''. The annotations of the slots, which indicate for example which words are adjectives and head nouns, are used by the sentence aggregation component of \nlowl to appropriately combine the two sentences. The referring expression generation component also uses the  slot annotations to identify the gender of the head noun,
when a pronoun has to be generated (e.g., ``it'' when the head noun is neuter). 

We can now define more precisely 
\nl names. A \nl name is a sequence of one or more slots. Each slot is accompanied by annotations requiring it to be filled in with 
exactly one of the following:\footnote{\nlowl also supports Greek. The possible annotations for Greek \nl names (and sentence plans, see below) are 
slightly different, but in this paper we consider only English \nl names (and sentence plans).}

\smallskip
(i) \textit{An article}, definite or indefinite, 
possibly to agree with another slot filled in by a noun. 

(ii) \textit{A noun flagged as the head}.
The number of the head noun must also be specified. 

(iii) \textit{An adjective flagged as the head}.
For example, the \nl name name of the individual
\code{:red} may consist of a single slot, to be filled in with the adjective `red'; in this case, the adjective is the head of the \nl name. The number and gender of the head adjective must also be specified. 

(iv) \textit{Any other noun or adjective}, (v) \textit{a preposition}, or (vi) \textit{any fixed string}. 

\smallskip
\noindent Exactly one head (noun or adjective) must be specified per \nl name.  For nouns and adjectives, the \nl name may require a particular inflectional form to be used (e.g., in a particular number, case, or gender), or it may require an inflectional form that agrees with another noun or adjective slot.\footnote{We use \textsc{simplenlg} \cite{Gatt2009} to generate the inflectional forms of nouns, adjectives, and 
verbs.}

When providing \nl names, 
an individual or class can also be declared to be \emph{anonymous}, indicating that \nlowl should avoid referring to it by name. For example, in 
a museum ontology, there may be a 
coin whose \owl identifier is \code{:exhibit49}. We may not wish to provide an \nl name for this individual (it may not have an English name); and we may 
want \nlowl to avoid referring to the coin 
by tokenizing its identifier (``exhibit 49''). By declaring the coin as anonymous, \nlowl would use only the \nl name of its class (e.g., ``this coin''), simply ``this'', or a pronoun. 

\subsection{The sentence plans of NaturalOWL} \label{sentencePlansBackground}

In \nlowl, a sentence plan for a relation $R$ specifies how to construct a sentence to express any message triple of the form $\left<S, R, O\right>$. Like \nl names, sentence plans are sequences of slots with annotations specifying how to fill the slots in. The contents of the slots are concatenated to produce the sentence. For example, the following is a sentence plan for the relation \code{:madeFrom}.
\begin{center}
[$\mathit{ref}(S)$]$^{1}_{\textit{nom}}$
[make]$^{2}_{\textit{verb}, \, \textit{passive}, \, \textit{present}, \, \textit{agr}=1, \, \textit{polarity}=+}$ 
[from]$^{3}_{prep}$ 
[$\mathit{ref}(O)$]$^{4}_{\textit{acc}}$
\end{center}
Given the message triple $<$\code{:StEmilion}, \code{:madeFrom}, \code{:cabernetSauvignonGrape}$>$, the sentence plan would lead to sentences like ``St.\ Emilion is made from Cabernet Sauvignon grapes'', or ``It is made from  Cabernet Sauvignon grapes'', assuming that appropriate \nl names have been provided for \code{:StEmilion} and \code{:cabernetSauvignonGrape}. Similarly, given  $<$\code{:Wine}, \code{:madeFrom}, \code{:Grape}$>$, the sentence plan above would lead to sentences like ``Wines are made from grapes'' or ``They are made from grapes'', assuming again appropriate \nl names. As another example, the following sentence plan can be used with the relations \code{:hasColor} and \code{:hasFlavor}. 
\begin{center}
[$\mathit{ref}(S)$]$^{1}_{\textit{nom}}$
[be]$^{2}_{\textit{verb}, \, \textit{active}, \, \textit{present}, \, \textit{agr}=1, \, \textit{polarity}=+}$
[$\mathit{ref}(O)$]$^{3}_{\textit{nom}}$
\end{center}
Given the message triples $<$\code{:StEmilion}, \code{:hasColor}, \code{:red}$>$ and $<$\code{:StEmilion}, \code{:hasFlavor}, \code{:strong}$>$, it would produce the sentences ``St.\ Emilion is red'' and ``St.\ Emilion is strong'', respectively. 
 
The first sentence plan above, for \code{:madeFrom}, has four slots. The first 
slot is to be filled in with an automatically generated referring expression (e.g., pronoun or name) for $S$, in nominative case. The verb of the second slot is to be realized in passive voice, present tense, and positive polarity (as opposed to expressing negation) and should agree (in number and person) with the referring expression of the first slot ($\textit{agr}=1$). The third slot is filled in with the preposition `from', and the fourth slot with an automatically generated referring expression for $O$, in accusative case. 

\nlowl has built-in sentence plans for domain-independent relations (e.g., \code{isA}, \code{instanceOf}).
For example, 
$<$\code{:StEmilion}, \code{isA}, \code{:Bordeaux}$>$ is expressed as ``St.\ Emilion is a kind of Bordeaux'' using the following built-in sentence 
plan; 
the last slot 
requires the \nl name of $O$ without article.
\begin{center}
[$\mathit{ref}(S)$]$^1_{\textit{nom}}$
[be]$^2_{\textit{verb}, \, \textit{active}, \, \textit{present}, \, \textit{agr}=1, \, \textit{polarity}=+}$
[``a kind of'']$^3_{\textit{string}}$
[$\mathit{name}(O)$]$^4_{\textit{noarticle}, \textit{nom}}$
\end{center}

Notice that the sentence plans of \nlowl are not simply slotted string templates (e.g., ``$X$ is made from $Y$''). 
Their linguistic annotations (e.g., \pos tags, agreement,
voice, tense, cases) along with the 
annotations of the \nl names allow \nlowl to produce more natural sentences (e.g., turn the verb into plural when the subject is also plural), produce appropriate referring expressions (e.g., pronouns in the correct 
cases and genders), and aggregate shorter sentences into longer ones. For example, the linguistic annotations of the \nl names and sentence plans allow \nlowl to produce the aggregated sentence ``St.\ Emilion is a kind of red Bordeaux made from Cabernet Sauvignon grapes'' from the  triples $<$\code{:StEmilion}, \code{isA}, \code{:Bordeaux}$>$, $<$\code{:StEmilion}, \code{:hasColor}, \code{:red}$>$, $<$\code{:StEmilion}, \code{:madeFrom}, \code{:cabernetSauvignonGrape}$>$, instead of three separate sentences. 

We can now define more precisely 
sentence plans. A sentence plan is a sequence of
slots. Each slot is accompanied by annotations requiring it to be filled in with 
exactly one of the following:

\smallskip
(i) \textit{A referring expression for the $S$} 
(a.k.a.\  the \emph{owner}) of the message triple in a particular case.

(ii) \textit{A verb} in a particular polarity and 
form (e.g., tense), possibly to agree with another slot. 

(iii) \textit{A noun or adjective} in a particular form,
possibly to agree with another slot.

(iv) \textit{A preposition}, or (v) \textit{a fixed string}. 

(vi) \textit{A referring expression for the $O$}
(a.k.a.\ the \emph{filler}) of the message triple. 
\smallskip

\begin{figure}
\center
\includegraphics[width=\columnwidth]{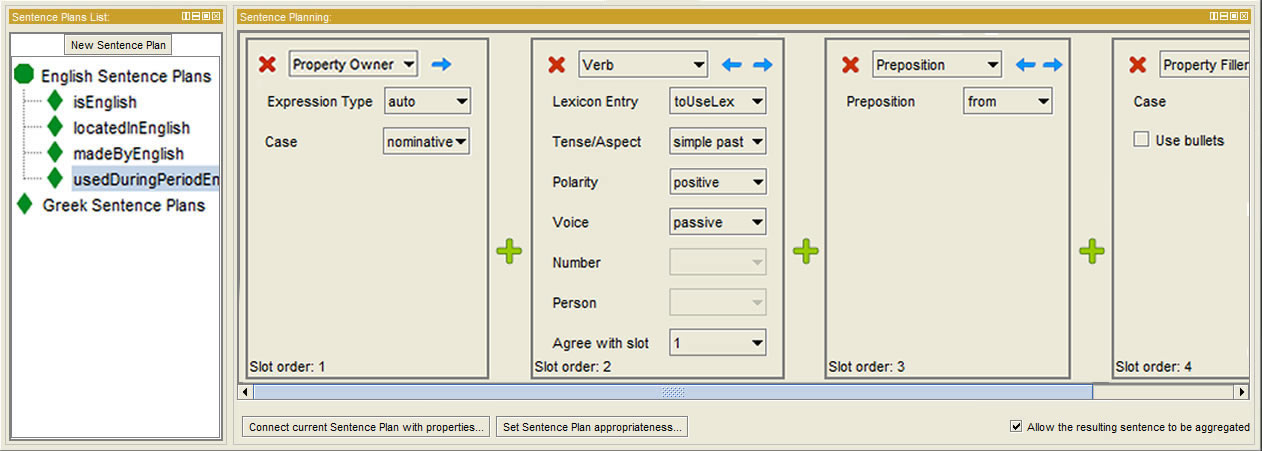}
\caption{A manually authored sentence plan in the \protege plug-in of \nlowl.}
\label{sentencePlanScreenshot}
\end{figure}

More details about the \nl names and sentence plans of \nlowl and their roles in sentence aggregation, referring expressions etc.\ can be found elsewhere \cite{Androutsopoulos2013}. Both sentence plans and \nl names were so far authored manually, using a \protege plug-in 
(Fig.~\ref{sentencePlanScreenshot}).\footnote{Consult \url{http://protege.stanford.edu/}
and \url{http://nlp.cs.aueb.gr/software.html}.}

\section{Our method to 
extract 
natural language names from the Web} \label{OurMethodNLN}

Given a target class or individual $t$ that we want to produce an \nl name for, 
we first extract from the Web noun phrases that are similar to the \owl identifier of $t$. 
The noun phrases are  ranked by aligning their words to the tokens of the 
identifier. The 
top-ranked noun phrases are then enhanced with linguistic annotations (e.g., \pos tags, agreement,
number), missing articles etc., turning them into \nl names. 
We aim to identify the best few (up to 5) candidate \nl names for $t$. In a fully automatic scenario, the candidate \nl name that the method considers best for $t$ is then used. In a semi-automatic scenario, the few top (according to the method) \nl names of $t$ are shown to a human author, who picks the best one; this is much easier than manually authoring \nl names.

\subsection{Extracting noun phrases from the Web} \label{NounPhraseExtraction}

We first
collect the \owl statements of the ontology that describe $t$, the individual or class  we want to produce an \nl name for, and turn them into message triples $\left<S=t, R, O\right>$, as when generating texts. For example, for the class
$t=$ \code{:KalinCellarsSemillon} of the Wine Ontology, one of the ontologies of our experiments, three of the resulting message triples are:

\begin{center}
$<$\code{:KalinCellarsSemillon}, \code{isA}, \code{:Semillon}$>$\\
$<$\code{:KalinCellarsSemillon}, \code{:hasMaker}, \code{:KalinCellars}$>$\\
$<$\code{:KalinCellarsSemillon}, \code{:hasFlavor}, \code{:Strong}$>$
\end{center}

For each 
collected message triple 
$\left<S=t, R, O\right>$,
we then produce $\textit{tokName}(S)$ and $\textit{tokName}(O)$, where $\textit{tokName}(X)$ is the tokenized identifier of $X$.\footnote{Most \owl ontologies use identifiers written in CamelCase, or identifiers that can be easily broken into tokens at underscores, hyphens etc. If the ontology provides an \code{rdfs:label} for $X$, we use
its tokens as $\textit{tokName}(X)$.} From the three triples above, we obtain:  

\begin{center}
{\small \textit{tokName}(\code{:KalinCellarsSemillon}) $=$ ``Kalin Cellars Semillon''} , 
{\small \textit{tokName}(\code{:Semillon}) $=$ ``Semillon''} \\
{\small \textit{tokName}(\code{:KalinCellars}) $=$ ``Kalin Cellars''}, 
{\small \textit{tokName}(\code{:Strong}) $=$ ``Strong''} \\
\end{center}

\subsubsection{Shortening the tokenized identifiers} \label{shorteningIdentifiers}

Subsequently, we attempt to shorten $\textit{tokName}(t)$, i.e., the tokenized identifier of the individual or class we wish to produce an \nl name for, by removing any part (token sequence) of $\textit{tokName}(t)$ that is identical to the tokenized identifier of the $O$ of any triple $\left<S=t, R, O\right>$ that we 
collected for $t$. If the shortened tokenized identifier of $t$ is the empty string or contains only numbers, $t$ is marked as anonymous (Section~\ref{NLnamesBackground}). In our example, where $t =$ \code{:KalinCellarsSemillon}, the tokenized identifier of $t$ is initially $\textit{tokName}(t)=$ ``Kalin Cellars Semillon''.  We remove the part ``Semillon'', because of the triple $<$\code{:KalinCellarsSemillon}, \code{:isA}, \code{:Semillon}$>$ and the fact that \textit{tokName}(\code{:Semillon}) $=$ ``Semillon'',
as illustrated below. We also remove the remaining part ``Kalin Cellars'', because of  
$<$\code{:KalinCellarsSemillon}, \code{:hasMaker}, \code{:KalinCellars}$>$ and the fact that \textit{tokName}(\code{:KalinCellars}) $=$ ``Kalin Cellars''. Hence, \code{:KalinCellarsSemillon} is marked as anonymous. 
\begin{center}
\vspace*{-5mm}
\noindent\begin{minipage}[c]{.3\linewidth}
\begin{align*}
  \overbrace{
  \rlap{$\underbrace{\phantom{\text{\mystrut Kalin\ Cellars}}}_{\text{:KalinCellars}}$}\text{\mystrut Kalin\ Cellars}\ 
  \rlap{$\underbrace{\phantom{\text{\mystrut Semillon}}}_{\text{:Semillon}}$}\text{\mystrut Semillon}\
  }^{\text{:KalinCellarsSemillon}}
\end{align*}
\end{minipage}
\noindent\begin{minipage}[c]{.3\linewidth}
\begin{align*}
  \overbrace{
  \text{\mystrut South\ Australia}\ 
  \rlap{$\underbrace{\phantom{\text{\mystrut Region}}}_{\text{:Region}}$}\text{\mystrut Region}\
  }^{\text{:SouthAustraliaRegion}}
\end{align*}
\end{minipage}
\noindent\begin{minipage}[c]{.3\linewidth}
\begin{align*}
  \overbrace{
  \rlap{$\underbrace{\phantom{\text{\mystrut Exhibit}}}_{\text{:Exhibit}}$}\text{\mystrut exhibit}\
  \text{\mystrut 23}\ 
  }^{\text{exhibit23}}
\end{align*}
\end{minipage}
\end{center}

\noindent 
Anonymizing \code{:KalinCellarsSemillon} causes \nlowl to produce texts like (a) below when asked to describe \code{:KalinCellarsSemillon}, rather than (b), which repeats ``Semillon'' and ``Kalin Cellars'':
\begin{quoting}
{\small 
\noindent (a) This is a strong, dry Semillon. It has a full body. It is made by Kalin Cellars.\\
(b) Kalin Cellars Semillon is a strong, dry Semillon. It has a full body. It is made by Kalin Cellars.
}
\end{quoting}

Similarly, if $t=$ \code{:SouthAustraliaRegion} and we have 
collected the following message triple, the tokenized identifier of $t$ would be shortened from ``South Australia Region'' to ``South Australia''. We use \textit{altTokName} to denote the resulting shortened tokenized identifiers.\footnote{
Strictly speaking, the values of \textit{altTokName} should be shown as sets, since an individual or class can have multiple shortened tokenized identifiers (see below), but we show them as single values for simplicity.}
\begin{center}
$<$\code{:SouthAustraliaRegion}, \code{:isA}, \code{:Region}$>$\\
{\small \textit{tokName}(\code{:SouthAustraliaRegion}) $=$ ``South Australia Region''}, 
{\small \textit{tokName}(\code{:Region}) $=$ ``Region''} \\
{\small \textit{altTokName}(\code{:SouthAustraliaRegion}) $=$ ``South Australia''}\\
\end{center}

\noindent 
Also, if $t=$ \code{:exhibit23} and we have 
collected the following triple, 
\textit{altTokName}(\code{:exhibit23}) would end up containing only numbers (``23''). Hence, \code{:exhibit23}
is marked as anonymous.
\begin{center}
$<$\code{:exhibit23}, \code{:isA}, \code{:Exhibit}$>$\\
{\small \textit{tokName}(\code{:exhibit23}) $=$ ``exhibit 23''}, 
{\small \textit{tokName}(\code{:Exhibit}) $=$ ``exhibit''}
\end{center}

\subsubsection{Obtaining additional alternative tokenized identifiers} \label{improvingTokenizedIdentifiers}

We then collect the tokenized identifiers of all the ancestor classes of $t$, also taking into account equivalent classes; for example, if $t$ has an equivalent class $t'$, we also collect the tokenized identifiers of the ancestor classes of $t'$.
For $t=$ \code{:KalinCellarsSemillon}, we collect the following tokenized identifiers, because \code{:Semillon}, \code{:SemillonOrSauvignonBlanc}, and \code{:Wine} are ancestors of $t$.\footnote{We consider only classes with \owl identifiers when traversing the hierarchy, ignoring classes constructed using \owl operators (e.g., class intersection) that have not been assigned \owl identifiers. In ontologies with multiple inheritance, the same tokenized identifiers of ancestors of $t$ may be obtained by following different paths in the class hierarchy, but we remove duplicate tokenized identifiers. 
}

\begin{center}
{\small \textit{tokName}(\code{:Semillon}) $=$ ``Semillon''}, 
{\small \textit{tokName}(\code{:SemillonOrSauvignonBlanc}) $=$ ``Semillon Or Sauvignon Blanc''}, 
{\small \textit{tokName}(\code{:Wine}) $=$ ``Wine''}
\end{center}

\noindent If 
$\textit{tokName}(t)$ does not contain any of the collected tokenized identifiers of the ancestor classes of $t$, we create additional alternative tokenized identifiers for $t$, also denoted $\textit{altTokName}(t)$, by appending to $\textit{tokName}(t)$ the collected tokenized identifiers of the ancestor classes of $t$.
For example, if $t=$ \code{:red} and \code{:Color} is the parent class of $t$ 
($<$\code{:red}, \code{isA}, \code{:Color}$>$), we 
also obtain ``red color'':
\begin{center}
{\small \textit{tokName}(\code{:red}) $=$ ``red''}, 
{\small \textit{tokName}(\code{:Color}) $=$ ``color''}, 
{\small \textit{altTokName}(\code{:red}) $=$ ``red color''}
\end{center}

\noindent By contrast, if $t=$ \code{:KalinCellarsSemillon}, no  $\textit{altTokName}(t)$  is produced from the ancestors of $t$, because $\textit{tokName}(t) =$ ``Kalin Cellars Semillon'' contains 
\textit{tokName}(\code{:Semillon}) 
$=$ ``Semillon'',
and \code{:Semillon} is an ancestor of \code{:KalinCellarsSemillon}.

Furthermore, we create an additional $\textit{altTokName}(t)$ by removing all numbers from $\textit{tokName}(t)$; for example, from $\textit{tokName}(t)= $ ``Semillon 2006'' we obtain  $\textit{altTokName}(t)= $``Semillon''. Lastly, if $\textit{tokName}(t)$ contains brackets, we create an $\textit{altTokName}(t)$ for each part outside and inside the brackets; 
for example, 
from ``gerbil (dessert rat)'' we get ``gerbil'' and ``dessert rat''.

\subsubsection{Retrieving web pages, extracting and ranking noun phrases} \label{rankingNPs}

\begin{figure}
\begin{tikzpicture}[scale=.7]
\Tree[.S 	[.\textbf{NP} [.DT \textbf{the} ] [.NNP \textbf{Naples} ] [.NNP \textbf{National} ] [.NNP \textbf{Archaeological} ] [.NNP \textbf{Museum} ] ]
					[.VP [.VBZ houses ]
							[.\textbf{NP} 							
										[.NP [.DT \textbf{some} ] ]
										[.PP [.IN \textbf{of} ] 
													[.\textbf{NP} 
																	[.DT \textbf{the} ]
																	[.ADJP [.RBS \textbf{most} ] [.JJ \textbf{important} ] ]
																	[.JJ \textbf{classical} ]
																	[.NNS \textbf{collections} ] ] ] ] ] ]
\end{tikzpicture}
\vspace*{-2.2cm}
\\
{\footnotesize Extracted noun phrases (\nps):}\\
{\footnotesize ``\textbf{the Naples National Archaeological Museum}''}\\
{\footnotesize ``\textbf{some of the most important classical collections}''}\\
{\footnotesize ``\textbf{the most important classical collections}''}
\caption{Parse tree of a retrieved sentence and its noun phrases.}
\label{parseTreeNLName}
\end{figure}
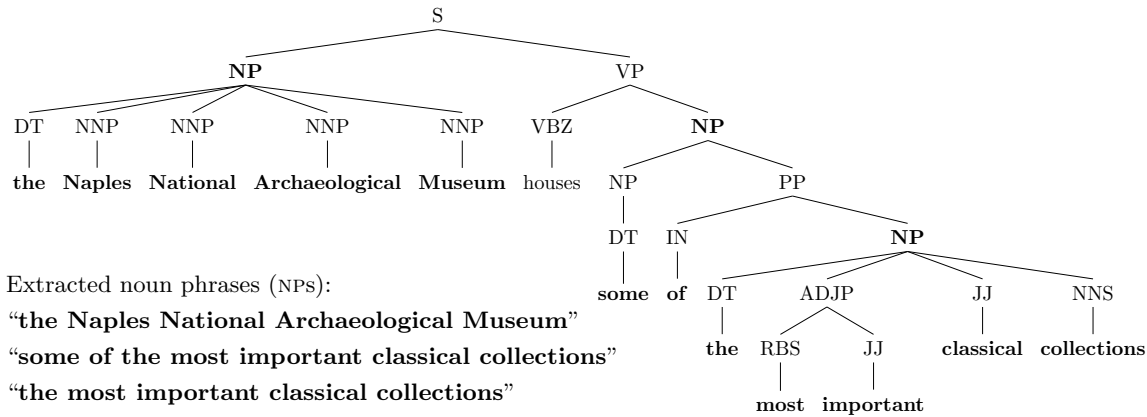

Subsequently, we formulate a Boolean Web search query for $\textit{tokName}(t)$  (e.g., ``South'' \code{AND} ``Australia'' \code{AND} ``Region'') and each  $\textit{altTokName}(t)$ (e.g., ``South'' \code{AND} ``Australia''); recall that $t$ is the individual or class we wish to produce an \nl name for.\footnote{
If the search engine suggests corrections to any query, we use the corrected form of the query as well.}
We convert the retrieved pages of all the queries to plain text documents and parse every sentence of the text, if any stemmed word of the sentence is the same as any stemmed word of any $\textit{tokName}(t)$ or $\textit{altTokName}(t)$.\footnote{We use \textsc{jsoup} (\url{http://jsoup.org/}) to obtain the plain texts from the Web pages, Porter's stemmer (\url{http://tartarus.org/martin/PorterStemmer/}), and the Stanford parser (\url{http://nlp.stanford.edu/}).} We then extract the noun phrases (\nps) from every parsed sentence. For example, from the sentence ``the Naples National Archaeological Museum houses some of the most important classical collections'' we extract the \nps ``the Naples National Archaeological Museum'', ``some of the most important classical collections'', and ``the most important classical collections''  (Fig.~\ref{parseTreeNLName}). 

For each extracted \np, we compute its similarity to $\textit{tokName}(t)$ and each $\textit{altTokName}(t)$. Let \textit{np} be an extracted \np and let \textit{name} be $\textit{tokName}(t)$ or an $\textit{altTokName}(t)$. To compute the similarity between \textit{np} and \textit{name}, we first compute the character-based Levenshtein distance between each token of \textit{np} and each token of \textit{name}; we ignore upper/lower case differences, articles, and connectives
(e.g. ``or''), which are often omitted from \owl identifiers. In the following example, \textit{np} $=$ ``the Naples National Archaeological Museum'' (but 
``the'' is ignored) and \textit{name} = ``national arch napoli museum''; 
this \textit{name} is an  $\textit{altTokName}(t)$ produced by appending to  $\textit{tokName}(t)$ the tokenized identifier of the parent class (\code{:Museum}) of $t$  (Section~\ref{improvingTokenizedIdentifiers}). The Levenshtein distance between 
``national'' and ``National'' is $0$ (upper/lower case differences are ignored).
The distance between ``napoli'' and ``Naples'' is $4$; a character deletion or insertion costs 1, 
a replacement costs 2. 
\begin{center}
\begin{tikzpicture}
\matrix[column sep=0em,row sep=.4in] {
\node[outer sep=0pt,anchor=base] (T) {}; & \node[outer sep=0pt,anchor=base] (A) {national}; & \node[outer sep=0pt,anchor=base] (B) {arch}; & \node[outer sep=0pt,anchor=base] (C) {napoli}; & \node[outer sep=0pt,anchor=base] (D) {museum}; &\\
\node[outer sep=0pt,anchor=base] (t) {(the)}; & \node[outer sep=0pt,anchor=base] (a) {Naples}; & \node[outer sep=0pt,anchor=base] (b) {National}; & \node[outer sep=0pt,anchor=base] (c) {Archaeological};  & \node[outer sep=0pt,anchor=base] (d) {Museum}; & \\
};
\draw (A) -- (b) node[draw=none,fill=none,font=\scriptsize,midway,left] {0};
\draw (B) -- (c) node[draw=none,fill=none,font=\scriptsize,midway,right] {10};
\draw (C) -- (a) node[draw=none,fill=none,font=\scriptsize,midway,above] {4};
\draw (D) -- (d) node[draw=none,fill=none,font=\scriptsize,midway,left] {0};
\end{tikzpicture}
\end{center}

\noindent We then form pairs of aligned tokens $\left<t_{\mathit{name}}, t_{\mathit{np}} \right>$, where $t_{\mathit{name}}$,  $t_{\mathit{np}}$  are tokens from \textit{name}, \textit{np}, respectively, such that each token of \textit{name} is aligned to at most one token of \textit{np} and vice versa, and any other, not formed pair $\left<t_{\mathit{name}}', t_{\mathit{np}}' \right>$ would have a Levenshtein distance (between $t_{\mathit{name}}'$, 
$t'_{\mathit{np}}$) larger or equal to the minimum Levensthein distance of the formed pairs.\footnote{In our experiments, we actually formed the pairs 
greedily, by computing the 
distances of all the possible pairs and iteratively selecting the pair with the smallest 
distance whose elements 
did not occur in any other already selected pair. A non-greedy search 
(e.g., using Integer Linear Programming) 
makes little difference in practice.} In 
our example, the pairs of alinged tokens are $<$``national'', ``National''$>$, $<$``arch'', ``Archaeological''$>$, $<$``napoli'', ``Naples''$>$, $<$``museum'', ``Museum''$>$.

The similarity between \textit{np} and \textit{name} is then computed as follows, where $A$ is the set of aligned token pairs, $\mathit{Levenshtein}(a)$ is the Levenshtein distance
(normalized to $[0,1]$)
between the $t_{\mathit{name}}$ and $t_{\mathit{np}}$ of pair $a$, 
and $|\mathit{np}|$, $|\mathit{name}|$ are the lengths (in tokens) of \textit{np} and \textit{name}, respectively.

\begin{equation}\label{levenSimilarityFunction}
\mathit{similarity}(\mathit{np}, \mathit{name}) = 
\frac{\sum_{a \in A}(1 - \mathit{Levenshtein}(a))}
{max\{|\mathit{np}|, |\mathit{name}|\}}
\end{equation}

For each extracted \np of $t$, we compute its similarity to every possible \textit{name}, i.e., $\textit{tokName}(t)$ or $\textit{altTokName}(t)$, as discussed above, and we assign to the \np a score equal to the largest of these similarities. Finally, we rank the extracted \nps of $t$ by decreasing score. If two \nps have the same score, we rank higher the \np with the fewest crossed edges in its best alignment with a \textit{name}.
If two \nps still cannot be distinguished, we rank them by decreasing frequency in the parsed sentences of $t$; and if their frequencies are equal, we rank them randomly.

\subsection{Turning the extracted noun phrases into 
natural language names} \label{NLNameGeneration}

The extracted \nps 
are not yet \nl names, because they lack the  linguistic annotations that \nlowl requires (e.g., \pos tags, 
agreement, number); they may also lack appropriate articles. To convert an \np to an \nl name, we first obtain the \pos tags of its words from the parse tree of the sentence the \np was extracted from.\footnote{If an \np has been extracted from multiple sentences and their parse trees provide different \pos tag assignments to the words of the \np, we create a separate \nl name for each \pos tag assignment.} For example, the \np ``the Red Wine'' becomes:
\smallskip
\begin{center}
\noindent
the$_{\pos=\textit{DT}}$ Red$_{\pos=\textit{JJ}}$ Wine$_{\pos=\textit{NN}}$
\end{center}

For every noun, adjective, article, 
preposition, we create a corresponding slot in the \nl name; all the other words of the \np become slots
containing the words as fixed strings (Section~\ref{NLnamesBackground}). For nouns and adjectives, the base form 
is used in the slot (e.g., ``wine'' istead of ``wines''), but slot annotations 
indicate the particular 
inflectional form that was used in the \np; 
e.g., the \textsc{nn} \pos tag shows that ``wine'' is 
singular.
A named-entity recognizer (\textsc{ner}) and an on-line dictionary are employed to detect nouns that refer to persons and locations.\footnote{We use the Stanford \textsc{ner} (\url{http://nlp.stanford.edu/}), and Wiktionary (\url{http://en.wiktionary.org/}).} The genders of these nouns are determined using the on-line dictionary, when possible, or defaults otherwise (e.g., the default for person nouns is a `person' pseudo-gender, which leads to ``he/she'' or ``they'' when generating a pronoun). Nouns
not referring to persons and locations are marked as neuter. Since
the \nps are extracted from Web pages, there is a
risk of wrong capitalization 
(e.g.,  ``the RED wine''). For each word of the \nl name, we pick the capitalization that is most frequent in the retrieved texts of 
the individual or class we generate the \nl name for. Hence, the \np ``the Red Wines'' becomes:
\begin{center}
[]$^{1}_{\textit{article}, \,  \textit{def}}$ 
[red]$^{2}_{\textit{adj}}$ 
[wine]$^{3}_{\textit{noun}, \, \textit{sing}, \, \textit{neut}}$ 
\end{center}

\noindent 
which requires a definite article, followed by the adjective ``red'', 
and the neuter ``wine'' in singular. 

A dependency parser is then used to identify the head 
of each \nl name (Section~\ref{NLnamesBackground}) and to obtain agreement information.\footnote{
The Stanford parser also produces  dependency trees. The parser is applied to the sentences the \nps were extracted from. If multiple parses are obtained for the \np an \nl name is based on, we keep the most frequent parse.} Adjectives are required to agree with the nouns they modify, and the same applies to articles and nouns. 
At this stage, the \np ``the Red Wines'' will have become:
\begin{center}
[]$^{1}_{\textit{article}, \,  \textit{def}, \, \textit{agr}=3}$ 
[red]$^{2}_{\textit{adj}}$ 
[wine]$^{3}_{\textit{headnoun}, \, \textit{sing}, \, \textit{neut}}$ 
\end{center}

We then consider the main article (or, more generally, determiner) of the \nl name, i.e., the article
that agrees with the head
(e.g., ``a'' in 
``a traditional wine from the Piemonte Region''). Although the \nl name may already include a main article, it is not necessarily an appropriate one. For example, it would be inappropriate to use a definite article in ``\emph{The} red wine is a kind of wine with red color'', when describing the class of red wines. We modify the \nl name to use an indefinite article if the \nl name refers to a class, and a definite article if it refers to an individual (e.g., ``the South Australia region'').\footnote{This arrangement works well in the ontologies we have experimented with, but it may have to be modified 
for other ontologies, for example if kinds of entities are modelled as individuals, rather than classes.} The article is omitted if the head is an adjective (e.g., ``strong''), 
or in plural (e.g., ``Semillon grapes''),  
or if the entire \nl name (excluding the article, if present) is a proper name (e.g., ``South Australia'')
or a mass noun phrase 
without article (e.g., ``gold''). Before inserting or modifying the main article, we also remove any demonstratives (e.g., ``\emph{this} statue'') or other non-article determiners (e.g., ``some'', ``all'') from the beginning of the \nl name . In our example, the \nl name is to be used to refer to the class \code{:RedWine}, so the final \nl name is the following, which would lead to sentences like ``\emph{A} red wine is a kind of wine with red color''.

\begin{center}
[]$^{1}_{\textit{article}, \,  \textit{indef}, \, \textit{agr}=3}$ 
[red]$^{2}_{\textit{adj}}$ 
[wine]$^{3}_{\textit{headnoun}, \, \textit{sing}, \, \textit{neut}}$ 
\end{center}

\noindent Recall that \nlowl can automatically adjust \nl names when generating texts (Section~\ref{NLnamesBackground}). For example, in a comparison like ``Unlike \emph{the previous red wines} that you have seen, this one is from France'', it would use a definite article and it would turn the head noun of the \nl name to plural, also adding the adjective ``previous''.
The resulting \nl names are 
finally ranked by the scores of the \np{s} they were obtained from (Section~\ref{rankingNPs}).

\subsection{Inferring interest scores from
natural language names} \label{InterestReasoning}

The reader may have already noticed that the sentence ``A red wine is a kind of wine with red color'' that we used above sounds redundant. 
Some message triples lead to sentences that sound redundant, because they report relations that are obvious (to humans) from the \nl names of the 
individuals or classes. In our 
example, the sentence reports the following two message triples. 
\begin{center}
$<$\code{:RedWine}, \code{isA}, \code{:Wine}$>$, 
$<$\code{:RedWine}, \code{:hasColor}, \code{:Red}$>$\\
\end{center}

\noindent Expressed separately, the two triples would lead to the sentences ``A red wine is a kind of wine'' and ``A red wine has red color'', but \nlowl aggregates them into a single sentence. The ``red color'' derives from an  $\textit{altTokName}$ of \code{:Red} obtained by considering the parent class \code{:Color} of \code{:Red} (Section~\ref{improvingTokenizedIdentifiers}). It is obvious that a red wine is a wine with red color and, hence, the two triples above should not be expressed. Similarly, the following triple leads to the sentence ``A white Bordeaux wine is a kind of Bordeaux'', which again seems redundant.
\begin{center}
$<$\code{:WhiteBordeaux}, \code{isA}, \code{:Bordeaux}$>$
\end{center}

\nlowl provides mechanisms 
to manually assign \emph{interest scores} to message triples \cite{Androutsopoulos2013}. Assigning a zero interest score to a triple instructs \nlowl to avoid expressing it. Manually assigning interest scores, however, can be tedious. Hence, we aimed to automatically assign zero scores to triples like the ones above, which report relations that are obvious from the \nl names. To identify triples of this kind, we follow a procedure similar to the one 
we use to identify individuals or classes that should be anonymous (Section~\ref{shorteningIdentifiers}). For each $\left<S, R, O\right>$ triple that involves the individual or class $S$ being described, we 
examine the \nl names
of $S$ and $O$.
If all the (lemmatized) words of the phrase produced by the \nl name of $O$ (e.g., ``a Bordeaux''),
excluding articles, appear in the phrase 
of the \nl name of $S$ (e.g., ``a white Bordeaux''), 
we assign a zero interest score to $\left<S, R, O\right>$.
\begin{align*}
  \overbrace{
  \text{\mystrut white}\ 
  \rlap{$\underbrace{\phantom{\text{\mystrut Bordeaux}}}_{\text{:Bordeaux}}$}\text{\mystrut Bordeaux}\
  }^{\text{:WhiteBordeaux}}
\end{align*}

\section{Our method to automatically extract sentence plans from the Web} \label{OurMethodSP}

To produce a sentence plan for a relation, we first extract slotted string templates (e.g., ``$X$ is made from $Y$'') from the Web using seeds 
(values of $X,Y$) from the ontology. 
We then enhance the templates by adding linguistic annotations (e.g., \pos tags, 
agreement, voice, tense) and missing components (e.g., auxiliary verbs) turning the templates into candidate sentence plans. 
The candidate sentence plans are then scored by a Maximum Entropy classifier
to identify the best few (again up to 5) candidate sentence plans for each relation.\footnote{
We use the Stanford classifier; consult \url{http://nlp.stanford.edu/software/classifier.shtml}.} 
In a fully automatic scenario, the sentence plan that the classifier considers best for each relation is used. In a semi-automatic scenario, the few top sentence plans of each relation are shown to a human author, who picks the best one.

\subsection{Extracting templates from the Web} \label{PatternExtraction}

For each relation $R$ that we want to generate a sentence plan for, our method first obtains the \owl statements of the ontology that involve the relation and turns them into message triples $\left<S, R, O\right>$, as when generating texts.
For example, if the relation is \code{:madeFrom}, two of the 
triples may be:
\begin{center}
$<$\code{:StEmilion}, \code{:madeFrom}, \code{:cabernetSauvignonGrape}$>$,
$<$\code{:Semillon}, \code{:madeFrom}, \code{:SemillonGrape}$>$
\end{center}

\noindent To these triples, we add more by replacing the $S$, $O$, or both of each originally obtained triple by their classes (if $S$ or $O$ are individuals), their parent classes, or their equivalent classes. For example,  from $<$\code{:StEmilion}, \code{:madeFrom}, \code{:cabernetSauvignonGrape}$>$ we also obtain the following three triples, because \code{Wine} is a parent class of \code{StEmilion}, and \code{Grape} is a parent class of \code{:cabernetSauvignonGrape}. 
\begin{center}
$<$\code{:Wine}, \code{:madeFrom}, \code{:cabernetSauvignonGrape}$>$\\
$<$\code{:StEmilion}, \code{:madeFrom}, \code{:Grape}$>$, 
$<$\code{:Wine}, \code{:madeFrom}, \code{:Grape}$>$
\end{center}
We obtain the same additional triples from $<$\code{:Semillon}, \code{:madeFrom}, \code{:SemillonGrape}$>$, because \code{Wine} and \code{Grape} are also parent classes of \code{Semillon} 
and \code{SemillonGrape}, 
but we remove duplicates. 

Each $\left<S, R, O\right>$ triple is then replaced by a pair $\left<\textit{seedName}(S), \textit{seedName}(O)\right>$, where $\textit{seedName}(S)$ is a word sequence generated by the \nl name of $S$, and similarly for $O$. 
We assume that the \nl names are manually authored,
or that they are generated by our method of Section \ref{OurMethodNLN}. In the latter case, we keep only one \nl name per individual or class, the one selected by the human author (in a semi-automatic setting of \nl name generation) or the top ranked \nl name (in a fully automatic setting). The five triples above become the following pairs. We call pairs of this kind \emph{seed name pairs}, and their elements \emph{seed names}. If a seed name results from a class, parent-class, or an equivalent class of the original $S$ or $O$, we consider it a \emph{secondary seed name}.
\begin{center}
{\small $<$``St.\ Emilion'',  ``Cabernet Sauvignon grape''$>$}, 
{\small $<$``Semillon'', ``Semillon grape''$>$}\\
{\small $<$``wine'', ``Cabernet Sauvignon grape''$>$}, 
{\small $<$``St.\ Emilion'', ``grape''$>$}, 
{\small $<$ ``wine'', ``grape''$>$}
\end{center}

\begin{figure}
\noindent\begin{minipage}[c]{.56\linewidth}
\begin{tikzpicture}[scale=.69]
\Tree[.S [.ADVP [.RB \textit{obviously} ] ] 
				 [.\textbf{NP} [.NNP \textbf{Semillon} ] ] 
         [.VP [.VBZ \textit{is} ] 
              [.VP [.VBN \textit{made} ] 
									 [.PP [.IN \textit{from} ] 
												[.NP [.\textbf{NP} [.JJ \textbf{Semillon} ] [.NNS \textbf{grapes} ] ] [.PP [.IN \textit{in} [.NP [.NNP California ] ] ] ] ] ] ] ] ]
\end{tikzpicture}
\end{minipage}
\noindent\begin{minipage}[c]{.44\linewidth}
{\footnotesize Seed name pairs:}\\
{\footnotesize $<$``St.\ Emilion'',  ``Cabernet Sauvignon grape''$>$}\\
{\footnotesize $<$``\textbf{Semillon}'', ``\textbf{Semillon grape}''$>$}\\
{\footnotesize $<$``wine'', ``Cabernet Sauvignon grape''$>$}\\
{\footnotesize $<$``St.\ Emilion'', ``grape''$>$}\\
{\footnotesize $<$ ``wine'', ``grape''$>$}\\
\\
{\footnotesize \np anchor pair:}\\
{\footnotesize $<$``\textbf{Semillon}'', ``\textbf{Semillon grapes}''$>$}\\
\end{minipage}
\caption{Parse tree of a retrieved sentence and its single \np anchor pair.}
\label{parseTree}
\end{figure}
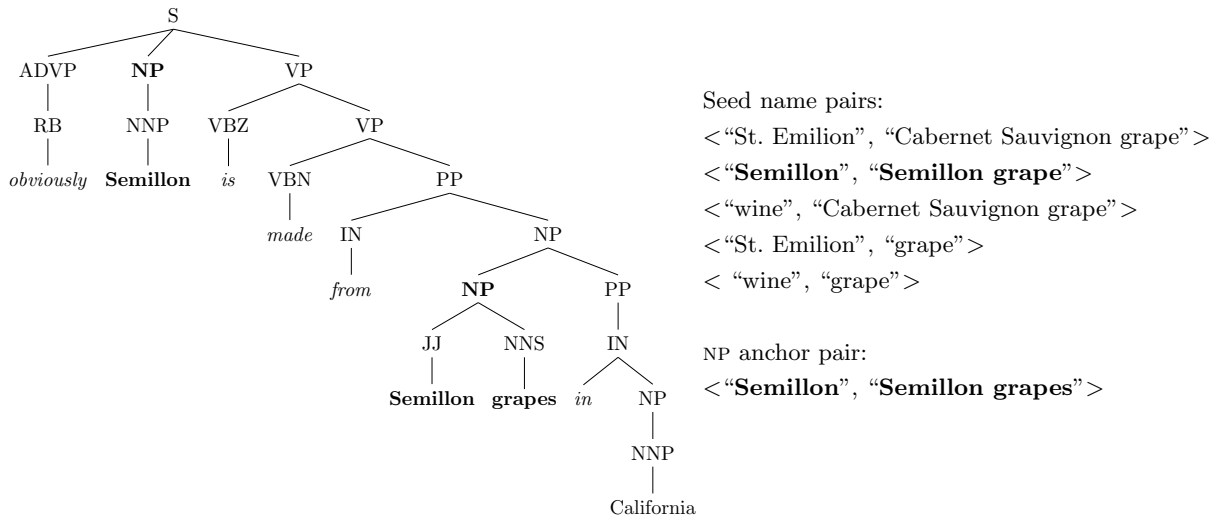

We then retrieve Web pages using the seed name pairs (of the relation that we want to generate a sentence plan for) as queries. For each seed name pair, we use the conjunction of its seed names (e.g., ``St. Emilion'' \code{AND} ``Cabernet Sauvignon grape'') as a Boolean query.\footnote{
We do not consider corrections suggested by the search engine, since the \nl names are assumed to be correct.} We convert all the retrieved pages (of all the seed name pairs) to plain text documents and parse every sentence of the retrieved documents, if at least one stemmed word from
each seed name of a particular pair is the same as a stemmed word of the sentence.
We then keep every parsed sentence that contains at least two \nps matching a seed name pair. For example, the sentence ``obviously Semillon is made from Semillon grapes in California'' contains the \nps ``Semillon'' and ``Semillon grapes'' that match the seed name pair $<$``Semillon'', ``Semillon grape''$>$ (Fig.~\ref{parseTree}). Two \nps of a sentence match a seed name pair if the similarity between any of the two \nps and any of the two seed names (e.g., the first \np and the second seed name) is above a threshold 
$T$ and the similarity between the other \np and the other seed name is also above $T$. The similarity between an \np and a seed name is computed as their weighted cosine similarity, with $\textit{tf}\cdot\textit{idf}$ weights, applied to stemmed \nps and seed names, ignoring stop-words.\footnote{We also remove accents, and replace all numeric tokens by a particular pseudo-token.} The $\textit{tf}$ of a word of the \np or seed name is the frequency (usually 0 or 1) of the word in the \np or seed name, respectively; the $\textit{idf}$ is the inverse document frequency of the word in all the retrieved documents of the relation. 
We call \emph{\np anchor pair} any 
two \nps of a parsed sentence that match a seed name pair,
and \emph{\np anchors} the elements of an \np anchor pair. 

From every parsed sentence that contains an \np anchor pair, we produce a slotted string template by replacing the first \np anchor by $S$, the second \np anchor by $O$, including between $S$ and $O$ in the template the words of the sentence that were between the two \np anchors, and discarding the other words of the sentence. In the example of Fig.~\ref{parseTree}, we would obtain the template ``$S$ is made from $O$''. Multiple templates may be extracted from the same sentence, if a sentence contains more than one \np anchor pairs; and the same template may be extracted from multiple sentences, possibly retrieved by different seed name pairs. We retain only the templates that were extracted from at least two different sentences. We then produce additional templates by increasingly extending the retained ones up to the boundaries of
the sentences they were extracted from. In 
Fig.~\ref{parseTree}, 
if the template ``$S$ is made from $O$'' has been retained, we would also produce the following templates. 

\medskip
{\small
\noindent
\itab{obviously $S$ is made from $O$}  \tab{obviously $S$ is made from $O$ in} \tab{obviously $S$ is made from $O$ in California}\\
\itab{$S$ is made from $O$ in}  \tab{$S$ is made from $O$ in California}
}
\medskip

\noindent Again, we discard extended templates that did not result from at least two sentences.

\subsection{Turning the templates into candidate sentence plans} \label{SentencePlanGeneration}

The templates 
 (e.g., ``$S$ is made from $O$'') are not yet sentence plans, because they lack the linguistic annotations that \nlowl requires 
(e.g., \pos tags, agreement, voice, tense, cases), and they may also not correspond to well-formed sentences (e.g., they may lack verbs). The conversion of a template to a (candidate) sentence plan is similar to the conversion of 
Section \ref{NLNameGeneration}. We start by obtaining \pos tags from the
parse trees of the sentences the template was obtained from. Recall that a template may have been extracted from multiple sentences. We obtain a \pos tag sequence for the words of the template from each one of the sentences the template was extracted from, and we keep the most frequent \pos tag sequence. We ignore the \pos tags of the anchor \nps, which become $S$ and $O$ in the template. For example, the template ``$S$ is made from $O$'' becomes:
\smallskip
\begin{center}
\noindent
$S$ is$_{\pos=\textit{VBZ}}$ made$_{\pos=\textit{VBN}}$ from$_{\pos=\textit{IN}}$ $O$
\end{center}

For every verb form (e.g., ``is made''), noun, adjective, and preposition, we create a corresponding slot in the sentence plan. For verbs, nouns, and adjectives, the base form is used in the slot; each verb slot is also annotated with the voice and tense of the corresponding verb form in the template.
If a negation expression (e.g., ``not'', ``aren't'') is used with the verb form in the template, the negation expression is not included as a separate slot in the  sentence plan, but the polarity of the verb slot is marked as negative; otherwise the polarity is positive. We determine the genders and capitalizations of nouns (and proper names) 
as in Section \ref{NLNameGeneration}. The $S$ and $O$ are also replaced by slots requiring referring expressions. For example, the template ``$S$ is made from $O$'' 
becomes:
\begin{center}
\noindent
[$\mathit{ref}(S)$]$^{1}$
[make]$^{2}_{\textit{verb}, \, \textit{passive}, \, \textit{present}, \, \textit{polarity}=+}$ 
[from]$^{3}_{prep}$ 
[$\mathit{ref}(O)$]$^{4}$
\end{center}

Agreement and case information is obtained using a dependency parser.
The parser is applied to the sentences the templates were extracted from, keeping the most frequent parse per template. Referring expressions obtained from \np anchors that were verb subjects are marked with nominative case, and they are required to agree with their verbs. Referring expressions corresponding to verb objects or preposition complements are marked with accusative case (e.g., ``from \emph{him}''). Referring expressions corresponding to \np anchors with head nouns in possessive form (e.g., ``Piemonte's'') are marked with possessive case. In our example, we obtain: 
\begin{center}
[$\mathit{ref}(S)$]$^{1}_{\textit{case}=\textit{nom}}$
[make]$^{2}_{\textit{verb}, \, \textit{passive}, \, \textit{present}, \, \textit{agr}=1, \, \textit{polarity}=+}$ 
[from]$^{3}_{prep}$ 
[$\mathit{ref}(O)$]$^{4}_{\textit{case}=\textit{acc}}$
\end{center}

Any remaining words of the template that have not been replaced by slots (e.g., ``obviously'' in ``obviously $S$ is made from $O$'') are turned into fixed string slots. Subsequently, any sentence plan that has only two slots, starts with a verb, or contains no verb, 
is discarded, because sentence plans of these kinds tend to 
be poor.
Also, if a sentence plan contains a single verb in the past participle, in agreement with either $S$ or $O$, followed by a preposition (e.g. ``$S$ made in $O$''), we insert an auxiliary verb to turn the verb form into present passive (e.g., ``$S$ is made in $O$''); in domains other than those of our experiments, a past passive may be more appropriate (``$S$ was made in $O$). Similarly, if a single verb appears in the present participle (e.g. ``$S$ making $O$''), we insert an auxiliary verb to obtain a present continuous form. Both cases are illustrated below. 

\smallskip
\noindent
\begin{minipage}{.5\linewidth}
\begin{equation*}
\text{$S$ made in $O$} \Rightarrow \text{$S$ \underline{is made} in $O$}
\end{equation*}
\end{minipage}
\begin{minipage}{.5\linewidth}
\begin{equation*}
\text{$S$ making $O$} \Rightarrow \text{$S$ \underline{is} making $O$}
\end{equation*}
\end{minipage}
\smallskip

Lastly, we filter the remaining sentence plans through a Web search engine. For this step, we replace referring expression slots by wildcards, we generate the rest of the sentence (e.g., ``* is made from *''), and we do a phrase search.
If no results are returned, the sentence plan is discarded.

\subsection{Applying a Maximum Entropy classifier to the candidate sentence plans} \label{EvaluatingSentencePlans}

The retained candidate sentence plans are then scored using a Maximum Entropy (MaxEnt) classifier.
The classifier views each candidate sentence plan $p$ for a relation $R$ as a vector of 251 features, and attempts to estimate the probability that $p$ is a good sentence plan (positive class) for $R$ or not (negative class). The 251 features provide information about  
$p$ itself, but also about the templates, seed name pairs, and \np anchor pairs of $p$, meaning the templates that $p$ was obtained from,
and the seed name pairs and \np anchor pairs (Fig.~\ref{parseTree}) that matched to produce the templates of $p$.

\subsubsection{Productivity features} \label{productivityFeatures}

The \emph{productivity of the $i$-th seed name pair} $\left<n_{i,1}, n_{i,2}\right>$ (e.g., $<$$n_{i,1} =$``Semillon'', $n_{i,2} =$ ``Semillon grape''$>$) of a relation $R$ (e.g., $R=$ \code{:madeFrom}) is defined as follows:
\begin{equation}\label{productivityParentSeedNamePairFunction}
\mathit{productivity}(\left<n_{i,1}, n_{i,2}\right>|R) = 
\frac{\mathit{hits}(\left<n_{i,1}, n_{i,2}\right>|R)}
{\sum_{j=1}^{N}{\mathit{hits}(\left<n_{j,1}, n_{j,2}\right>|R)}}
\end{equation}
where: $\mathit{hits}(\left<n_{i,1}, n_{i,2}\right>|R)$ is the number of times $\left<n_{i,1}, n_{i,2}\right>$ matched any \np anchor pair of the parsed sentences of $R$, counting only matches that contributed to the extraction (Section \ref{PatternExtraction}) of \emph{any} template of $R$; $N$ is the total number of seed name pairs of $R$; and $\left<n_{j,1}, n_{j,2}\right>$ is the $j$-th seed name pair of $R$.\footnote{In 
the function $\mathit{hits}(\left<n_1, n_2\right>|R)$, we actually multiply by  $\frac{1}{2}$ the value 
of the function if exactly one of $n_1, n_2$ is a secondary seed name (Section~\ref{PatternExtraction}), and we multiply by $\frac{1}{4}$ if both $n_1, n_2$ are secondary seed names. The same applies to the function $\mathit{hits}(\left<n_1, n_2\right>,t|R)$ of Eq.~\ref{productivityParentTemplateParentSeedNameFunction} and the function $ \mathit{hits}(\left<n_1, n_2\right>, \left<a_1, a_2\right>, t|R)$ of Eq.~\ref{productivityTemplateSeedNameAnchorFunction} below.} The intuition behind $\mathit{productivity}(\left<n_{i,1}, n_{i,2}\right>|R)$ is that seed name pairs that match \np anchor pairs of many sentences of $R$ are more likely to be indicative of $R$. When using the MaxEnt classifier to score a sentence plan $p$ for a relation $R$, we compute the $\mathit{productivity}(\left<n_{i,1}, n_{i,2}\right>|R)$ of 
\emph{all} the seed name pairs $\left<n_{i,1}, n_{i,2}\right>$ of $p$, and we use the \emph{maximum}, \emph{minimum}, \emph{average}, \emph{total}, and \emph{standard deviation} of these productivity scores as five features of $p$. 

The \emph{productivity of a seed name} $n_1$ (considered on its own) that occurs as the first element of at least one seed name pair $\left<n_{i,1}, n_{i,2}\right> = \left<n_1, n_{i,2}\right>$ of a relation $R$ is defined as follows:
\begin{equation}\label{productivityParentSeedNameFunction}
\mathit{productivity}(n_1| R) = 
\frac{\sum_{i=1}^{N_2}{\mathit{hits}(\left<n_1, n_{i,2}\right>|R)}}
{\sum_{j=1}^{N}{\mathit{hits}(\left<n_{j,1}, n_{j,2}\right>|R)}}
\end{equation}
where: $N_2$ is the number of seed name pairs $\left<n_1, n_{i,2}\right>$ of $R$ that have $n_1$ as their first element; $\mathit{hits}(\left<n_1, n_{i,2}\right>|R)$ is the number of times $n_1$ (as part of a seed name pair $\left<n_1, n_{i,2}\right>$ of $R$) matched any element of any \np anchor pair of the parsed sentences of $R$ and contributed to the extraction of \emph{any} template of $R$; $N$ and $\mathit{hits}(\left<n_{j,1}, n_{j,2}\right>|R)$ are as in Eq.~\ref{productivityParentSeedNamePairFunction}. Again, when using the classifier to score a sentence plan $p$ for a relation $R$, we calculate the $\mathit{productivity}(n_1|R)$ values of 
\emph{all} the  
(distinct) seed names $n_1$ that occur as first elements in the seed name pairs of $p$, and we use the \emph{maximum}, \emph{minimum}, \emph{average}, \emph{total}, and \emph{standard deviation} of these productivity scores as five more features of $p$. We define similarly $\mathit{productivity}(n_2| R)$ for a seed name $n_2$ that occurs as the \emph{second} element in any seed name pair $\left<n_{i,1}, n_2\right>$ of $R$, obtaining five more features for $p$.

Similarly to Eq.~\ref{productivityParentSeedNamePairFunction}, we define the \emph{productivity of the $i$-th NP anchor pair} $\left<a_{i,1}, a_{i,2}\right>$ (e.g., $<$$a_{1,1} =$``Semillon'', $a_{1,2} =$ ``Semillon grapes''$>$ in Fig.~\ref{parseTree}) of a relation $R$ as follows:
\begin{equation}\label{productivityParentAnchorPairFunction}
\mathit{productivity}(\left<a_{i,1}, a_{i,2}\right>|R) = 
\frac{\mathit{hits}(\left<a_{i,1}, a_{i,2}\right>|R)}
{\sum_{j=1}^{A}{\mathit{hits}}(\left<a_{j,1}, a_{j,2}\right>|R)}
\end{equation}
where: $\mathit{hits}(\left<a_{i,1}, a_{i,2}\right>|R)$ is the number of times a seed name pair of $R$ matched  $\left<a_{i,1}, a_{i,2}\right>$ in the parsed sentences of $R$ and contributed to the extraction of \emph{any} template of $R$; and $A$ is the total number of \np anchor pairs of $R$.\footnote{In 
the function $\mathit{hits}(\left<a_1, a_2\right>|R)$, we actually multiply by $\frac{1}{2}$ any match of $\left<a_1, a_2\right>$ with a seed name pair $\left<n_1, n_2\right>$ if exactly one of $n_1, n_2$ is a secondary seed name, 
and by $\frac{1}{4}$ if both $n_1, n_2$ are secondary seed names.} As with $\mathit{productivity}(\left<n_{i,1}, n_{i,2}\right>|R)$, the intuition behind $\mathit{productivity}(\left<a_{i,1}, a_{i,2}\right>|R)$ is that \np anchor pairs that match many seed name pairs of $R$ are more indicative of $R$. When using the classifier to score a sentence plan $p$ for a relation $R$, we compute the $\mathit{productivity}(\left<a_{i,1}, a_{i,2}\right>|R)$ of 
\emph{all} the \np anchor pairs of $p$, and we use the \emph{maximum}, \emph{minimum}, \emph{average}, \emph{total}, and \emph{standard deviation} of these scores as five additional features of $p$. 

Similarly to Eq.~\ref{productivityParentSeedNameFunction}, the \emph{productivity of an NP anchor} $a_1$ (considered on its own) that occurs as the first element of at least one \np anchor pair $\left<a_{i,1}, a_{i,2}\right> = \left<a_1, a_{i,2}\right>$ of $R$ is defined as follows:
\begin{equation}\label{productivityParentAnchorFunction}
\mathit{productivity}(a_1|R) = 
\frac{\sum_{i=1}^{A_2}{\mathit{hits}(\left<a_1, a_{i,2}\right>|R)}}
{\sum_{j=1}^{A}{\mathit{hits}(\left<a_{j,1}, a_{j,2}\right>|R)}}
\end{equation}
where: $A_2$ is the number of \np anchor pairs $\left<a_1, a_{i,2}\right>$ of $R$ that have $a_1$ as their first element; $\mathit{hits}(\left<a_1, a_{i,2}\right>|R)$ is the number of times $a_1$ (as part of an \np anchor pair $\left<a_1, a_{i,2}\right>$ of $R$) matched any element of any seed name pair of $R$ and contributed to the extraction of \emph{any} template of $R$; and $A$, $\mathit{hits}(\left<a_{j,1}, a_{j,2}\right>|R)$ are as in Eq.~\ref{productivityParentAnchorPairFunction}. Again, 
we calculate the $\mathit{productivity}(a_1|R)$ values of 
\emph{all} the 
(distinct) \np anchors $a_1$ that occur as first elements in the \np anchor pairs of $p$, and we use the \emph{maximum}, \emph{minimum}, \emph{average}, \emph{total}, and \emph{standard deviation} of these productivity scores as five more features of $p$. We define similarly $\mathit{productivity}(a_2| R)$ for a seed name $a_2$ that occurs as the \emph{second} element in any \np anchor pair $\left<a_{i,1}, a_2\right>$ of $R$, obtaining five more features for $p$. 

The \emph{productivity of a template} $t$ (e.g., ``$S$ is made from $O$'') of a relation $R$ is defined as follows:
\begin{equation}\label{productivityParentTemplateFunction}
\mathit{productivity}(t|R) = 
\frac{\mathit{hits}(t|R)}
{\sum_{k=1}^{T}{\mathit{hits}(t_k|R)}}
\end{equation}
where: $\mathit{hits}(t|R)$ is the number of times the \emph{particular} template $t$ was extracted from any of the parsed sentences of $R$; $T$ is the total number of templates of $R$; and $t_k$ is the $k$-th template of $R$. The intuition is that templates that are produced more often for $R$ are more indicative of $R$. 
Again, we calculate the $\mathit{productivity}(t|R)$ 
of 
\emph{all} the templates $t$ of $p$, and we use the \emph{maximum}, \emph{minimum}, \emph{average}, \emph{total}, and \emph{standard deviation} of these productivity scores as five more features of $p$. 

The \emph{productivity of a parsed sentence} $s$ (e.g., ``obviously Semillon is made from Semillon grapes in California'') of a relation $R$ is defined as follows:
\begin{equation}\label{productivityParentSentenceFunction}
\mathit{productivity}(s|R) = 
\frac{\mathit{hits}(s|R)}
{\sum_{l=1}^{L}{\mathit{hits}(s_l|R)}}
\end{equation}
where: $\mathit{hits}(s|R)$ is the number of times any template of $R$ was extracted from the \emph{particular} parsed sentence $s$; $L$ is the total number of parsed sentences of $R$; and $s_l$ is the $l$-th parsed sentence of $R$. The intuition is that sentences that produce more templates for $R$ are more indicative of $R$. 
Again, we calculate the $\mathit{productivity}(s|R)$ 
of 
\emph{all} the parsed sentences $s$ of $p$, and we use the \emph{maximum}, \emph{minimum}, \emph{average}, \emph{total}, and \emph{standard deviation} of these productivity scores as 
features of $p$. 

The \emph{joint productivity} of a seed name pair $\left<n_1, n_2\right>$ and a template $t$ of a relation $R$ is:
\begin{equation}\label{productivityParentTemplateParentSeedNameFunction}
\mathit{productivity}(\left<n_1, n_2\right>, t|R) = 
\frac{\mathit{hits}(\left<n_1, n_2\right>, t|R)}
{\sum_{j=1}^{N}{\sum_{k=1}^{T}{{\mathit{hits}(\left<n_{j,1}, n_{j,2}\right>, t_k|R)}}}}
\end{equation}
where: $\mathit{hits}(\left<n_1, n_2\right>, t|R)$ is the number of times the \emph{particular} seed name pair $\left<n_1, n_2\right>$ matched any \np anchor pair of the parsed sentences of $R$ and contributed to the extraction of the \emph{particular} template $t$; and $N,T$ are again the total numbers of seed name pairs and templates, respectively, of $R$. 
Again, when scoring a sentence plan $p$ for a relation $R$, we calculate the $\mathit{productivity}(\left<n_1, n_2\right>, t|R)$ of 
\emph{all} the combinations of seed name pairs $\left<n_1, n_2\right>$ and templates $t$ that led to $p$, and we use the \emph{maximum}, \emph{minimum}, \emph{average}, \emph{total}, and \emph{standard deviation} of these
scores 
as features of $p$. We define very similarly $\mathit{productivity}(n_1, t|R)$, $\mathit{productivity}(n_2, t|R)$, $\mathit{productivity}(\left<a_1, a_2\right>, t|R)$, $\mathit{productivity}(a_1, t|R)$, $\mathit{productivity}(a_2, t|R)$, $\mathit{productivity}($ 
$\left<n_1, n_2\right>, \left<a_1, a_2\right>|R)$, $\mathit{productivity}(n_1, a_1|R)$, $\mathit{productivity}(n_2, a_2|R)$, obtaining five additional features of $p$ from each one. We also define:
\begin{equation}\label{productivityTemplateSeedNameAnchorFunction}
\begin{split}
\mathit{productivity}(\left<n_1, n_2\right>, \left<a_1, a_2\right>, t|R) = 
\frac{\mathit{hits}(\left<n_1, n_2\right>, \left<a_1, a_2\right>, t|R)}
{\sum_{i=1}^{N}{\sum_{j=1}^{A}{\sum_{k=1}^{T}{\mathit{hits}(\left<n_{i,1}, n_{i,2}\right>, \left<a_{j,1}, a_{j,2}\right>, t_k|R)}}}}
\end{split}
\end{equation}
where: $\mathit{hits}(\left<n_1, n_2\right>, \left<a_1, a_2\right>, t|R)$ is the number of times 
$\left<n_1, n_2\right>$ matched the \np anchor pair $\left<a_1, a_2\right>$ in a parsed sentence of $R$ 
contributing to the extraction of
template $t$; 
$N, A, T$ are the
numbers of seed name pairs, \np anchor pairs, 
templates of $R$. We define 
similarly $\mathit{productivity}(n_1, a_1, t|R)$ and 
$\mathit{productivity}(n_2, a_2, t|R)$, obtaining five features 
from each
type of productivity score.

\subsubsection{Prominence features} 

For each 
\textit{productivity} version of Section~\ref{productivityFeatures}, we define a 
\emph{prominence} variant. For example, based on the productivity of a seed name pair $\left<n_{i,1}, n_{i,2}\right>$ of a relation $R$ (Eq.~\ref{productivityParentSeedNamePairFunction}, repeated as Eq.~\ref{productivityParentSeedNamePairFunctionCopy}),  
\begin{equation}\label{productivityParentSeedNamePairFunctionCopy}
\mathit{productivity}(\left<n_{i,1}, n_{i,2}\right>|R) = 
\frac{\mathit{hits}(\left<n_{i,1}, n_{i,2}\right>|R)}
{\sum_{j=1}^{N}{\mathit{hits}(\left<n_{j,1}, n_{j,2}\right>|R)}}
\end{equation}
we define the \emph{prominence of a candidate sentence plan} $p$ with respect to the \emph{seed name pairs} of $R$:
\begin{equation}
\mathit{prominence}_{\textit{seed\_name\_pairs}}(p|R) = 
\frac{ \sum_{i=1}^{N} 1\{ \mathit{hits}(\left<n_{i,1}, n_{i,2}\right>|p,R) > 0 \} }
{\sum_{j=1}^{N}{ 1\{ \mathit{hits}(\left<n_{j,1}, n_{j,2}\right>|R) >0 \}}}
\end{equation}
where: the 
$1\{\xi\}$ denotes $1$ if condition $\xi$ holds and $0$ otherwise; $\mathit{hits}(\left<n_{i,1}, n_{i,2}\right>|p, R)$ (in the numerator) is the number of times $\left<n_{i,1}, n_{i,2}\right>$ matched any \np anchor pair of the parsed sentences of $R$, counting only matches that contributed to the extraction of a template of $R$ that led to the \emph{particular sentence plan} $p$; by contrast, $\mathit{hits}(\left<n_{j,1}, n_{j,2}\right>|R)$ (in the denominator) is the number of times $\left<n_{j,1}, n_{j,2}\right>$ matched any \np anchor pair of the parsed sentences of $R$, counting only matches that contributed to the extraction of \emph{any template} of $R$; and $N$ is the total number of seed name pairs of $R$. In other words, we count how many (distinct) seed name pairs of $R$ produced $p$, dividing by the number of (distinct) seed name pairs of $R$ that produced at least one template of $R$. The intuition is that the more
seed name pairs of $R$ lead to the 
sentence plan $p$, the better $p$ is.

In a similar manner, we define 
$\mathit{prominence}_{\textit{anchor\_pairs}}(p|R)$ based on  Eq.~\ref{productivityParentAnchorPairFunction}, and similarly for all the other productivity versions of Section~\ref{productivityFeatures}. We obtain one feature for the candidate sentence plan $p$ from each prominence variant, i.e., we do not compute any maximum, minimum, average, sum, 
standard deviation values, unlike the productivity versions, which lead to five features each.

\subsubsection{pmi features}

To estimate the extent to which two seed names $n_1$, $n_2$ of a relation $R$ co-occur when they match \np anchors to produce templates of $R$, we use 
a Pointwise Mutual Information (\pmi) score:
\begin{equation}\label{connectivityPMIparentSeedNamePair}
\pmi(\left<n_1, n_2\right>|R) = 
\frac{1}{-\log\mathit{productivity}(\left<n_1, n_2\right>|R)} %
\log\frac{\mathit{productivity}(\left<n_1, n_2\right>|R)}
{\mathit{productivity}(n_1|R) \cdot \mathit{productivity}(n_2|R)}
\end{equation}

\noindent The second factor of the right-hand side of Eq.~\ref{connectivityPMIparentSeedNamePair} is the standard \pmi definition, using productivity scores instead of probabilities. The first factor normalizes the \pmi scores to $[-1,1]$ ($-1$ if $n_1, n_2$ never co-occur when producing templates of $R$, $0$ if they are independent, $1$ if they always co-occur). Intuitively, if $n_1, n_2$ co-occur frequently when they produce templates of $R$, they are strongly connected and, hence, they are more indicative of $R$. Again, when using the classifier to score a sentence plan $p$ for a relation $R$, we calculate $\pmi(\left<n_1, n_2\right>|R)$ for 
\emph{all} the seed name pairs $\left<n_1, n_2\right>$ of $R$, and we use the \emph{maximum}, \emph{minimum}, \emph{average}, \emph{total}, and \emph{standard deviation} of these \pmi scores as five more features of $p$. We define similarly $\pmi(\left<n_1, n_2\right>, t|R)$, $\pmi(n_1, t|R)$, $\pmi(n_2, t|R)$, $\pmi(\left<a_1, a_2\right>|R)$, $\pmi(\left<a_1, a_2\right>, t|R)$, $\pmi(a_1, t|R)$, $\pmi(a_2, t|R)$,  $\pmi(\left<n_1, n_2\right>, \left<a_1, a_2\right>|R)$,  $\pmi(n_1, a_1|R)$, $\pmi(n_2, a_2|R)$, 
obtaining five features for $p$ from each one. 

\subsubsection{Token-based features}

These features view seed names, \np anchors, templates, and \owl identifiers
as sequences of tokens. 

For each seed name $n_1$ and \np anchor $a_1$ that matched (as first elements of a seed name pair $\left<n_1, n_{i,2}\right>$ and \np anchor pair $\left<a_1, a_{j,2}\right>$) to produce a particular sentence plan $p$, we calculate their \emph{cosine similarity} $\cos(n_1, a_1)$ with $\textit{tf}\cdot\textit{idf}$ weights (defined as in Section~\ref{PatternExtraction}).\footnote{When computing the features of this section, all tokens are stemmed, stop-words are ignored, accents are removed, and numeric tokens are replaced by a particular pseudo-token, as in Section~\ref{PatternExtraction}.} We then use the \emph{maximum}, \emph{minimum}, \emph{average}, \emph{total}, and \emph{standard deviation} of these cosine similarities as features of $p$. Intuitively, they show how good the matches that produced $p$ were. We repeat for each seed name $n_2$ and \np anchor $a_2$ that matched (as second elements of their pairs) to produce $p$, this time computing  $\cos(n_2, a_2)$, obtaining five additional features of $p$. 

We do the same using $\avgpmi(n_1, a_1)$,
defined below, instead of $\cos(n_1, a_1)$:
\begin{equation}
\avgpmi(n_1, a_1|R) = 
\frac{1}{|n_1| \cdot |a_1|}
\sum_{\tau \in \mathit{toks}(n_1)}
\sum_{\tau' \in \mathit{toks}(a_1)}
\frac{1}{-\log P(\tau, \tau'|R)} \cdot
\log\frac{P(\tau, \tau'|R)}
{P(\tau|R) \cdot P(\tau'|R)}
\end{equation}

\noindent where: $|n_1|$, $|a_1|$ are the lengths (in tokens) of $n_1, a_1$; $\mathit{toks}(n_1)$, $\mathit{toks}(a_1)$ are the token sequences of $n_1, a_1$, respectively; $P(\tau|R)$ is the probability of encountering 
token $\tau$ in a parsed sentence of $R$; and $P(\tau, \tau'|R)$ is the probability of encountering both $\tau$ and $\tau'$ in the same parsed sentence of $R$; we use Laplace estimates for these probabilities. Again, we compute $\avgpmi(n_1, a_1)$ for every seed name $n_1$ and \np anchor $a_1$ that matched to produce a particular sentence plan $p$, and we use the \emph{maximum}, \emph{minimum}, \emph{average}, \emph{total}, and \emph{standard deviation} of these scores as features of $p$. We repeat using $\avgpmi(n_2, a_2)$ instead of $\cos(n_2, a_2)$, obtaining five more features. 

Similarly, we compute $\avgpmi(a_1, t|R)$ and $\avgpmi(a_2, t|R)$ for each \np anchor $a_1$ or $a_2$ (left, or right element of an \np anchor pair) and template $t$ (ignoring the $S$ and $O$) 
that led to a particular sentence plan $p$, and we use their \emph{maximum}, \emph{minimum}, \emph{average}, \emph{total}, and \emph{standard deviation} as ten additional features of $p$. We also compute $\cos(t, r)$ and $\avgpmi(t, r|R)$ for each template $t$ and tokenized identifier $r$ of a relation $R$ (e.g., $R=$ \code{:madeFrom} becomes $r =$ ``made from'') that led to the sentence plan $p$, obtaining ten more features. Finally, we compute $\avgpmi(a_1, a_2|R)$, $\avgpmi(a_1, a_1|R)$, and $\avgpmi(a_2, a_2|R)$  for all the $a_1$ and $a_2$ \np anchors (first or second elements in their pairs) of $p$, obtaining fifteen more features of $p$. Although they may look strange, in effect $\avgpmi(a_1, a_1|R)$ and $\avgpmi(a_2, a_2|R)$ examine how strongly connected the words inside each \np anchor ($a_1$ or $a_2$) are.

\subsubsection{Other features}

Another group of features try to estimate the grammaticality of a candidate sentence plan $p$. Let us assume that $p$ is
for relation $R$. For every seed name pair of $R$ (not only seed name pairs that led to $p$), we generate a sentence using $p$; we ignore only seed name pairs that produced no sentence plans at all, which are assumed to be poor. For example, for the seed name pair $<$$n_1 =$``Semillon'', $n_2 =$ ``Semillon grape''$>$ of the relation $R =$ \code{:madeFrom} and the following candidate sentence plan:

\begin{center}
[$\mathit{ref}(S)$]$^{1}_{\textit{nom}}$
[make]$^{2}_{\textit{verb}, \, \textit{passive}, \, \textit{present}, \, \textit{agr}=1, \, \textit{polarity}=+}$ 
[from]$^{3}_{prep}$ 
[$\mathit{ref}(O)$]$^{4}_{\textit{acc}}$
\end{center}

 \noindent the sentence ``Semillon is made from Semillon grapes'' is generated. We do not generate referring expressions, even when required by the sentence plan (e.g., [$\mathit{ref}(S)$]$^{1}_{\textit{nom}}$); we use the seed names instead. We obtain confidence scores for these sentences from the parser, and we normalize these scores dividing by each sentence's length. The \emph{maximum}, \emph{minimum}, \emph{average}, 
 and \emph{standard deviation} of these scores are used as features of $p$. 

Some additional features for a candidate sentence plan $p$ follow:\footnote{All the non-Boolean features that we use are normalized to $[0, 1]$.}
\begin{itemize}
	\item True if $p$ contains a 
	present participle without an auxiliary; otherwise false. 
	\item True if $p$ has a main verb in active voice; otherwise false. %
	\item True if $p$ contains a referring expression for $S$ before a referring expression for $O$; otherwise false. Sentence plans that refer to $S$ before $O$ are usually simpler and better. %
	\item True if a referring expression of $p$ is the \emph{subject} of a verb of $p$; otherwise false. This information is obtained from the parsed sentences that led to $p$. We use the most frequent dependency tree, if $p$ was derived from many sentences. Sentence plans with no subjects are often ill-formed.
	\item True if a referring expression of $p$ is the \emph{object} of a verb of $p$; otherwise false. Again, 
	we consider the parsed sentences that led to $p$
	using the most frequent dependency tree. Sentence plans with no objects are often ill-formed, because most relations are expressed by transitive verbs.
	\item True if all the sentences 
	$p$ was derived from were 
	well-formed, according to the parser.
     \item True if $p$ required a repair at the end of the sentence plan generation (Section~\ref{SentencePlanGeneration}); otherwise false. Repaired sentence plans can be poor. %
     	\item The number of slots of $p$, the number of slots before the slot for $S$, the number of slots after the slot for $O$, the number of slots between the slots for $S$ and $O$ (4 features). 
	\item The \emph{maximum}, \emph{minimum}, \emph{average}, \emph{total}, 
	\emph{standard deviation} of the ranks of the Web pages 
	(returned by the search engine, Section~\ref{PatternExtraction}) that contained the sentences $p$ was obtained from. Sentences from higher-ranked 
	pages are usually more relevant to the seed name pairs we use as queries. Hence, sentence plans obtained from higher-ranked Web pages are usually better.
	\item The number of Web pages that contained the sentences from which $p$ was obtained. Uncommon sentences often lead to poor sentence plans.
\end{itemize}

\subsection{Ranking the candidate sentence plans} \label{rankingSPs}

Each candidate sentence plan of a relation $R$ is represented as a feature vector $\vec{v}$, containing the 251 features discussed above.
Each vector is given to the MaxEnt classifier to obtain a probability estimate $P(c_+|\vec{v}, R)$ that it belongs in the positive class $c_+$, i.e., that 
the sentence plan is correct for $R$. The candidate sentence plans of each relation $R$ are then ranked by decreasing estimated (by the classifier) $P(c_+|\vec{v}, R)$. We call \spg our overall sentence plan generation method that uses the probability estimates of the classifier to rank the candidate sentence plans. 

In an alternative configuration of our sentence plan generation method, denoted \spgrerank, the probability estimate $P(c_+|\vec{v}, R)$ of each candidate sentence plan is multiplied by its \emph{coverage}. To compute the coverage of a sentence plan for a relation $R$, we use the sentence plan to produce a sentence for each seed name pair of $R$ (as when computing the grammaticality of a sentence plan in Section~\ref{EvaluatingSentencePlans}). Subsequently, we use each sentence as a phrase query in a Web search engine. The coverage of the sentence plan is the number of seed name pairs for which the search engine retrieved at least one document containing the search sentence (verbatim), divided by the total number of seed name pairs of $R$. Coverage 
helps avoid sentence plans that produce very uncommon sentences.
Computing the coverage of \emph{every} candidate sentence plan is time consuming, however, because of the Web searches; this is also why we do not include coverage in the features of the classifier. Hence, we first rank the candidate sentence plans of each relation $R$ by decreasing $P(c_+|\vec{v}, R)$, and we then re-rank only the top ten  
of them (per $R$) after multiplying the $P(c_+|\vec{v}, R)$ of each one by its coverage. 

In both \spg and \spgrerank, in a semi-automatic scenario we return to a human inspector the top five candidate sentence plans 
per relation. In a fully automatic scenario, we return only the top one. 

\section{Experiments} \label{Experiments}

We now present the experiments we performed to evaluate our methods that generate \nl names and sentence plans. We first discuss the ontologies that we used in our experiments.

\subsection{The ontologies of our experiments} \label{ontologiesOfExperiments}

We used three ontologies:
(i) the Wine Ontology, one of the most commonly used examples of \owl ontologies;
(ii) the \mpiro ontology, which describes a collection of museum exhibits,
was originally developed in the \mpiro project \cite{Isard2003}, was later ported to \owl, and accompanies \nlowl \cite{Androutsopoulos2013}; and (iii) the Disease Ontology, which describes diseases, including their symptoms, causes etc.\footnote{See also \url{http://www.w3.org/TR/owl-guide/wine.rdf/} and  \url{http://disease-ontology.org/}.}  

The Wine Ontology involves a wide variety of \owl constructs and, hence, is a good test case for ontology verbalizers and \nlg systems
for \owl. The \mpiro ontology has been used to demonstrate the high quality texts that \nlowl can produce, when appropriate manually authored linguistic resources are provided \cite{Galanis2009}. We wanted to investigate if texts of similar quality can be generated with 
automatically or semi-automatically acquired \nl names and sentence plans. The Disease Ontology 
was developed by biomedical experts to address real-life information needs; hence, it constitutes a good real-world test case.

The Wine Ontology contains 77 classes, 161 individuals, and 14 relations (properties). 
We aimed to produce \nl names and sentence plans for the 49 classes, 146 individuals, and 7 relations that are directly involved in non-trivial definitions 
of wines (43 definitions of wine classes, 52 definitions of wine individuals), 
excluding classes, individuals, and relations that are 
only used to define 
wineries, wine-producing regions etc. By ``non-trivial definitions'' we mean that we ignored 
definitions that humans understand as repeating information that is obvious from the name of the defined class or individual (e.g., the definition of 
\code{:RedWine} in effect says that a red wine is a wine with red color). 

The 
\mpiro ontology 
currently contains 76 classes, 508 individuals, and 41 relations. Many individuals, however, are used to represent canned texts (e.g., manually written descriptions of particular types of exhibits) that are difficult to generate from symbolic information. For example, there is a pseudo-individual \code{:aryballos-def} whose \nl name is the fixed string ``An aryballos was a small spherical vase with a narrow neck, in which the athletes kept the oil they spread their bodies with''. Several properties are also used only to link these pseudo-individuals (in effect, the canned texts) to other individuals or classes
(e.g., to link \code{:aryballos-def} to the class \code{:Aryballos}); and many other classes are used only to group pseudo-individuals (e.g., pseudo-individuals whose canned texts describe types of vessels all belong in a common class). In our experiments, we ignored pseudo-individuals, properties, and classes that are used to represent, link, and group canned texts, since we focus on generating texts from symbolic information. We aimed to produce \nl names and sentence plans for the remaining 30 classes, 127 individuals, and 12 relations, which are all involved in the definitions (descriptions) of the 49 exhibits of the collection the ontology is about. 

The Disease Ontology currently contains information about 6,286 diseases, all represented as classes. 
Apart from \textsc{is-a} 
relations, synonyms, and pointers to related terms, however, all the other information 
is represented using strings containing quasi-English sentences with relation names used mostly as verbs. For example, there is an axiom in the ontology stating that the Rift Valley Fever (\textsc{doid}\_1328) is a kind of viral infectious disease (\textsc{doid}\_934). 
All the other information
about the Rift Valley Fever is provided in a string, shown below as `Definition'. The tokens that contain underscores (e.g., \code{results\_in}) are relation names. The ontology declares all the relation names, but uses them only inside `Definition' strings. Apart from diseases, it does not define any of the other entities mentioned in the `Definition' strings (e.g., symptoms, viruses).

\begin{quoting}
{\small \noindent Name: Rift Valley Fever (\textsc{doid}\_1328)\\
\textsc{is-a}: viral infectious disease (\textsc{doid}\_934) \\
Definition: A viral infectious disease that \code{results\_in} infection, \code{has\_material\_basis\_in} Rift Valley fever virus, which is \code{transmitted\_by} Aedes mosquitoes. The virus affects domestic animals (cattle, buffalo, sheep, goats, and camels) and humans. The infection \code{has\_symptom} jaundice, \code{has\_symptom} vomiting blood, \code{has\_symptom} passing blood in the feces, \code{has\_symptom} ecchymoses (caused by bleeding in the skin), \code{has\_symptom} bleeding from the nose or gums, \code{has\_symptom} menorrhagia and \code{has\_symptom} bleeding from venepuncture sites. 
}
\end{quoting}

We defined as individuals 
all the non-disease entities mentioned in the `Definition' strings, 
also adding statements to formally express the relations mentioned in the original `Definition' strings. For example, the resulting 
ontology contains the following definition of Rift Valley Fever, where \code{:infection}, \code{:Rift\_Valley\_fever\_virus}, \code{:Aedes\_mosquitoes}, \code{:jaundice} etc.\ are new individuals. 
\label{DOID1328}
{\small
\begin{verbatim}
   SubClassOf(:DOID_1328 
      ObjectIntersectionOf(:DOID_934
         ObjectHasValue(:results_in :infection)
         ObjectHasValue(:has_material_basis_in :Rift_Valley_fever_virus)
         ObjectHasValue(:transmitted_by :Aedes_mosquitoes)
         ObjectHasValue(:has_symptom :jaundice)
         ObjectHasValue(:has_symptom :vomiting_blood)
         ObjectHasValue(:has_symptom :passing_blood_in_the_feces)
         ObjectHasValue(:has_symptom :ecchymoses_(caused_by_bleeding_in_the_skin))
         ObjectHasValue(:has_symptom :bleeding_from_the_nose_or_gums)
         ObjectHasValue(:has_symptom :menorrhagia)
         ObjectHasValue(:has_symptom :bleeding_from_venepuncture_sites)))
\end{verbatim}
}

\noindent The new 
form of the ontology was produced automatically, using patterns that searched the definition strings for 
relation names (e.g., \code{results\_in}), sentence breaks, and words that introduce secondary clauses (e.g., ``that'', ``which'').\footnote{The 
new form of the Disease Ontology that we produced is available upon request.} Some sentences of the original definition strings that did not include declared relation names, like the sentence  ``The virus affects\dots and humans'' in the `Definition' string of Rift Valley Fever 
above, were discarded during the conversion, because it was not always possible to reliably convert them to appropriate \owl statements.

The new 
form of the Disease Ontology contains 6,746 classes, 15 relations, and 1,545 individuals. 
We aimed to automatically produce \nl names and sentence plans for the 94 classes, 99 individuals, and 8 relations 
that are involved in the definitions of 30 randomly selected diseases. We 
manually authored \nl names and sentence plans for the same classes, individuals, and relations, to be able to compare the quality of the resulting texts. Manually authored \nl names and sentence plans for the Wine and \mpiro ontologies are also available (they are included in the software of \nlowl). 

We note that the relations  (properties) $R$ of our experiments are all used in message triples $\left<S, R, O\right>$, where $O$ is an individual or class, i.e., they are \emph{object properties} in \owl's terminology. Datatype properties, where $O$ is a datatype value (e.g., integer, string, date), can in principle be handled using the same methods, but appropriate recognizers may be needed to obtain appropriate anchors, instead of \np anchors. For example, a datatype property may map persons to dates of birth; then a recognizer of dates would be needed to extract (and possibly normalize) appropriate date anchors from Web pages, since a parser may not treat dates as \np{s}.

\subsection{Experiments with  
automatically or semi-automatically produced NL names} \label{NLNameExperiments}

We now present 
experiments we performed with our method that generates \nl names (Section~\ref{OurMethodNLN}). 

\subsubsection{Anonymity experiments}

In a first experiment, we measured how well our \nl names method determines which individuals and classes should be anonymous
(Sections~\ref{NLnamesBackground} and \ref{shorteningIdentifiers}). We compared the decisions of our method against the corresponding anonymity declarations in the manually authored \nl names of the three ontologies. 
Table~\ref{anonymousTable} summarizes the results of this experiment. \emph{Precision} is the total number of individuals and classes our \nl names method \emph{correctly} (in agreement with the manually authored \nl names) declared as anonymous, divided by the total number of individuals and classes our method declared as anonymous. \emph{Recall} is the total number of individuals and classes our \nl names method correctly declared as anonymous, divided by the total number of individuals and classes (among those we aimed to produce \nl names for) that the manually authored \nl names declared as anonymous. For the Disease Ontology, the manually authored \nl names and our \nl names method agreed that no individuals and classes (that we aimed to produce \nl names for) should be anonymous, which is why precision and recall are undefined. \emph{Accuracy} is the number of correct decisions (individuals and classes correctly declared, or correctly not declared as anonymous), divided by the total number of individuals and classes (that we aimed to produce \nl names for). 

\begin{table}
\center
{\footnotesize
\begin{tabular}{|l|c|c|c|}
\hline               & \textsc{wine} & \mpiro & \textsc{disease} \\
\hline 
\hline precision     & $1.00$ (38/38) & $1.00$ (49/49) & undefined\ (0/0)\\
\hline recall        & $0.73$ (38/52) & $1.00$ (49/49) & undefined (0/0)\\
\hline accuracy      & $0.93$ (184/198) & $1.00$ (157/157) & 1.00 (195/195)\\
\hline
\end{tabular}
}
\caption{Results of the anonymity experiments.}
\label{anonymousTable}
\end{table}

The anonymity decisions of our method were perfect in the \mpiro ontology and Disease Ontology. In the Wine Ontology, the precision of our method was also perfect, i.e., whenever our method decided to declare an individual or class as anonymous, this was a correct decision; but recall was lower, i.e., our method did not anonymize all the individual and classes that the manually authored \nl names did. The latter is due to the fact that the manually authored \nl names of the Wine ontology also anonymize 14 individuals and classes with complex identifiers (e.g., \code{:SchlossVolradTrochenbierenausleseRiesling}) to produce more readable texts. By contrast, our method declares individuals and classes as anonymous only to avoid redundancy in the generated texts (Section~\ref{shorteningIdentifiers}), hence it does not anonymize the 14 individuals and classes.

\subsubsection{Inspecting the produced natural language names} \label{inspectingNLnames}

We then invoked our
\nl name generation method for the individuals and classes
it had not declared as anonymous (160 in the Wine Ontology, 108 in the \mpiro ontology, 195 in the Disease Ontology), 
using the top 10 returned documents per Web search (or 
top 20, when the search engine proposed spelling 
corrections -- see Section~\ref{rankingNPs}). We ranked the produced \nl names (as in Sections~\ref{rankingNPs} and 
\ref{NLNameGeneration}), and kept the top 5 \nl names per individual or class. The first author 
then inspected the resulting \nl names and marked each one as correct or incorrect. An \nl name was considered correct if and only if: (i) it would produce morphologically, syntactically, and semantically correct and unambiguous noun phrases 
(e.g., ``Cabernet Sauvignon grape'' is correct for \code{:CabernetSauvignonGrape}, but ``Cabernet Sauvignon wine'', ``Cabernet Sauvignon'', or ``grape'' are incorrect); and (ii)  
its slot annotations 
(e.g., \pos tags, gender, agreement) were all correct.

Table \ref{ontologyMRR_NLN_Table} shows the results of this experiment. The ``\emph{1-in-1}'' score is the ratio of individuals and classes for which the top returned \nl name was correct. The ``\emph{1-in-3}'' score is the ratio of individuals and classes for which there was at least one correct \nl name among the top three, and similarly for ``\emph{1-in-5}''. The ``1-in-1'' score corresponds to a fully automatic scenario, where the top \nl name 
is used for each individual or class, without  human intervention. By contrast, the ``1-in-3'' and ``1-in-5'' scores correspond to a semi-automatic scenario, where a human 
inspects the top three or five, respectively, \nl names per individual or class, looking for a correct one to select. The \emph{mean reciprocal rank} (\textsc{mrr}) is the mean (over all the individuals and classes we asked our method to produce \nl names for) of the reciprocal rank $r_i = 1/k_i$, where $k_i$ is the rank (1 to 5) of the top-most correct \nl name returned for the $i$-th individual or class; if no correct \nl name exists among the top five, then $r_i = 0$. 
\textsc{mrr} rewards more those methods that place correct \nl names towards the top of the five returned 
ones. The \emph{weighted} scores of Table~\ref{ontologyMRR_NLN_Table} are similar, but they weigh
each individual or class 
by the number of \owl statements that mention it in the ontology.

\begin{table}
\center
{\footnotesize
\begin{tabular}{|c|c|c|c|}
\hline               & \textsc{wine} 					& \mpiro							& \textsc{disease} \\
\hline 
\hline 1-in-1        & $0.69$ (110/160)				& $0.77$ (83/108)			& $0.74$ (145/195)\\
\hline 1-in-3        & $0.93$ (145/160)				& $0.94$ (102/108)		& $0.89$ (173/195)\\
\hline 1-in-5        & $0.95$ (152/160)  			& $0.95$ (103/108)		& $0.90$ (176/195)\\
\hline MRR     			 & $0.79$ 								& $0.85$ 							& $0.80$ \\
\hline 
\hline weighted 1-in-1        & $0.69$ 				& $0.77$		& $0.76$\\
\hline weighted 1-in-3        & $0.93$				& $0.94$		& $0.91$\\
\hline weighted 1-in-5        & $0.96$   				& $0.95$		& $0.93$\\
\hline weighted MRR     	  & $0.80$ 				& $0.85$ 		& $0.82$ \\
\hline
\end{tabular}
}
\caption{Results of the inspection of the produced \nl names.}
\label{ontologyMRR_NLN_Table}
\end{table}

The results of Table~\ref{ontologyMRR_NLN_Table} 
show that our \nl names method performs very well in a semi-automatic scenario.
In a fully automatic scenario, however, 
there is
large scope for improvement. We note, 
though, that our definition of correctness (of \nl names) in the experiment of this section was 
very strict. For example, an \nl name with only a single error in its slot annotations (e.g., a wrong gender in a noun slot) was counted as incorrect, even if in practice the error might have a minimal effect on the generated texts that would use the \nl name. The experiment of 
Section~\ref{evaluatingNLNtexts} below, where \nl names are considered in the context of generated texts, sheds more light on this point.

By inspecting the produced \nl names, we noticed that our method is very resilient to spelling errors and abbreviations in the \owl identifiers of individuals and classes. For example, it returns \nl names producing ``a Côte d'Or wine'' for \code{:CotesDOr}, and ``the Naples National Archaeological Museum'' for \code{:national-arch-napoli}. Several wrongly produced \nl names are due to errors of 
tools that our method invokes  (e.g., parser, 
\textsc{ner}). 
Other errors 
are due to over-shortened \code{altTokNames} (Section~\ref{NounPhraseExtraction}); 
e.g., one of the \code{altTokNames} of \code{:CorbansDryWhiteRiesling} was simply ``Dry White'', which leads to an \nl name that does not identify the particular wine clearly enough. 
Finally, in the Disease Ontology, our automatic conversion 
of the `Description' strings produced many individuals whose identifiers are in effect long phrases (see, for example, the \owl description of \code{:\textsc{doid}\_1328} in Section~\ref{ontologiesOfExperiments}). Our \nl names method manages to produce appropriate \nl names (with correct slot annotations etc.) for some of them (e.g., \code{:mutation\_in\_the\_SLC26A2\_gene}), but produces no \nl names in other cases (e.g., \code{:infection\_of\_the\_keratinized\_layers}). Some of these errors, however, may not have a significant effect on the generated texts (e.g., using the tokenized identifer ``infection of the keratinized layers'', which is the default when no \nl name is provided, may still lead to a reasonable text). Again, the experiment of
Section~\ref{evaluatingNLNtexts} below sheds more light on this point.

\subsubsection{Annotator agreement and effort to semi-automatically author nl names}
\label{agreementNLnames}
 
The top five automatically produced \nl names of each individual and class were also shown to a second human judge. The second judge was a computer science researcher not involved in \nlg, fluent in English, though not a native speaker. For each individual or class, and for each one of its top five \nl names, the judge was shown a phrase produced by the \nl name (e.g., ``Cabernet Sauvignon''), an automatically generated sentence about the individual or class 
(expressing a message triple of the ontology) illustrating the use of the \nl name (e.g., ``Cabernet Sauvignon is a kind of wine.''), and 
a sentence where the \nl name had been automatically replaced by a pronoun (e.g., ``It is a kind of wine.'') to check the gender of the \nl name. The judge was asked to consider the phrases and sentences, and mark the best correct \nl name for each individual or class. The judge could also mark more than one \nl names for the same individual or class, if more than one seemed correct and equally good; the judge was instructed not to mark any of the five \nl names, if none seemed correct.
The judge completed this task in 49, 45, and 75 minutes for the Wine, 
\mpiro, and Disease Ontology (727, 540, 965 candidate \nl names), respectively;
by contrast, manually authoring the \nl names of the three ontologies took approximately 2, 2, and 3 working days, respectively.
These times and the fact that the second judge was not 
aware of the internals of \nlowl
and its resources suggest that the semi-automatic authoring scenario is 
viable and very useful in practice.

Table~\ref{ontologyAgreement_NLN_Table} compares the decisions of the second judge, hereafter called \judgetwo, to those of the first author, hereafter called \judgeone. \judgeone was able to view the full details of the \nl names using \nlowl's \protege plug-in, unlike \judgetwo who viewed only phrases and example sentences. For the purposes of this study, \judgeone marked \emph{all} the correct \nl names (not only the best ones) among the top five of each individual or class. In Table~\ref{ontologyAgreement_NLN_Table}, \emph{micro-precision} is the number of \nl names (across all the individuals and classes) that were marked as correct by both \judgeone and \judgetwo, divided by the number of \nl names marked as correct by \judgetwo, i.e., we treat the decisions of \judgeone as gold. \emph{Macro-precision} is similar, but we first compute the precision of \judgetwo against \judgeone separately for each individual or class, and we then average over all the individuals and classes. \judgeone \emph{1-in-5} is the percentage of individuals and classes for which \judgeone marked at least one \nl name among the top five as correct,
and similarly for \judgetwo \emph{1-in-5}. 
\emph{Pseudo-recall} is the number of individuals and classes for which both \judgeone and \judgetwo marked at least one \nl name as correct, divided by the number of individuals and classes for which \judgeone marked at least one \nl name as correct; this measure shows how frequently \judgetwo managed to find at least one correct (according to \judgeone) \nl name, when there was at least one correct \nl name among the top five. Computing the true recall of the decisions of \judgetwo against those of \judgeone would be inappropriate, because \judgetwo was instructed to mark only the best \nl name(s) of each individual and class, unlike \judgeone who was instructed to mark all the correct ones.
We also calculated Cohen's Kappa between \judgeone and \judgetwo; for each individual or class, if \judgeone had marked more than one \nl names as correct, we kept only the 
top-most one, and similarly for \judgetwo, hence each judge had six possible choices (including marking no \nl name) per individual and class. The results of Table~\ref{ontologyAgreement_NLN_Table} indicate strong inter-annotator agreement in the semi-automatic authoring of \nl names in all three ontologies. 

\begin{table}
\center
{\footnotesize
\begin{tabular}{|c|c|c|c|}
\hline               & \textsc{wine}	& \mpiro	& \textsc{disease} \\
\hline 
\hline micro-precision 	& $0.97$ 		& $0.96$ 		& $0.97$ \\
\hline macro-precision  & $0.98$ 		& $0.94$ 		& $0.98$ \\
\hline 
\hline \judgeone 1-in-5 			& $0.95$ 		& $0.95$ 		& $0.90$ \\
\hline \judgetwo 1-in-5 		& $0.95$ 		& $0.93$ 		& $0.90$ \\

\hline pseudo-recall   				& $1.00$  	& $0.96$ 		& $1.00$ \\
\hline 
\hline Cohen's Kappa		& $0.77$ 		& $0.80$		& $0.98$\\
\hline
\end{tabular}
}
\caption{Inter-annotator agreement in the semi-automatic authoring of \nl names.}
\label{ontologyAgreement_NLN_Table}
\end{table}

\subsubsection{Evaluating natural language names in generated texts} \label{evaluatingNLNtexts}

In order to examine how the produced \nl names affect the perceived quality of the generated texts, we showed automatically generated texts describing individuals and classes of the three ontologies to six computer science students  not involved in the work of this article; they were all fluent, though not native, English speakers. We generated texts using \nlowl configured in four ways. The \nonl configuration uses no \nl names; in this case, \nlowl uses 
the tokenized \owl identifiers of the individuals and classes as their names.\footnote{If the ontology provides an \code{rdfs:label} for an individual or class, \nonl uses 
its tokens.} \authnl uses manually authored \nl names. \autonl uses the top-ranked \nl name that our \nl names method produces for each individual and class. Finally, \semiautonl uses the \nl name (of each individual or class) that a human inspector (the first author of this article)  selected among the top five \nl names produced by our method. Additionally, both \autonl and \semiautonl use the methods of Sections~\ref{shorteningIdentifiers} and \ref{InterestReasoning} to anonymize individuals or classes and to infer interest scores from \nl names, whereas \authnl uses the anonymity declarations and interest scores of the manually authored linguistic resources, and \nonl uses no anonymity declarations and no interest scores. Apart from the \nl names, anonymity declarations, and interest scores, all four configurations use the same, manually authored other types of  linguistic resources (e.g., sentence plans, text plans to order the message triples). Below are example texts generated from the three ontologies by the four configurations.

\medskip
{\small
\noindent
\authnl: This is a moderate, dry Zinfandel. It has a medium body. It is made by Saucelito Canyon in the city of Arroyo Grande.\\
\semiautonl: This is a moderate, dry Zinfandel wine. It has a medium body. It is made by the Saucelito Canyon Winery in the Arroyo Grande area. \\
\autonl: This is a dry Zinfandel and has the medium body. It is the moderate. It is made by Saucelito Canyon in Arroyo Grande.\\
\nonl: Saucelito Canyon Zinfandel is Zinfandel. It is Dry. It has a Medium body. It is Moderate. It is made by Saucelito Canyon. It is made in Arroyo Grande Region. 
}

\medskip
{\small
\noindent
\authnl: This is a statue, created during the classical period and sculpted by Polykleitos. Currently it is exhibited in the National Archaeological Museum of Napoli.\\
\semiautonl: This is a statue, created during the classical period and sculpted by the sculptor polyclitus. Currently it is exhibited in the Naples National Archaeological Museum. \\
\autonl: This is a statue, created during classical and sculpted by the polyclitus. Currently it is exhibited in national arch napoli.\\
\nonl: Exhibit 4 is statue, created during classical period and sculpted by polyclitus. Today it is exhibited in national arch napoli. 
}

\medskip
{\small
\noindent
\authnl: Systemic mycosis is a kind of fungal infectious disease that affects the human body. It results in infection of internal organs. It is caused by fungi.\\
\semiautonl: A systemic mycosis is a kind of fungal infectious disease that affects human body. It results in infections of internal organs and it is caused by the fungi. \\
\autonl: A systemic mycosis is fungal that affects human body. It results in infections of internal organs and it is caused by the fungi.\\
\nonl: Systemic mycosis is a kind of fungal infectious disease. It affects human body. It results in infection of internal organs. It is caused by Fungi. 
}
\medskip

\noindent We note that some \nl names of \semiautonl and \autonl can be easily improved using the \protege plug-in of \nlowl. For example, the \nl name of the human body can be easily modified to include a definite article, which would improve the texts of \semiautonl and \autonl in the Disease Ontology examples above 
(``affects the human body'' instead of ``affects human body'').\footnote{
The missing article is due to the fact that the on-line dictionary we used marks ``body'' as (possibly) non-countable.} Nevertheless, we made no such improvements. 

Recall that there are $43 + 52 = 95$ non-trivial definitions of wine classes and wine individuals in the Wine Ontology, $49$ exhibits in the \mpiro ontology, and that we randomly selected $30$ diseases from the Disease Ontology (Section~\ref{ontologiesOfExperiments}). Hence, we generated $95 \times 4 = 380$ texts from the Wine Ontology (with the four configurations of \nlowl), $49 \times 4 = 196$ texts from the \mpiro ontology, and $30 \times 4 = 120$ texts from the Disease ontology. For each individual or class,
the message triples of its definition (regardless of their interest scores) along with the corresponding texts 
were given to exactly one student. The four texts of each individual or class were randomly ordered and the students did not know which configuration had generated each one of the four texts. For each individual or class, the students were asked to compare the four texts to each other and to the message triples, and score each text by stating how strongly they agreed or disagreed with statements $S_1$--$S_4$ below. A scale from 1 to 5 was used (1: strong disagreement, 3: ambivalent, 5: strong agreement). Examples and more detailed guidelines were also provided to the students.

\smallskip
($S_1$) \emph{Sentence fluency}: 
Each sentence of the text (on its own) is grammatical and sounds natural.

($S_2$) \emph{Clarity}: The text is easy to understand, provided that the reader is familiar with the terminology and concepts of the domain (e.g., historical periods, grape varieties, virus names).

($S_3$) \emph{Semantic correctness}: The text accurately conveys the information of the message triples.

($S_4$) \emph{Non-redundancy}: 
There is no redundancy in the text (e.g., stating the obvious,
repetitions).

\begin{table*}[ht]
\begin{center}
{\footnotesize
\begin{tabular}{|l||c|c|c|c|}
\hline Criteria              	& \nonl  					& \autonl  							& \semiautonl  						& \authnl \\
\hline 
\hline Sentence fluency      			& $3.99$								& $3.50$  							& $\mathbf{4.70}^{1}$ 		& $\mathbf{4.83}^{1}$\\
\hline Clarity        			 			& $4.41$ 								& $3.43$  							& $\mathbf{4.79}^{1}$ 		& $\mathbf{4.79}^{1}$\\
\hline Semantic correctness  				& $4.44^{1}$ 						& $3.54$  							& $\mathbf{4.66}^{1,2}$ 	& $\mathbf{4.85}^{2}$\\
\hline Non-redundancy  										& $3.17$ 								& $3.93^{1}$  					& $\mathbf{4.31}^{1,2}$ 	& $\mathbf{4.56}^{2}$\\
\hline
\end{tabular}
}
\caption{Human scores for the Wine Ontology with different methods to obtain \nl names.}
\label{nln_wine_human_results}
\end{center}
\vspace*{-7mm}
\end{table*}

\begin{table*}[ht]
\begin{center}
{\footnotesize
\begin{tabular}{|l||c|c|c|c|}
\hline Criteria              		& \nonl  						& \autonl 							& \semiautonl  						& \authnl \\
\hline 
\hline Sentence fluency      			& $4.14^{1}$		& $4.22^{1}$  					& $\mathbf{4.90}^{2}$ 	& $\mathbf{4.98}^{2}$\\
\hline Clarity        							& $4.00^{1}$ 		& $3.69^{1}$ 						& $\mathbf{4.82}^{2}$ 	& $\mathbf{4.98}^{2}$\\
\hline Semantic correctness  			& $4.06^{1}$ 		& $4.04^{1}$  					& $\mathbf{4.82}^{2}$ 	& $\mathbf{4.98}^{2}$\\
\hline Non-redundancy  										& $3.18$ 				& $4.06$  							& $\mathbf{4.86}^{1}$ 	& $\mathbf{4.96}^{1}$\\
\hline
\end{tabular}
}
\caption{Human scores for the \mpiro ontology with different methods to obtain \nl names.}
\label{nln_mpiro_human_results}
\end{center}
\vspace*{-7mm}
\end{table*}

\begin{table*}[ht]
\begin{center}
{\footnotesize
\begin{tabular}{|l||c|c|c|c|}
\hline Criteria              		& \nonl  						& \autonl 							& \semiautonl  						& \authnl \\
\hline 
\hline Sentence fluency      					& $\mathbf{4.27}^{1,2}$		& $4.03^{1}$  					& $\mathbf{4.40}^{1,2}$ 	& $\mathbf{4.73}^{2}$\\
\hline Clarity        						& $\mathbf{4.73}^{1}$ 		& $3.80$  							& $\mathbf{4.57}^{1}$ 		& $\mathbf{4.90}^{1}$\\
\hline Semantic correctness  				& $\mathbf{4.83}^{1}$ 		& $4.23^{2}$  					& $\mathbf{4.47}^{1,2}$ 	& $\mathbf{4.87}^{1}$\\
\hline Non-redundancy  									& $\mathbf{4.23}^{1}$ 		& $\mathbf{4.30}^{1}$  	& $\mathbf{4.43}^{1}$ 		& $\mathbf{4.33}^{1}$\\
\hline
\end{tabular}
}
\caption{Human scores for the Disease Ontology with different methods to obtain \nl names.}
\label{nln_disease_human_results}
\end{center}
\vspace*{-4mm}
\end{table*}

Tables \ref{nln_wine_human_results}--\ref{nln_disease_human_results} show the scores of the four configurations of \nlowl, averaged over the texts of each ontology. For each criterion, the best scores are shown in bold. 
In each criterion (row), we detected no statistically significant differences between scores marked with the same superscript; all the other differences (in the same row) were statistically significant.\footnote{
We performed Analysis of Variance (\textsc{anova}) and post-hoc Tukey tests 
($a = 0.05$) in the four scores of each criterion in each ontology.  A post-hoc power analysis of the \textsc{anova} values resulted in power values greater or equal to $0.98$, with the exception of the non-redundancy scores of the Disease Ontology,
where power was 0.25.}
Overall, the manually authored \nl names led to the best (near-perfect) scores, as one might expect. The scores of \semiautonl were overall slightly lower, but still high (always
$\geq 4.3/5$) and no statistically significant differences to the corresponding scores of \authnl were detected. These findings confirm 
that our \nl names method performs very well in a semi-automatic scenario, where a human inspects and selects among the top-ranked automatically produced \nl names. By contrast, \autonl performed overall much worse than \semiautonl and \authnl, and often worse than \nonl, which 
again indicates that our \nl names method cannot be used in a fully automatic manner. 

The \nonl configuration, which uses tokenized identifiers of individuals and classes, performed overall much worse than \authnl and \semiautonl in the Wine and \mpiro ontologies, which shows the importance of \nl names in the perceived quality of generated texts. The differences between \nonl, \semiautonl, and \authnl were smaller in the Disease Ontology, where no statistically significant differences between the three configurations were detected. These smaller differences are due to the fact that the conversion of the Disease Ontology 
(Section~\ref{ontologiesOfExperiments}) produced many individuals whose \owl identifiers are in effect long phrases,
easily readable, and sometimes better than then top-ranked 
\nl names of our methods; furthermore, our \nl names method does not manage to produce any \nl names for many of these individuals and, hence, \semiautonl ends up using their tokenized identifiers, like \nonl. We also note that there are very few redundant message triples and no anonymous individuals or classes in the Disease Ontology, which explains the higher non-redundancy scores of \nonl in the Disease Ontology, compared to the much lower non-redundancy scores of \nonl in the other two ontologies.

\subsection{Experiments with  
automatically or semi-automatically produced sentence plans} \label{SentencePlanExperiments}

We now present the experiments we performed to evaluate our method that generates sentence plans (Section~\ref{OurMethodSP}). Recall that our method employs a MaxEnt classifier to predict the probability that a candidate sentence plan is correct (positive class) or incorrect (negative class).\footnote{We also experimented with SVM{s} and different kernels, but 
saw no significant improvements compared to MaxEnt.}

\subsubsection{Training the classifier of the sentence plan generation method} \label{trainingClassifier}

To create training instances for the MaxEnt classifier, we used our sentence plan generation method without the classifier to obtain candidate sentence plans (as in Sections~\ref{PatternExtraction} and \ref{SentencePlanGeneration}) from Wikipedia for the seven relations of the Wine Ontology (Section~\ref{ontologiesOfExperiments}). We used the manually authored \nl names of the Wine Ontology to obtain seed names, and the top 50 Wikipedia articles of each search query.\footnote{We used Google's Custom Search API (\url{developers.google.com/custom-search/}) to search Wikipedia.} 
We searched Wikipedia exclusively at this stage, as opposed to querying the entire Web, to obtain high quality texts and, hence, hopefuly more positive training examples (correct sentence plans). The first author 
then manually tagged the resulting 655 candidate sentence plans as positive or negative training instances, depending on whether or not they were correct.
A candidate sentence plan was considered correct if and only if: (i) it would produce morphologically, syntactically, and semantically correct sentences; and (ii) the annotations of its slots (e.g., \pos tags, voice, tense, agreement) were all correct. To compensate for class imbalance in the training set (16\% positive vs.\ 84\% negative candidate sentence plans), we replicated all the positive training instances (over-sampling) to obtain an equal number of positive and negative training instances. 

\begin{figure}
\center
\includegraphics[width=0.5\columnwidth]{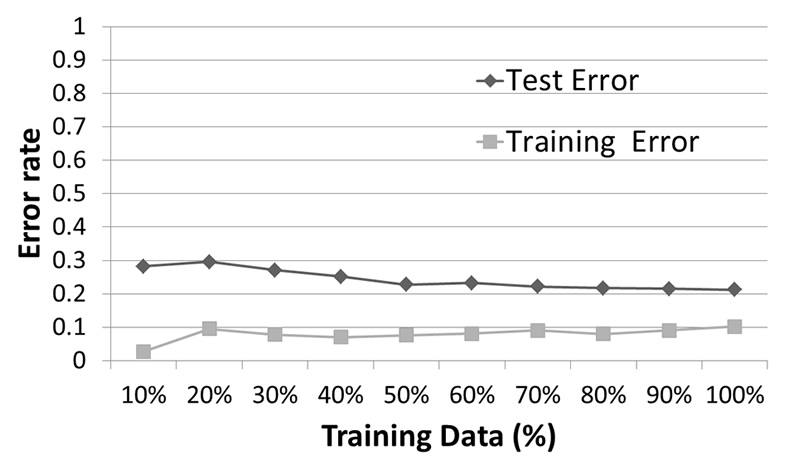}
\vspace*{-4mm}
\caption{Test and training error rate of the classifier of our sentence plan generation method.}
\label{TestTrainError}
\end{figure}

Figure~\ref{TestTrainError} shows the error rate of the classifier on (i) unseen instances (test error) and (ii) on the instances it has been trained on (training error). To obtain the curves of Fig.~\ref{TestTrainError}, we performed a leave-one-out cross validation on the 655 instances (candidate sentence plans) we had constructed, i.e., we repeated the experiment 655 times, each time using a different instance as the only test instance and the other 654 instances as the training dataset. Within each repetition of the cross-validation, we iteratively trained the classifier on 10\%, 20\%, \dots, 100\% of the training dataset 
(654 instances, with over-sampling applied to them).
The \emph{training error} counts how many of the instances that were used to train the classifier were also correctly classified by the classifier.
The \emph{test error} counts how many of the test (unseen) instances (one in each repetition of the cross-validation) were correctly classified by the classifier (trained on the corresponding percentage of the training dataset). The error rates of Fig.~\ref{TestTrainError} are averaged over the 655 repetitions of the cross-validation.\footnote{We 
tuned the similarity threshold $T$
(Section~\ref{PatternExtraction})
to 0.1, based on additional cross-validation experiments.}
The training error curve can be thought of as a lower bound of the test error curve, since a classifier typically performs better on the instances it has been trained 
on than on unseen instances. The two curves indicate that the classifier might perform slightly better with more training data, though the test error rate would remain above $0.1$. The relatively small distance of the right ends of the two curves indicates only mild overfitting when the entire training dataset is used. 

\begin{figure}
\center
\includegraphics[width=\columnwidth]{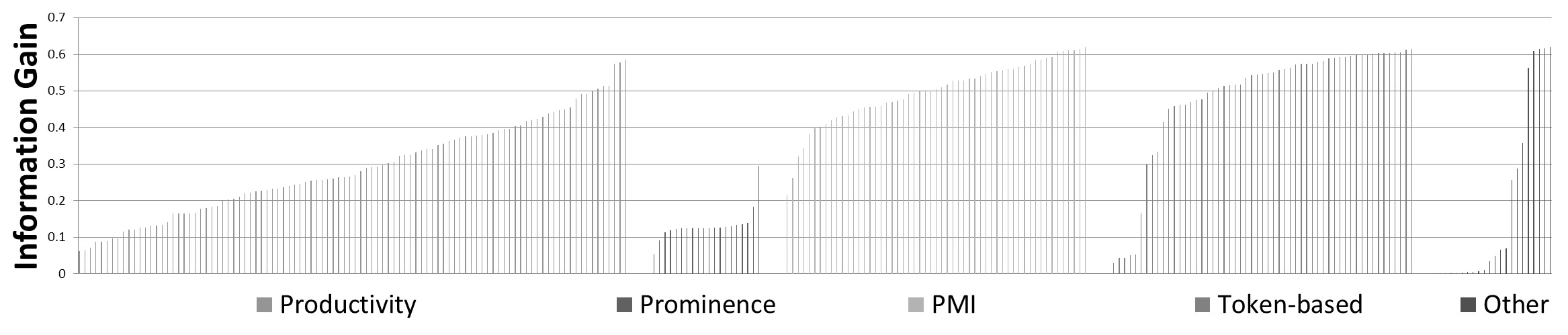}
\vspace*{-7mm}
\caption{Ascending 
\textsc{ig}  in each group of features used by
our sentence plan generation method.}
\label{InformationGainBars}
\end{figure}

To assess the contribution of the 251 features (Section~\ref{EvaluatingSentencePlans}), we ranked them by decreasing Information Gain (\textsc{ig}) \cite{Manning2000} computed on the 655 instances. Table~\ref{IGTable} shows the maximum, minimum, and average \textsc{ig} scores of the features in each group (subsection) of Section~\ref{EvaluatingSentencePlans}. On average, the \pmi and token-based features are the best predictors, whereas the prominence features are the worst. The maximum scores, however, show that there is at least one good feature in each group, with the prominence features being again the worst in this respect. The minimum scores indicate that there is also at least one weak feature in every group, with the exception of the \pmi features, where the minimum \textsc{ig} score (0.21) was much higher. Figure~\ref{InformationGainBars} shows the \textsc{ig} scores of all the features in each group, in ascending order. There are clearly many good features in every group, 
with the prominence features again being the weakest group overall.

\begin{figure}
\center
\includegraphics[width=0.5\columnwidth]{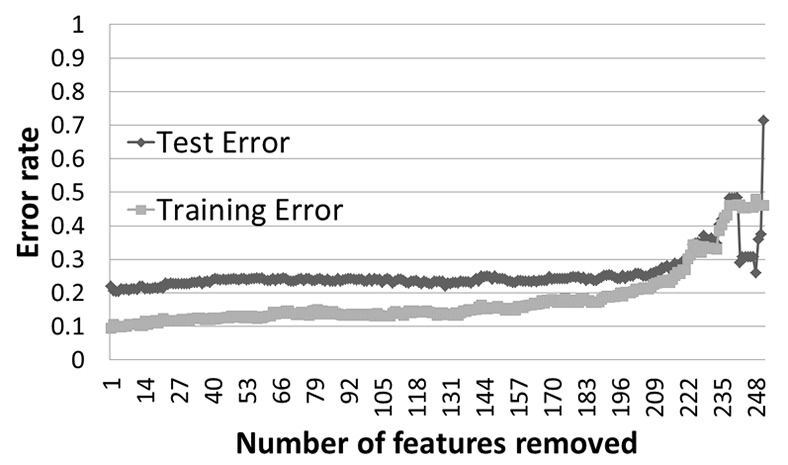}
\vspace*{-4mm}
\caption{
Error rate of the classifier of our sentence plan 
method, for different 
numbers of features.}
\label{FeatureSelection}
\end{figure}

We then iteratively trained the classifier on the entire training dataset ($100\%$), removing at each iteration the feature (among the remaining ones) with the smallest \textsc{ig} score. Figure~\ref{FeatureSelection} shows the resulting test and training error rate curves, again obtained using a leave-one-out cross-validation. As more features are removed, the distance between the training and test error decreases, because of reduced overfitting. When very few features are left (far right), the performance of the classifier on unseen instances becomes unstable. The best results are obtained using all (or almost all) of the features, but the test error is almost stable from approximately 50 to  200 removed features, indicating that there is a lot of redundancy 
(e.g., correlated features) 
in the feature set. Nevertheless, we did not remove any features in the subsequent experiments, since the overfitting was reasonably low and the training and test times of the MaxEnt classifier were also low (performing a leave-one-out cross-validation on the 655 instances with all the features took approximately 6 minutes). We hope to explore dimensionality reduction further (e.g., via \textsc{pca}) in future work.

\begin{table}
\center
{\footnotesize
\begin{tabular}{|l|c|c|c|}
\hline               & \textsc{Avg.\ \textsc{ig}} & \textsc{Max.\ \textsc{ig}} & \textsc{Min.\ \textsc{ig}} \\
\hline 
\hline Productivity features    & $0.29$ & $0.58$ & 0.06 \\
\hline Prominence features      & $0.13$ & $0.29$ & 0.05 \\
\hline \pmi features      				& $0.50$ & $0.62$ & 0.21 \\
\hline Token-based features     & $0.48$ & $0.61$ & 0.03 \\
\hline Other features      			& $0.20$ & $0.62$ & 0.00 \\
\hline
\end{tabular}
}
\caption{Information Gain of different groups of features 
of our sentence plan generation method.}
\label{IGTable}
\end{table}

\subsubsection{Inspecting the produced sentence plans} \label{inspectingSPs}

In a subsequent experiment, the classifier was trained on the 655 instances (candidate sentence plans) of the previous section; recall that those instances were obtained from Wikipedia articles for Wine Ontology relations. The classifier was then embedded (without retraining) in our overall sentence plan generation method (\spg or \spgrerank, see Section~\ref{rankingSPs}).
The sentence plan generation method 
was then invoked to produce sentence plans from the entire Web, not just Wikipedia, 
and for the relations of all three ontologies, not just those of the Wine Ontology.\footnote{
To avoid testing the classifier on sentence plans it had encountered during its training (more likely to happen with sentence plans for Wine Ontology relations),
we actually excluded from the training data of the classifier any of the 655 instances that had the same feature vector with any 
test candidate sentence plan of the experiments of this section,
when classifying that particular test candidate sentence plan.} We kept the top 10 returned documents per Web search (Section~\ref{PatternExtraction}),
to reduce the time 
to complete them. The first author
inspected the 
top 5 sentence plans
per relation (as ranked by \spg or \spgrerank),
marking them as correct or incorrect (as in Section~\ref{trainingClassifier}). We then computed the \emph{1-in1}, \emph{1-in-5}, and \textsc{mrr} scores of the produced sentence plans per ontology, along with weighted variants of the three measures. All six measures are defined as in Section~\ref{inspectingNLnames}, but for sentence plans instead of \nl names; the weighted variants weigh each relation by the number of \owl statements that mention it in the ontology.

Tables \ref{wineMRRTable}--\ref{diseaseMRRTable} show the results for the three ontologies. The 
configurations ``with seeds of \authnl'' use the \nl names from the manually authored linguistic resources to obtain seed names (Section~\ref{PatternExtraction}); by contrast, the configurations ``with seeds of \semiautonl'' use the semi-automatically produced \nl names to obtain seed names.
Recall that  \spgrerank reranks the candidate sentence plans using their coverage (Section~\ref{rankingSPs}). Tables \ref{wineMRRTable}--\ref{diseaseMRRTable} also include results for a bootstrapping baseline (\bootstrap), described below. For each measure, the best results  are shown in bold.

\begin{table}
\begin{center}
{\footnotesize
\begin{tabular}{|l|c|c|c|c|c|}
\hline               & \spg with  seeds of & \spgrerank with	 seeds of & \spg	 with seeds of & \spgrerank with seeds of & \bootstrap with seeds of \\
                       & \authnl     &  \authnl              &  \semiautonl                      & \semiautonl                               & \authnl \\
\hline 
\hline 1-in-1        & $0.71$ (5/7)						& $\mathbf{0.86}$ (6/7)	& $0.71$ (5/7)					& $0.71$ (5/7)					& $0.57$ (4/7)\\
\hline 1-in-3        & $0.86$ (6/7)						& $0.86$ (6/7)					& $\mathbf{1.00}$ (7/7)	& $\mathbf{1.00}$ (7/7)	& $0.71$ (5/7)\\
\hline 1-in-5        & $\mathbf{1.00}$ (7/7)	& $\mathbf{1.00}$ (7/7)	& $\mathbf{1.00}$ (7/7)	& $\mathbf{1.00}$ (7/7)	& $0.86$ (6/7)\\
\hline MRR     			 & $0.82$ 								& $\mathbf{0.89}$ 			& $0.83$ 								& $0.86$ 								& $0.68$ \\
\hline 
\hline weighted 1-in-1 & $0.73$ 					& $\mathbf{0.86}$ 	& $0.71$ 						& $0.71$ 						& $0.58$ \\
\hline weighted 1-in-3 & $0.86$ 					& $0.86$ 						& $\mathbf{1.00}$ 	& $\mathbf{1.00}$ 	& $0.71$ \\
\hline weighted 1-in-5 & $\mathbf{1.00}$ 	& $\mathbf{1.00}$ 	& $\mathbf{1.00}$ 	& $\mathbf{1.00}$ 	& $0.86$ \\
\hline weighted MRR   	& $0.83$ 					& $\mathbf{0.89}$ 	& $0.83$ 						& $0.86$ 						& $0.68$ \\
\hline
\end{tabular}
}
\caption{Results of the inspection of the produced sentence plans for the Wine Ontology.}
\label{wineMRRTable}
\end{center}
\vspace*{-5mm}
\end{table}

\begin{table}
\begin{center}
{\footnotesize
\begin{tabular}{|l|c|c|c|c|c|}
\hline               & \spg with  seeds of & \spgrerank with	 seeds of & \spg	 with seeds of & \spgrerank with seeds of & \bootstrap with seeds of \\
                       & \authnl     &  \authnl              &  \semiautonl                      & \semiautonl                               & \authnl \\
\hline 
\hline 1-in-1        & $0.58$ (7/12)			& $\mathbf{0.67}$ (8/12)	& $\mathbf{0.67}$ (8/12)	& $\mathbf{0.67}$ (8/12)	& $0.17$ (2/12)\\
\hline 1-in-3        & $0.67$ (8/12)			& $0.67$ (8/12)						& $\mathbf{0.75}$ (9/12)	& $\mathbf{0.75}$ (9/12)	& $0.33$ (4/12)\\
\hline 1-in-5        & $0.67$ (8/12)			& $0.67$ (8/12)						& $\mathbf{0.83}$ (10/12)	& $\mathbf{0.83}$ (10/12)	& $0.33$ (4/12)\\
\hline MRR     			 & $0.62$ 						& $0.67$ 									& $\mathbf{0.73}$ 				& $\mathbf{0.73}$ 				& $0.24$ \\
\hline 
\hline weighted 1-in-1 & $0.73$ 					& $\mathbf{0.85}$ & $0.61$ 						& $0.72$ 						& $0.04$ \\
\hline weighted 1-in-3 & $\mathbf{0.85}$ 	& $\mathbf{0.85}$ & $0.73$ 						& $0.73$						& $0.19$ \\
\hline weighted 1-in-5 & $0.85$ 					& $0.85$ 					& $\mathbf{0.98}$ 	& $\mathbf{0.98}$ 	& $0.45$ \\
\hline weighted MRR  	 & $0.79$ 					& $\mathbf{0.85}$ & $0.73$ 						& $0.79$ 						& $0.22$ \\
\hline
\end{tabular}
}
\caption{Results of the inspection of the produced sentence plans for the \mpiro ontology.}
\label{mpiroMRRTable}
\end{center}
\vspace*{-7mm}
\end{table}

Overall, \spgrerank performs better than \spg, though the scores of the two methods are very close or identical in many cases, and occasionally \spg performs better. 
Also, \spg and \spgrerank occasionally perform better when semi-automatically produced \nl names are used to obtain seed names,
than when manually authored \nl names are used.
It seems that manually authored \nl names occasionally produce seeds that are uncommon on the Web and, hence, do not help produce good sentence plans, unlike semi-automatically produced \nl names, which are extracted from the Web (and then manually filtered). The 1-in-5 results of Tables~\ref{wineMRRTable}--\ref{diseaseMRRTable} 
show that our sentence plan generation method 
(especially \spgrerank) performs 
very well in a semi-automatic scenario, 
especially if the weighted measures are considered. By contrast, our method does not always perform well in a fully automatic scenario (1-in-1 results);
the Disease Ontology was the most difficult in that respect.
Overall, \spgrerank with seeds of \semiautonl seems to be the best version.
The \textsc{mrr} scores of our method (all versions) were higher in the Wine Ontology and lower in the other two ones, which may be due to the fact that the classifier was trained for Wine Ontology relations (but with texts from Wikipedia).

\begin{table}
\begin{center}
{\footnotesize
\begin{tabular}{|l|c|c|c|c|c|}
\hline               & \spg with  seeds of & \spgrerank with	 seeds of & \spg	 with seeds of & \spgrerank with seeds of & \bootstrap with seeds of \\
                       & \authnl     &  \authnl              &  \semiautonl                      & \semiautonl                               & \authnl \\
\hline 
\hline 1-in-1        & $0.00$ (0/8)						& $0.37$ (3/8)						& $0.12$ (1/8)						& $\mathbf{0.50}$ (4/8)	& $0.12$ (1/8)\\
\hline 1-in-3        & $0.75$ (6/8)						& $\mathbf{0.87}$ (7/8)		& $0.75$ (6/8)						& $\mathbf{0.87}$ (7/8)	& $0.37$ (3/8)\\
\hline 1-in-5        & $\mathbf{1.00}$ (8/8)	& $\mathbf{1.00}$ (8/8)		& $\mathbf{1.00}$ (8/8)		& $\mathbf{1.00}$ (8/8) 
& $0.75$ (6/8)\\
\hline MRR     			 & $0.41$ 								& $0.65$ 									& $0.47$ 									& $\mathbf{0.70}$ 			& $0.33$ \\
\hline 
\hline weighted 1-in-1 	& $0.00$ 					& $0.16$ 						& $0.01$ 						& $\mathbf{0.17}$ 	& 
$0.00$ \\
\hline weighted 1-in-3 	& $0.42$ 					& $\mathbf{0.89}$ 	& $\mathbf{0.89}$ 	& $\mathbf{0.89}$ 	& $0.15$ \\
\hline weighted 1-in-5 	& $\mathbf{1.00}$ & $\mathbf{1.00}$ 	& $\mathbf{1.00}$ 	& $\mathbf{1.00}$ 	& $0.85$ \\
\hline weighted MRR  		& $0.35$ 					& $\mathbf{0.55}$ 	& $0.45$ 						& $0.48$ 						& $0.22$ \\
\hline
\end{tabular}
}
\caption{Results of the inspection of the produced sentence plans for the Disease Ontology.}
\label{diseaseMRRTable}
\end{center}
\vspace{-5mm}
\end{table}

While inspecting the 
sentence plans, we noticed 
several cases where the \owl identifier of the relation was poor (e.g., the \code{:locatedIn} relation of the Wine Ontology connects wines to the regions
producing them, but our method  
produced good sentence plans (e.g., [$\mathit{ref}(S)$] [is produced] [in] [$\mathit{ref}(O)$]). 
On the other hand, our method (all versions) 
produced very few (or none) 
sentence plans for relations with fewer than 10 seed name pairs. 
Also, the most commonly produced sentence plans are [$\mathit{ref}(S)$] [is] [$\mathit{ref}(O)$] and [$\mathit{ref}(O)$] [is] [$\mathit{ref}(S)$]. While the former may be 
appropriate for a message triple $\left<S, R, O\right>$, the latter is almost never appropriate, so we always discard it.

\subsubsection{Agreement and effort to semi-automatically author sentence plans}

The top five sentence plans of \spgrerank for each relation were also shown to the second human judge (\judgetwo) who had examined the automatically produced \nl names in the experiments of Section~\ref{agreementNLnames}. For each relation, and for each one of its top five sentence plans, the second judge was shown a template view of the sentence plan (e.g., ``$S$ is made from $O$''), and an automatically generated sentence illustrating
its use (e.g., ``Cabernet Sauvignon is made from Cabernet Sauvignon grapes.''). The judge was asked to consider the templates and sentences, and mark the best correct sentence plan for each relation. The judge could also mark more than one sentence plans for the same relation, if more than one seemed correct and equally good; the judge was instructed not to mark any of the five sentence plans, if none seemed correct. 
The second judge completed this task 
(inspecting 40, 29, and 40 candidate sentence plans of the Wine, \mpiro, and Disease Ontology, respectively) 
in approximately 5 minutes per ontology (15 minutes for all three ontologies); 
by contrast, 
manually authoring the sentence plans took approximately one working day per ontology.
Again, these times suggest that the semi-automatic authoring scenario is 
viable and very useful in practice.

We also measured the agreement between the second judge (\judgetwo) and the first author 
(\judgeone) in the semi-automatic authoring (selection) of sentence plans, as in Section~\ref{agreementNLnames}. The results, reported in Table~\ref{ontologyAgreement_SP_Table}, 
show perfect agreement in the Wine and Disease Ontologies. In the \mpiro ontology, the agreement was lower, but still reasonably high; the pseudo-recall score shows \judgetwo did not select any sentence plan for some relations where \judgeone believed there was a correct one among the top five.

\begin{table}
\center
{\footnotesize
\begin{tabular}{|c|c|c|c|}
\hline               & \textsc{wine}	& \mpiro	& \textsc{disease} \\
\hline 
\hline micro-precision 	& $1.00$ 		& $1.00$ 		& $1.00$ \\
\hline macro-precision  & $1.00$ 		& $1.00$ 		& $1.00$ \\
\hline 
\hline \judgeone 1-in-5 			& $1.00$ 		& $0.83$ 		& $1.00$ \\
\hline \judgetwo 1-in-5 		& $1.00$ 		& $0.75$ 		& $1.00$ \\
\hline pseudo-recall   				& $1.00$  	& $0.75$ 		& $1.00$ \\
\hline 
\hline Cohen's Kappa		& $1.00$ 		& $0.62$		& $0.74$\\
\hline
\end{tabular}
}
\caption{Inter-annotator agreement in the semi-automatic authoring of sentence plans.}
\label{ontologyAgreement_SP_Table}
\end{table}

\subsubsection{The sentence plan generation baseline that uses bootstrapping} \label{bootstrapping}

As a baseline, we also implemented a sentence plan generation method that uses a bootstrapping template extraction approach. Bootstrapping is often used in information extraction to obtain templates that extract instances of a particular relation (e.g., \code{:makeFrom}) from texts, starting from seed pairs of entity names (e.g., $<$``cream'', ``milk''$>$) for which the relation is known to hold
\cite{Riloff1999,Muslea1999,Xu2007,gupta2014}. The seed pairs are used as queries in a search engine to obtain documents that contain them in the same sentence (e.g., ``cream is made from milk''). Templates are then obtained 
by replacing the seeds with slots in the retrieved sentences (e.g., ``$X$ is made from $Y$''). The templates (without their slots, e.g., ``is made from'') are then used as phrasal search queries to obtain new sentences (e.g., ``gasoline is made from petroleum''), from which new 
seed pairs ($<$``gasoline'', ``petroleum''$>$) are obtained. A new iteration can then start with the new 
seed pairs, leading to new templates, and so on. 

Given a relation $R$, our baseline, denoted \bootstrap, first constructs seed name pairs using the ontology and the \nl names, as in Section~\ref{PatternExtraction}; 
we used only manually authored \nl names in the experiments with \bootstrap. Then, \bootstrap uses the seed name pairs to obtain templates (``$X$ is made from $Y$'') from the Web, again as in Section~\ref{PatternExtraction}. If the number of obtained templates is smaller than $L$,  the templates (without their
slots) are used as phrasal search queries to obtain new documents and new sentences (from the documents) that match the templates. For each new sentence (e.g., ``gasoline is made from petroleum''), \bootstrap finds the \np{s} (``gasoline'', ``petroleum'') immediately before and after the search phrase, and treats them as a new seed name pair,
discarding 
pairs that occur in only one retrieved document. The new 
pairs are then used to obtain new 
templates, again discarding templates occurring in only one document. This process is repeated until we have at least $L$ templates for $R$, or until no new templates can be produced. In our experiments, we set $L = 150$ to obtain approximately the same number of templates  
as with \spg and \spgrerank.

At the end of the bootstrapping, instead of using a MaxEnt classifier (Sections~\ref{EvaluatingSentencePlans} and \ref{rankingSPs}), \bootstrap scores the templates of each relation $R$ using the following confidence function:\footnote{
This function is from
Jurafsky \& Marting \citeyear{JurafskyMartin2008}, and is based on a similar function of 
Riloff \& Jones \citeyear{Riloff1999}. Unlike Jurafsky 
\& Martin, we count \np anchor pairs rather than seed name pairs in $\mathit{hits}(t)$ and $\mathit{misses}(t)$, 
placing more emphasis on extracting all the 
\np{s} (of the retrieved documents of $R$) that correspond to the seed names.}

\begin{equation}\label{bootstrapConfidenceFunction}
\mathit{conf}(t) = 
\frac{\mathit{hits}(t)}{\mathit{hits}(t) + \mathit{misses}(t)}
\cdot \log \mathit{finds}(t)
\end{equation}

\noindent where $t$ is a template being scored, $\mathit{hits}(t)$ is the 
number of (distinct) 
\np anchor pairs of $R$ extracted by $t$ (from the documents retrieved by all the seed name pairs of $R$), $\mathit{misses}(t)$ is the number of (distinct) 
\np anchor pairs of $R$ \emph{not} extracted by $t$ (from the same documents), and $\mathit{finds}(t)$ is the number of sentences (of the same documents) that match $t$.
The five templates with the highest $\mathit{conf}(t)$ (in a semi-automatic scenario) or the single template with the highest $\mathit{conf}(t)$ (in a fully automatic scenario) are then converted to sentence plans, as in Section~\ref{SentencePlanGeneration}. 
  
Functions like Eq.~\ref{bootstrapConfidenceFunction} can also be applied \emph{within} each iteration of the bootstrapping, not only at the end of the entire bootstrapping, to keep only the best new templates of each iteration. This may help avoid concept drift, i.e., gradually obtaining templates that are more appropriate for other relations that 
share seed name pairs with the relation we wish to generate templates for. We did not use Eq.~\ref{bootstrapConfidenceFunction} within each iteration,
because in our experiments very few iterations (most often only the initial one) were needed. Also, using a function like Eq.~\ref{bootstrapConfidenceFunction} within each iteration requires a threshold $d$, to discard templates with $\mathit{conf}(t)<d$ at the end of each iteration, which 
is not trivial to tune. Similar functions can be used to score the new seed name pairs within each iteration or at the end of the bootstrapping.\footnote{
Consult Chapter 22 of 
Jurafsky \& Martin \citeyear{JurafskyMartin2008} for a broader discussion of bootstrapping template extraction methods. See also \url{http://nlp.stanford.edu/software/patternslearning.shtml} for a publicly available bootstrapping template extraction system (\textsc{spied}), which supports however only templates with a single slot.}
Since very few iterations (most often only the initial one) were needed in our experiments, we ended up using mostly (and most often only) the initial seed name pairs, which are known to be correct; hence, scoring the seed name pairs seemed unnecessary.  

Tables~\ref{wineMRRTable}--\ref{diseaseMRRTable} show that the results of \bootstrap are consistently worse than the results of \spg and \spgrerank.
As already noted, for most relations more than $L$ templates had been produced at the end of the first iteration (with the initial seed name pairs) of \bootstrap. Additional iterations were used only for 5 relations of the \mpiro ontology. Hence, the differences in the performance of \bootstrap compared to \spg and \spgrerank are almost entirely due to the fact that \bootstrap uses the confidence function of Eq.~\ref{bootstrapConfidenceFunction} instead of the MaxEnt classifier (and the coverage of the sentence plans, in the case of \spgrerank). Hence, the MaxEnt classifier has an important contribution in the performance of \spg and \spgrerank.
Tables~\ref{wineMRRTable}--\ref{diseaseMRRTable} show that this contribution is large in all three ontologies, despite the fact that the classifier was trained on Wine Ontology relations only (but with texts from Wikipedia). 

\subsubsection{Evaluating sentence plans in generated texts} \label{evaluatingSPtexts}

To examine how sentence plans produced by different methods affect the perceived quality of generated texts, we showed automatically generated texts describing individuals and classes of the three ontologies to six computer science students, the same students as in the experiments of Section~\ref{evaluatingNLNtexts}. We used six configurations of \nlowl in this experiment. The \nosp configuration is given no sentence plans; in this case, \nlowl automatically produces sentence plans by tokenizing the \owl identifiers of the relations, acting like a simple verbalizer.\footnote{If the ontology provides an \code{rdfs:label} for a relation, \nosp 
uses its tokens.} \authsp uses manually authored sentence plans. \autospgrerank uses the \spgrerank method (Section~\ref{rankingSPs}) with no human selection of sentence plans, i.e., the top-ranked sentence plan of each relation. We did not consider \spg in this experiment, since the 
previous experiments indicated that \spgrerank was overall better. In \semiautospgrerank, a human inspector (the first author)
selected the best sentence plan of each relation among the five top-ranked sentence plans 
of \spgrerank. Similarly, \autobootstrap and \semiautobootstrap use the \bootstrap baseline of Section~\ref{bootstrapping} with no human selection or with a human 
selecting among the top five, respectively. Apart from the sentence plans, all six configurations use the same, manually authored other types of linguistic resources (e.g., \nl names, interest scores, text plans to order the message triples). Below are example texts generated from the three ontologies by the six configurations. 

\medskip
{\small
\noindent
\authsp: This is a moderate, dry Zinfandel. It has a full body. It is made by Elyse in the Napa County.\\
\semiautospgrerank: This is a full, dry Zinfandel. It is moderate. It is made at Elyse in the Napa County.\\
\semiautobootstrap: This is a full, dry Zinfandel. It is moderate. Elyse produced it and the Napa County is Home to it. \\
\autospgrerank: This is a full, dry Zinfandel. It is moderate. It is made at Elyse and it is the Napa County.\\
\autobootstrap: This is a full, dry Zinfandel. It is moderate. It is Elyse and the Napa County. \\
\nosp: This is a Zinfandel. It has sugar dry, it has body full and it has flavor moderate. It has maker Elyse and it located in the Napa County.
}

\medskip
{\small
\noindent
\authsp: This is a kantharos, created during the Hellenistic period and it originates from Amphipolis. Today it is exhibited in the Archaeological Museum of Kavala.\\
\semiautospgrerank: This is a kantharos, produced during the Hellenistic period and almost certainly from Amphipolis. It is found in the Archaeological Museum of Kavala. \\
\semiautobootstrap: This is a kantharos, produced during the Hellenistic period and almost certainly from Amphipolis. It is among the Archaeological Museum of Kavala. \\
\autospgrerank: This is a kantharos that handles from the Hellenistic period and is almost certainly from Amphipolis. It derives from the Archaeological Museum of Kavala.\\
\autobootstrap: This is a kantharos that is the Hellenistic period and is Amphipolis. It is the Archaeological Museum of Kavala. \\
\nosp: This is a kantharos. It creation period the Hellenistic period, it original location Amphipolis and it current location the Archaeological Museum of Kavala. 
}

\medskip
{\small
\noindent
\authsp: Molluscum contagiosum is a kind of viral infectious disease that affects the skin. It results in infections and its symptom is lesions. It is transmitted by fomites and contact with the skin, and it is caused by the molluscum contagiosum virus. \\
\semiautospgrerank: Molluscum contagiosum is a kind of viral infectious disease that can occur in the skin. Infections are caused by molluscum contagiosum. Molluscum contagiosum often causes lesions. It is transmissible by fomites and contact with the skin, and it is caused by the molluscum contagiosum virus.\\
\semiautobootstrap: Molluscum contagiosum is a kind of viral infectious disease that occurs when the skin. Infections are caused by molluscum contagiosum. Molluscum contagiosum can cause lesions. Fomites and contact with the skin can transmit molluscum contagiosum. Molluscum contagiosum is caused by the molluscum contagiosum virus.\\
\autospgrerank: Molluscum contagiosum is a kind of viral infectious disease that is the skin. It is infections. It is lesions. It is fomites and contact with the skin, and it is caused by the molluscum contagiosum virus. \\
\autobootstrap: Molluscum contagiosum is a kind of viral infectious disease that is the skin. It is infections. It is lesions. It is fomites, the molluscum contagiosum virus and contact with the skin.\\
\nosp: Molluscum contagiosum is a kind of viral infectious disease and it located in the skin. It results in infections. It has symptom lesions. It transmitted by fomites and contact with the skin, and it has material basis in the molluscum contagiosum virus. 
}
\medskip

As in the corresponding experiment with \nl names (Section~\ref{evaluatingNLNtexts}), we generated 
$95 \times 6 = 570$ texts from the Wine ontology (this time with six configurations), $49 \times 6 = 294$ texts from the \mpiro ontology, and $30 \times 6 = 180$ texts from the Disease ontology.\footnote{Compared to the Wine and Disease ontologies, the \mpiro ontology provides fewer seeds, which leads to no sentence plans for some of its relations. For those relations, we relaxed 
the constraint of Section~\ref{PatternExtraction} that requires templates to be extracted from at least two sentences, but this did not always produce good sentence plans.} The students were asked to score each text by stating how strongly they agreed or disagreed with statements $S_1$--$S_3$ below; the non-redundancy criterion was not used in this experiment, because all six configurations used the same (manually authored) \nl names and interest scores. Otherwise, the experimental setup was the same as in the corresponding experiment with \nl names (Section~\ref{evaluatingNLNtexts}).

\smallskip
($S_1$) \emph{Sentence fluency}:  Each sentence of the text (on its own) is grammatical and sounds natural.

($S_2$) \emph{Clarity}: The text is easy to understand, provided that the reader is familiar with the terminology and concepts of the domain (e.g., historical periods, grape varieties, virus names).

($S_3$) \emph{Semantic correctness}: The text accurately conveys the information of the message triples.

\vspace{1mm}
Tables \ref{sp_wine_human_results}--\ref{sp_disease_human_results} show the scores of the six configurations of \nlowl, averaged over the texts of each ontology. For each criterion, the best scores are shown in bold. 
In each criterion (row), we detected no statistically significant differences between scores marked with the same superscript; all the other differences (in the same row) were statistically significant.\footnote{
We performed \textsc{anova} and post-hoc Tukey tests
($a = 0.05$) in the six scores of each criterion in each ontology. A post-hoc power analysis of the \textsc{anova} values resulted in power values greater or equal to $0.95$ in all cases.}
\authsp was the best overall configuration, as one would expect, but \semiautospgrerank performed only slightly worse, with no detected statistically significant difference between the two configurations in most cases. The only notable exception was the semantic correctness of the \mpiro ontology, where the difference between \semiautospgrerank and \authsp was larger, because for some relations \semiautospgrerank produced sentence plans that did not correctly express  the corresponding message triples, because of too few seeds. These findings confirm 
that \spgrerank performs very well in a semi-automatic scenario.
\semiautobootstrap performed clearly worse than \semiautospgrerank in this scenario. 

In the fully automatic scenario, with no human selection of sentence plans, \autospgrerank was overall better than \autobootstrap, but still not good enough to be used in practice. The \nosp configuration, which uses sentence plans constructed by tokenizing the identifiers of the \owl relations, obtained much lower sentence fluency and clarity scores than \authsp, which shows the importance of sentence plans in the perceived quality of the texts. The semantic correctness scores of \nosp were also much lower than those of \authsp in the Wine and \mpiro ontologies, but the difference was smaller (with no detected statistically significant difference) in the Disease Ontology, because tokenizing the \owl identifiers of the relations of the Disease Ontology (e.g., \code{:has\_symptom})
leads to 
sentences that convey the correct information in most cases,
even if the sentences are not particularly fluent and clear. The sentence fluency and clarity scores of \nosp 
in the Disease Ontology 
were also higher, compared to the scores of \nosp in the other two ontologies, for the same reason.

\begin{table*}[htbp]
\begin{center}
{\footnotesize
\begin{tabular}{|l||c|c|c|c|c|c|}
\hline Criteria              		& \nosp  	& \autobootstrap  & \semiautobootstrap	& \autospgrerank  		& \semiautospgrerank  				& \authsp \\
\hline 
\hline Sentence fluency     		& $2.69$				& $3.67^{1}$  & $4.18^{2}$ 			& $3.88^{1,2}$  & $\mathbf{4.71}^{3}$ 	& $\mathbf{4.75}^{3}$\\
\hline Clarity        			 			& $3.26^{1,2}$ 	& $2.89^{1}$  & $4.28$ 					& $3.47^{2}$  	& $\mathbf{4.81}^{3}$ 	& $\mathbf{4.90}^{3}$\\
\hline Semantic correctness  		& $3.02^{1,2}$ 	& $2.71^{1}$  & $4.27$ 					& $3.21^{2}$  	& $\mathbf{4.77}^{3}$ 	& $\mathbf{4.93}^{3}$\\
\hline
\end{tabular}
}
\caption{Human scores for 
the Wine Ontology with different methods to obtain sentence plans.}
\label{sp_wine_human_results}
\end{center}
\vspace{-7mm}
\end{table*}

\begin{table*}[htbp]
\begin{center}
{\footnotesize
\begin{tabular}{|l||c|c|c|c|c|c|}
\hline Criteria              	& \nosp  	& \autobootstrap  & \semiautobootstrap	& \autospgrerank  		& \semiautospgrerank  				& \authsp \\
\hline 
\hline Sentence fluency      		& $1.18$				& $2.73$ 			& $3.63^{1}$ 			& $4.08^{1}$  	& $\mathbf{4.67}^{2}$ 	& $\mathbf{4.98}^{2}$\\
\hline Clarity        			 		& $2.39$ 				& $1.67$  		& $3.12^{1}$ 			& $3.28^{1}$  	& $\mathbf{4.65}^{2}$ 	& $\mathbf{5.00}^{2}$\\
\hline Semantic correctness  			& $2.96^{1}$ 		& $1.90$  		& $2.90^{1}$ 			& $2.86^{1}$  	& $4.08$ 								& $\mathbf{5.00}$\\
\hline
\end{tabular}
}
\caption{Human scores for 
the \mpiro ontology with different methods to obtain sentence plans.}
\label{sp_mpiro_human_results}
\end{center}
\vspace{-7mm}
\end{table*}

\begin{table*}[htbp]
\begin{center}
{\footnotesize
\begin{tabular}{|l||c|c|c|c|c|c|}
\hline Criteria              	& \nosp  	& \autobootstrap  & \semiautobootstrap	& \autospgrerank  		& \semiautospgrerank  				& \authsp \\
\hline 
\hline Sentence fluency      				& $3.40^{1}$							& $3.67^{1}$  & $\mathbf{4.07}^{1,2}$ 	& $3.77^{1}$  	& $\mathbf{4.77}^{2}$ 	& $\mathbf{4.83}^{2}$\\
\hline Clarity        						& $3.80^{1,2}$ 						& $2.37^{1}$  & $3.13$ 									& $2.53^{2}$  	& $\mathbf{4.73}^{3}$ 	& $\mathbf{4.83}^{3}$\\
\hline Semantic correctness  			& $\mathbf{4.10}^{1,2}$ 	& $2.40^{3}$  & $3.27^{1,3}$ 						& $2.63^{2}$  	& $\mathbf{4.57}^{2}$ 	& $\mathbf{4.83}^{2}$\\
\hline
\end{tabular}
}
\caption{Human scores for 
the Disease Ontology with different methods to obtain sentence plans.}
\label{sp_disease_human_results}
\end{center}
\end{table*}

\subsection{Joint experiments with extracted NL names and sentence plans} \label{JointExperiments}

In a final set of experiments, we examined the effect of combining our methods that produce \nl names and sentence plans. We experimented with four configurations of \nlowl. The \jointauto configuration produces \nl names using the method of Section~\ref{OurMethodNLN}; it then uses the most highly ranked \nl name of each individual or class to produce seed name pairs, and invokes the \spgrerank method of Section~\ref{OurMethodSP} to produce sentence plans; it then uses the most highly ranked sentence plan for each relation. \jointsemiauto also produces \nl names using the method of Section~\ref{OurMethodNLN}, but a human selects the best \nl name of each individual or class among the five most highly ranked ones; the selected \nl names are then used to produce seed name pairs, and the \spgrerank method is invoked to produce sentence plans; a human then selects the best sentence plan of each relation among the five most highly ranked ones. The \manual configuration uses manually authored \nl names and sentence plans. In the \verbalizer configuration, no \nl names and sentence plans are provided to \nlowl; hence, acting like a simple verbalizer, \nlowl produces \nl names and sentence plans by tokenizing the \owl identifiers of individuals, classes, and relations.\footnote{
Whenever the ontology provides an \code{rdfs:label} for an individual, class, or relation, \verbalizer
uses its tokens.} Furthermore, \manual uses manually authored intrest scores, \verbalizer uses no interest scores, whereas \jointauto and \jointsemiauto use interest scores obtained from the (top-ranked or selected) \nl names using the method of Section~\ref{InterestReasoning}. All the other linguistic resources (most notably, text plans)
are the same (manually authored) across all four configurations. We did not experiment with the \bootstrap and \spg sentence plan generation methods in this section, since \spgrerank performed overall better in the previous experiments. Below are example texts generated from the three ontologies by the four configurations we considered.

\medskip
{\small
\noindent
\manual: This is a moderate, dry Chenin Blanc. It has a full body. It is made by Foxen in the Santa Barbara County.\\
\jointsemiauto: This is a full, dry Chenin Blanc wine. It is moderate. It is made at the Foxen Winery in the Santa Barbara region.\\
\jointauto: This is dry Chenin Blanc and is the full. It is the moderate. It is made at Foxen and it is Santa Barbara.\\
\verbalizer: Foxen Chenin Blanc is Chenin Blanc. It has sugar Dry, it has maker Foxen and it has body Full. It located in Santa Barbara Region and it has flavor Moderate. 
}

\medskip
{\small
\noindent
\manual: This is a portrait that portrays Alexander the Great and it was created during the Roman period. It is made of marble and today it is exhibited in the Archaeological Museum of Thassos.\\
\jointsemiauto: This is a portrait. It is thought to be Alexander the Great, it is produced during the Roman period and it was all hand carved from marble. It is found in the Archaeological Museum of Thasos. \\
\jointauto: This is a portrait. It is thought to be Alexander the Great, it handles from Roman and it is marble. It derives from the Thasos Archaeological Museum.\\
\verbalizer: Exhibit 14 is portrait. It exhibit portrays alexander the great. It creation period roman period. It made of marble. It current location thasos archaelogical. 
}

\medskip
{\small
\noindent
\manual: Ebola hemorrhagic fever is a kind of viral infectious disease. Its symptoms are muscle aches, sore throat, fever, weakness, stomach pain, red eyes, joint pain, vomiting, headaches, rashes, internal and external bleeding, hiccups and diarrhea. It is transmitted by infected medical equipment, contact with the body fluids of an infected animal, and contaminated fomites and it is caused by Bundibugyo ebolavirus, Cote d'Ivoire ebolavirus, Sudan ebolavirus and Zaire ebolavirus. \\
\jointsemiauto: An Ebola hemorrhagic fever is a kind of viral Infectious disease. It often causes a muscle ache, a sore throat, a fever, weakness, stomach pain, a red eye symptom, joint pain, vomiting, a headache, rash, a severe internal bleeding, hiccups and diarrhea. It is transmissible by contaminated medical equipment, the direct contact with infected animals, and the contaminated fomite and it is caused by the species Bundibugyo ebolavirus, the Côte d’Ivoire ebolavirus, the Sudan ebolavirus and the Zaire ebolavirus. \\
\jointauto: An Ebola hemorrhagic fever is viral. It is a muscle aches, contaminated medical equipment, a sore throat, a fever, weakness, stomach pain, a red eye, joint pain, a vomiting, a headache, rash, a content internal bleeding symptom, a hiccups and diarrhea. It is caused by the Bundibugyo ebolavirus, a Côte d’Ivoire ebolavirus, the Sudan ebolavirus and the Zaire ebolavirus.\\
\verbalizer: Ebola hemorrhagic fever is a kind of viral infectious disease. It has symptom muscle aches, sore throat, fever, weakness, stomach pain, red eyes, joint pain, vomiting, headache, rash, internal and external bleeding, hiccups and diarrhea. It transmitted by infected medical equipment, contact with the body fluids of an infected animal, and contaminated fomites and it has material basis in Bundibugyo ebolavirus, Cote d'Ivoire ebolavirus, Sudan ebolavirus and Zaire ebolavirus.
}
\medskip

\noindent Again, 
some \nl names and sentence plans of \jointsemiauto and \jointauto can be easily improved using the \protege plug-in of \nlowl. For example, the sentence plan that 
reports the historical period of the exhibit in the second \jointsemiauto example above can be easily modified to use the simple past tense (``was produced'' instead of ``is produced''). Nevertheless, we made no such improvements. 

Apart from the configurations of \nlowl, the experimental setup was the same as in Sections~\ref{evaluatingNLNtexts} and \ref{evaluatingSPtexts}. We generated $95 \times 4 = 380$ texts from the Wine ontology,
$49 \times 4 = 196$ texts from the \mpiro ontology, and $30 \times 4 = 120$ texts from the Disease ontology. The students were now asked to score each text for sentence fluency, clarity, semantic correctness, and non-redundancy, by stating how strongly they agreed or disagreed with statements $S_1$--$S_4$ of Section~\ref{evaluatingNLNtexts}. 

\begin{table*}[htbp]
\begin{center}
{\footnotesize
\begin{tabular}{|l||c|c|c|c|}
\hline Criteria              		& \verbalizer  					& \jointauto  		& \jointsemiauto 						& \manual \\
\hline 
\hline Sentence fluency      		& $2.77$								& $3.90$  			& $\mathbf{4.67}^{1}$ 		& $\mathbf{4.81}^{1}$\\
\hline Clarity        			 		& $3.57$ 								& $2.79$  			& $\mathbf{4.87}^{1}$ 		& $\mathbf{4.93}^{1}$\\
\hline Semantic correctness  				& $3.57$ 								& $2.86$  			& $\mathbf{4.68}^{1}$ 		& $\mathbf{4.97}^{1}$\\
\hline Non-redundancy  									& $3.20^{1}$ 						& $3.47^{1}$  	& $\mathbf{4.57}^{2}$ 		& $\mathbf{4.79}^{2}$\\
\hline
\end{tabular}
}
\caption{Human scores for 
the Wine Ontology, 
different methods for \nl names, sentence plans.}
\label{nlnsp_wine_human_results}
\end{center}
\vspace{-5mm}
\end{table*}

\begin{table*}[ht]
\begin{center}
{\footnotesize
\begin{tabular}{|l||c|c|c|c|}
\hline Criteria              		& \verbalizer  					& \jointauto  		& \jointsemiauto 						& \manual \\
\hline 
\hline Sentence fluency      			& $2.10$								& $4.33$  								& $\mathbf{4.73}^{1}$ 		& $\mathbf{4.96}^{1}$\\
\hline Clarity        			 			& $3.02^{1}$ 						& $3.49^{1}$  						& $4.43$ 									& $\mathbf{5.00}$\\
\hline Semantic correctness  				& $3.33^{1}$ 						& $2.90^{1}$  						& $3.96$ 									& $\mathbf{5.00}$\\
\hline Non-redundancy  								& $2.67$ 								& $4.16^{1}$  						& $\mathbf{4.59}^{1,2}$ 	& $\mathbf{5.00}^{2}$\\
\hline
\end{tabular}
}
\caption{Human scores for 
the \mpiro ontology,
different methods for \nl names, sentence plans.}
\label{nlnsp_mpiro_human_results}
\end{center}
\vspace{-5mm}
\end{table*}

\begin{table*}[ht]
\begin{center}
{\footnotesize
\begin{tabular}{|l||c|c|c|c|}
\hline Criteria              	& \verbalizer  					& \jointauto  		& \jointsemiauto 						& \manual \\
\hline 
\hline Sentence fluency      				& $3.20^{1}$							& $3.00^{1}$  	& $\mathbf{4.33}^{2}$ 		& $\mathbf{4.86}^{2}$\\
\hline Clarity        			 				& $3.77$ 									& $2.13$  			& $\mathbf{4.47}^{1}$ 		& $\mathbf{4.90}^{1}$\\
\hline Semantic correctness 			& $3.77^{1}$ 							& $2.43$  			& $\mathbf{4.20}^{1,2}$ 	& $\mathbf{4.93}^{2}$\\
\hline Non-redundancy  									& $\mathbf{4.20}^{1,2}$ 	& $3.43^{1}$  		& $\mathbf{4.37}^{2}$ 		& $\mathbf{4.70}^{2}$\\
\hline
\end{tabular}
}
\caption{Human scores for 
the Disease Ontology,
different methods for \nl names, sentence plans.}
\label{nlnsp_disease_human_results}
\end{center}
\vspace{-7mm}
\end{table*}

Tables \ref{nlnsp_wine_human_results}--\ref{nlnsp_disease_human_results} show the scores of the four configurations of \nlowl, averaged over the texts of each ontology. For each criterion, the best scores are shown in bold. 
In each criterion (row), we detected no statistically significant differences between scores marked with the same superscript; all the other differences (in the same row) were statistically significant.\footnote{
Again, we performed \textsc{anova} and post-hoc Tukey tests 
($a = 0.05$) in the four scores of each criterion in each ontology. A post-hoc power analysis of the \textsc{anova} values resulted in power values equal to $1.0$ in all cases. }
\manual had the best overall scores, as one would expect, but the scores of \jointsemiauto were close, in most cases with no detected statistically significant difference, despite the combined errors of the methods that produce \nl names and sentence plans. The biggest difference between \jointsemiauto and \manual was in the semantic correctness criterion of the \mpiro ontology. 
This difference is mostly due to the fact that \jointsemiauto did not always manage to produce sentence plans to convey correctly the semantics of the 
message triples, because of too few seeds, as in the experiments of the previous section. This also affected the clarity score of \jointsemiauto in the \mpiro ontology. The scores of \jointauto were much lower, 
again indicating that our methods cannot be used in a fully automatic scenario. 

The scores of \verbalizer were overall much lower than those of \manual, 
again showing the importance of linguistic resources when generating texts from ontologies. 
The high non-redundancy score of \verbalizer in the Disease Ontology is due to the fact that there are very few redundant message triples and no anonymous individuals or classes in the Disease Ontology. Hence, \verbalizer, which treats all the message triples as important and does not anonymize any individuals or classes performs well in terms of non-redundancy. We made a similar observation in 
Section~\ref{evaluatingNLNtexts}.

\section{Related work} \label{RelatedWork}

Simple ontology verbalizers \cite{Cregan2007,Kaljurand2007,Schwitter2008,HalaschekWiener2008,Schutte2009,Power2010b,Power2010,Schwitter2010,Liang2011b} typically produce texts describing individuals and classes without requiring manually authored domain-dependent linguistic resources. They usually tokenize the \owl identifiers or labels (e.g., \code{rdfs:label}) of the individuals or classes to obtain \nl names and sentence plans. Androutsopoulos \etalt \citeyear{Androutsopoulos2013} showed that the texts of the \swat verbalizer \cite{Stevens2011,Williams2011}, one of the best publicly available verbalizers, are perceived as being of significantly lower quality compared to texts generated by \nlowl with domain-dependent linguistic resources; \nl names, sentence plans, 
and (to a lesser extent) text plans were found to contribute most to this difference. Without domain-dependent linguistic resources, \nlowl was found to generate texts of the same quality as the \swat verbalizer. 

\nlowl is based on ideas from \textsc{ilex} \cite{ODonnell2001} and \textsc{m-piro} \cite{Isard2003,Androutsopoulos2007}. Excluding simple verbalizers, it 
is the only publicly available \nlg system for \owl, which is why we based our work on it. Nevertheless, its processing stages and linguistic resources are typical of \nlg systems \cite{ReiterDale2000,Mellish2006c}. Hence, we believe that our work is also applicable, at least in principle, to other \nlg systems. For example, \ontosum \cite{Bontcheva2005}, which generates natural language descriptions of individuals, but apparently not classes, from \rdfs and \owl ontologies, uses similar processing stages, and linguistic resources  corresponding to \nl names and sentence plans.
Reiter et al.\ \citeyear{Reiter2003} discuss the different types of knowledge that \nlg systems require and the difficulties of obtaining them (e.g., by interviewing experts or analyzing corpora). Unlike Reiter et al., we assume that domain knowledge is already available, in the form of \owl ontologies. The domain-specific linguistic resources of \nlowl belong in the `domain communication knowledge' of Reiter et al., who do not describe 
particular corpus-based algorithms to acquire knowledge.  

Ngonga Ngomo \etalt \citeyear{NgongaNgomo2013} discuss \textsc{sparql2nl}, a system that translates \textsc{sparql} queries 
to English. \textsc{sparql2nl} uses techniques similar to those of simple ontology verbalizers. To express the \rdf triples $\left<S, R, O\right>$ that are involved in a \textsc{sparql} query, it assumes that the labels (e.g., \code{rdfs:label}, perhaps also identifiers) of the relations are verbs or nouns.
It determines if a relation label is a verb or noun using hand-crafted rules and the \pos tags of the label's synonyms in WordNet \cite{Fellbaum1998}. It then employs manually authored templates, corresponding to our sentence plans, to express the relation; e.g., the template ``$S$ writes $O$'' is used for a triple involving the relation \code{:write}, since ``write'' is a verb, but ``$S$'s author is $O$'' is used for the relation \code{:author}, since ``author'' is a noun. To express the $S$ or $O$ of a triple, \textsc{sparql2nl} tokenizes the label (or identifier) of the corresponding individual or class, pluralizing the resulting name if it refers to a class.

Ratnaparkhi \citeyear{Ratnaparkhi2000} aims to express a set of attribute-value pairs as a natural language phrase; 
e.g., $\{\textit{city-from} = \text{Athens}, \textit{city-to} = \text{New York}, \textit{depart-day} = \text{Wednesday}\}$ 
becomes ``flights from Athens to New York on Wednesday''. 
A parallel training corpus containing sets of attribute-value pairs, the corresponding phrases, and their dependency trees is required. A maximum entropy model is trained on the corpus, roughly speaking to be able to estimate the probability of a dependency tree given a set of attribute-value pairs. Then, given an unseen set of attribute-value pairs, multiple alternative dependency trees are constructed in a top-down manner, using beam search and the maximum entropy model to estimate the probabilities of the trees being constructed. The most probable tree that expresses all the attribute-value pairs is eventually chosen, and the corresponding phrase is returned. In later work \cite{Ratnaparkhi2002}, the generated dependency trees are further altered by a set of hand-crafted rules that add unmentioned attributes, and the trees are also ranked by language models. In our case, where we aim to express multiple message triples $\left<S, R_i, O_i\right>$ all describing an individual or class $S$, we can think of the message triples as attribute-value pairs $R_i = O_i$. To apply the methods of Ratnaparkhi, however, a parallel training corpus with sets of attribute-value pairs (or message triples) and the corresponding target texts would be needed; and corpora of this kind are difficult to obtain. By contrast, our methods require no parallel corpus and, hence, can be more easily applied to ontologies of new domains. Furthermore, the methods of Ratnaparkhi aim to produce a single sentence per set of attribute-value pairs, whereas 
we produce linguistic resources
that are used to generate multi-sentence texts (e.g., our \nl names and sentence plans include annotations used in sentence aggregation and referring expression generation). 

Angeli \etalt \citeyear{Angeli2010} generate multi-sentence texts describing database records. Their methods also require a parallel training corpus consisting of manually authored texts and the database records (and particular record fields) expressed by each text. The generative model of Liang \etalt \citeyear{Liang2009} is applied to the training corpus to align the words of each text to the database records and fields it expresses. Templates are then extracted from the aligned texts, by replacing words aligned to record fields with variables. To generate a new text from a set of database records, the system generates a sequence of phrases. For each phrase, it first decides which records and fields to express, then which templates to generate the phrase with, and finally which template variables to replace by which record fields. These decisions are made either greedily or by sampling probability distributions learnt during
training. This process is repeated until all the given record fields have been expressed. A language model is also employed to ensure that the transitions between phrases sound natural. As with the work of Ratnaparkhi, the methods of Angeli \etalt could in principle be applied to express message triples describing an individual or class, 
but again a parallel training corpus containing texts and the database records and fields expressed by each text would be needed. 

Wong and Mooney \citeyear{Wong2006,wong2007} employ Statistical Machine Translation (\textsc{smt}) methods to automatically obtain  formal semantic representations from natural language sentences. They automatically construct a synchronous context-free grammar, by applying a statistical word alignment model to a parallel training corpus of sentences and their semantic representations. The grammar generates both natural language sentences and their semantic representations. Given a new sentence, the grammar produces candidate semantic representations, and a maximum-entropy model 
estimates the probability of each canidate representation. Chen and Mooney \citeyear{Chen2008} use the same methods in the reverse direction, to convert formal semantic  representations to single sentences. In principle, similar \textsc{smt} methods could be employed to generate sentences from  message triples.
However, a parallel corpus of texts and message triples would again be needed. Furthermore, \textsc{smt} methods produce a single sentence at a time, whereas our work concerns 
multi-sentence texts.

Lu \etalt \citeyear{Lu2009} generate natural language sentences from tree-structured semantic representations \cite{Lu2008}. Given a parallel training corpus of sentences and 
tree-structured semantic representations, hybrid trees are created by expanding the original semantic representation trees of the corpus with nodes standing for phrases of the corresponding sentences. To generate a new sentence from a tree-structured semantic representation, a set of candidate hybrid trees is initially produced based on predefined tree patterns and a 
\textsc{crf} model trained on the hybrid trees of the parallel corpus. A sentence is then obtained from the most probable candidate hybrid tree. In later work, Lu and Ng \citeyear{Lu2011} extend their hybrid trees to support formal logic (typed lambda calculus) semantic representations. A synchronous context free grammar is obtained from the extended hybrid trees of the parallel corpus. The grammar is then used to map formal logic expressions to new sentences. We note that \owl is based on description logic \cite{Baader2002} and, hence, methods similar to those of Lu \etalt could in principle be used to map \owl statements to sentences, though the hybrid trees would have to be modified for description logic. A parallel training corpus of texts and description logic expressions (or corresponding \owl statements) would again be needed, however, and only single sentences would be obtained. 

Konstas and Lapata \citeyear{Konstas2012b} use a probabilistic context-free grammar to convert a given set of database entries to a single sentence (or phrase). Roughly speaking, in each parse tree of the grammar, the leaves are the words of a sentence, and the internal nodes indicate which database entries are expressed by each subtree. 
The grammar is constructed using hand-crafted templates of rewrite rules and a parallel training corpus of database entries and 
sentences; a generative model based on the work of Liang \etalt \citeyear{Liang2009} is employed to estimate the probabilities of the grammar.
Subsequently, all the parse trees of the grammar for the sentences of the training corpus and the corresponding database entries are represented as a weighted directed hypergraph 
\cite{Klein2004}. The hypergraph's weights are estimated using the inside-outside algorithm \cite{Li2009} on the training corpus. 
Following 
Huang and Chiang \citeyear{Huang2007}, the hypergraph nodes are then integrated with an $n$-gram language model trained on the 
sentences of the corpus. Given a new set of database entries, the most probable derivation is found in the hypergraph using a $k$-best Viterbi search with cube pruning \cite{Chiang2007} and the final sentence is obtained from 
the derivation. In later work, Konstas and Lapata \citeyear{Konstas2012} find the most probable derivation in the hypergraph by forest reranking, using 
features that include the decoding probability of the derivation according to their previous work, the frequency of rewrite rules in the derivation, as well as lexical 
(e.g., word $n$-grams) and structural features 
(e.g., $n$-grams of record fields). The weights of the features are estimated with a structured perceptron \cite{Collins2002} on the training corpus.

Apart from simple verbalizers,
all the other related methods discussed above require a parallel training corpus of texts (or sentences, or phrases) and their semantic representations,
unlike our work. 
A further difference from our work is that all the previous methods assume that the English names of the various entities (or individuals and classes) are already available in the semantic representations of the texts to be generated, or that they can be directly obtained from the identifiers
of the entities in the semantic representations. By contrast, we 
also proposed methods to produce appropriate \nl names for individuals and classes, and we 
showed experimentally (Section~\ref{evaluatingNLNtexts}) that without \nl names the perceived quality of the generated texts is significantly lower. 

Our sentence plan generation method contains a template extraction stage (Section~\ref{PatternExtraction}), which is similar to methods 
proposed to automatically obtain templates that extract instances of particular relations from texts. We discussed bootstrapping 
in Section~\ref{bootstrapping}. Xu \etalt \citeyear{Xu2007} adopt a similar 
bootstrapping approach with templates obtained from dependency trees. 
Bootstrappoing has also been used to obtain paraphrasing 
and textual entailment rules
\cite{Szpektor2004,Androutsopoulos2010}.
The sentence plans we produce are not just templates (e.g., ``$X$ bought $Y$''), but include additional annotations (e.g., \pos tags, agreement, voice, tense, cases). Furthermore, they are not intended to capture \emph{all} the alternative natural language expressions that convey a particular relation, unlike information extraction, paraphrase and textual entailment recognition; our goal is to obtain a single sentence plan per relation that leads to high quality texts. 

Bootstrapping approaches have also been used to obtain templates that extract named entities of a particular semantic class (e.g., person names) from texts \cite{Riloff1996,Patwardhan2006}. Methods of this kind aim to extract \emph{all} the named entities of a particular class from a corpus. By contast, we aim to assign a single high quality \nl name to each individual or class of a given ontology. Furthermore, our \nl names are not simply strings, but contain additional information (e.g., head, gender, number, agreement) that helps produce high quality texts.

\section{Conclusions and future work} \label{Conclusion}

Concept-to-text generation systems typically require domain-specific linguistic resources to produce high quality texts, but manually constructing these resources can be tedious and costly. Focusing on \nlowl, a publicly available state of the art natural language generator for \owl ontologies, we proposed methods to automatically or semi-automatically extract from the Web sentence plans and natural language names, two of the most important types of domain-specific 
generation resources.\footnote{The software of this article has been embedded in \nlowl and will be available from \url{http://nlp.cs.aueb.gr/software.html}.} We showed experimentally that texts generated using linguistic resources produced by our methods in a semi-automatic manner, with minimal human involvement, are perceived as being almost as good as texts generated using manually authored linguistic resources, and much better than texts produced by using linguistic resources extracted from the relation and entity identifiers of the 
ontologies. 
Using our methods, constructing sentence plans and natural language names requires human effort of a few minutes or hours, respectively, per ontology, whereas constructing them manually from scratch is typically a matter of days. Also, our methods do not require any familiarity with the internals of \nlowl and the details of its linguistic resources. Furthermore, unlike previous related work, no parallel corpus of sentences and semantic representations is required. On the downside, our methods do not perform sufficiently well in a fully-automatic scenario, with no human involvement during the construction of the linguistic resources. 

The processing stages and linguistic resources of \nlowl are typical of \nlg systems. Hence, we believe that our work is also applicable, at least in principle, to other \nlg systems. Our methods may also be useful in simpler ontology verbalizers, where the main concern seems to be to avoid manually authoring domain-specific linguistic resources, currently at the expense of producing texts of much lower quality. Future work could aim to improve our methods to allow using them in a fully automatic manner. Further work could 
also explore how other kinds of domain-specific linguistic resources for \nlg, most importantly text plans, could be constructed automatically or semi-automatically. Another future goal might be to consider languages other than English.

\vskip 0.2in
{\small
\bibliography{Bibliography}

\begin{thebibliography}{}

\bibitem[\protect\BCAY{Androutsopoulos, Lampouras,\ \BBA\
  Galanis}{Androutsopoulos et~al.}{2013}]{Androutsopoulos2013}
Androutsopoulos, I., Lampouras, G., \BBA\ Galanis, D. \BBOP2013\BBCP.
\newblock \BBOQ Generating natural language descriptions from {OWL} ontologies:
  the {NaturalOWL} system\BBCQ\
\newblock {\Bem Journal of Artificial Intelligence Research}, {\Bem 48\/}(1),
  671--715.

\bibitem[\protect\BCAY{Androutsopoulos\ \BBA\ Malakasiotis}{Androutsopoulos\
  \BBA\ Malakasiotis}{2010}]{Androutsopoulos2010}
Androutsopoulos, I.\BBACOMMA\  \BBA\ Malakasiotis, P. \BBOP2010\BBCP.
\newblock \BBOQ A survey of paraphrasing and textual entailment methods\BBCQ\
\newblock {\Bem Journal of Artificial Intelligence Research}, {\Bem 38\/}(1),
  135--187.

\bibitem[\protect\BCAY{Androutsopoulos, Oberlander,\ \BBA\
  Karkaletsis}{Androutsopoulos et~al.}{2007}]{Androutsopoulos2007}
Androutsopoulos, I., Oberlander, J., \BBA\ Karkaletsis, V. \BBOP2007\BBCP.
\newblock \BBOQ Source authoring for multilingual generation of personalised
  object descriptions\BBCQ\
\newblock {\Bem Natural Language Engineering}, {\Bem 13\/}(3), 191--233.

\bibitem[\protect\BCAY{Angeli, Liang,\ \BBA\ Klein}{Angeli
  et~al.}{2010}]{Angeli2010}
Angeli, G., Liang, P., \BBA\ Klein, D. \BBOP2010\BBCP.
\newblock \BBOQ A simple domain-independent probabilistic approach to
  generation\BBCQ\
\newblock In {\Bem Conference on Empirical Methods in Natural Language
  Processing}, \BPGS\ 502--512, Cambridge, MA.

\bibitem[\protect\BCAY{Antoniou\ \BBA\ van Harmelen}{Antoniou\ \BBA\ van
  Harmelen}{2008}]{Antoniou2008}
Antoniou, G.\BBACOMMA\  \BBA\ van Harmelen, F. \BBOP2008\BBCP.
\newblock {\Bem A {S}emantic {W}eb primer\/} (2nd \BEd).
\newblock MIT Press.

\bibitem[\protect\BCAY{Baader, Calvanese, McGuinness, Nardi,\ \BBA\
  Patel-Schneider}{Baader et~al.}{2002}]{Baader2002}
Baader, F., Calvanese, D., McGuinness, D., Nardi, D., \BBA\ Patel-Schneider,
  P.\BEDS. \BBOP2002\BBCP.
\newblock {\Bem The Description Logic Handbook}.
\newblock Cambridge University Press.

\bibitem[\protect\BCAY{Berners-Lee, Hendler,\ \BBA\ Lassila}{Berners-Lee
  et~al.}{2001}]{BernersLee2001}
Berners-Lee, T., Hendler, J., \BBA\ Lassila, O. \BBOP2001\BBCP.
\newblock \BBOQ The {S}emantic {W}eb\BBCQ\
\newblock {\Bem Scientific American}, {\Bem May\/}(1), 34--43.

\bibitem[\protect\BCAY{Bontcheva}{Bontcheva}{2005}]{Bontcheva2005}
Bontcheva, K. \BBOP2005\BBCP.
\newblock \BBOQ Generating tailored textual summaries from ontologies\BBCQ\
\newblock In {\Bem 2nd European Semantic Web Conference}, \BPGS\ 531--545,
  Heraklion, Greece.

\bibitem[\protect\BCAY{Chen\ \BBA\ Mooney}{Chen\ \BBA\ Mooney}{2008}]{Chen2008}
Chen, D.~L.\BBACOMMA\  \BBA\ Mooney, R.~J. \BBOP2008\BBCP.
\newblock \BBOQ Learning to sportscast: a test of grounded language
  acquisition\BBCQ\
\newblock In {\Bem 25th International Conference on Machine Learning}, \BPGS\
  128--135, Helsinki, Finland.

\bibitem[\protect\BCAY{Chiang}{Chiang}{2007}]{Chiang2007}
Chiang, D. \BBOP2007\BBCP.
\newblock \BBOQ Hierarchical phrase-based translation\BBCQ\
\newblock {\Bem Computational Linguistics}, {\Bem 33\/}(2), 201--228.

\bibitem[\protect\BCAY{Collins}{Collins}{2002}]{Collins2002}
Collins, M. \BBOP2002\BBCP.
\newblock \BBOQ Discriminative training methods for {H}idden {M}arkov {M}odels:
  Theory and experiments with {P}erceptron algorithms\BBCQ\
\newblock In {\Bem Conference on Empirical Methods in Natural Language
  Processing}, \BPGS\ 1--8, Stroudsburg, PA, USA.

\bibitem[\protect\BCAY{Cregan, Schwitter,\ \BBA\ Meyer}{Cregan
  et~al.}{2007}]{Cregan2007}
Cregan, A., Schwitter, R., \BBA\ Meyer, T. \BBOP2007\BBCP.
\newblock \BBOQ Sydney {OWL} syntax -- towards a controlled natural language
  syntax for {OWL}\BBCQ\
\newblock In {\Bem {OWL:} Experiences and Directions Workshop}, Innsbruck,
  Austria.

\bibitem[\protect\BCAY{Fellbaum}{Fellbaum}{1998}]{Fellbaum1998}
Fellbaum, C.\BED. \BBOP1998\BBCP.
\newblock {\Bem {WordNet}: an electronic lexical database}.
\newblock MIT Press.

\bibitem[\protect\BCAY{Galanis\ \BBA\ Androutsopoulos}{Galanis\ \BBA\
  Androutsopoulos}{2007}]{Galanis2007}
Galanis, D.\BBACOMMA\  \BBA\ Androutsopoulos, I. \BBOP2007\BBCP.
\newblock \BBOQ Generating multilingual descriptions from linguistically
  annotated {OWL} ontologies: the {NaturalOWL} system\BBCQ\
\newblock In {\Bem 11th European Workshop on Natural Language Generation},
  \BPGS\ 143--146, Schloss Dagstuhl, Germany.

\bibitem[\protect\BCAY{Galanis, Karakatsiotis, Lampouras,\ \BBA\
  Androutsopoulos}{Galanis et~al.}{2009}]{Galanis2009}
Galanis, D., Karakatsiotis, G., Lampouras, G., \BBA\ Androutsopoulos, I.
  \BBOP2009\BBCP.
\newblock \BBOQ An open-source natural language generator for {OWL} ontologies
  and its use in \protege and {S}econd {L}ife\BBCQ\
\newblock In {\Bem 12th Conference of the European Chapter of {ACL} (demos)},
  \BPGS\ 17--20, Athens, Greece.

\bibitem[\protect\BCAY{Gatt\ \BBA\ Reiter}{Gatt\ \BBA\ Reiter}{2009}]{Gatt2009}
Gatt, A.\BBACOMMA\  \BBA\ Reiter, E. \BBOP2009\BBCP.
\newblock \BBOQ {SimpleNLG}: A realisation engine for practical
  applications\BBCQ\
\newblock In {\Bem 12th European Workshop on Natural Language Generation},
  \BPGS\ 90--93, Athens, Greece.

\bibitem[\protect\BCAY{Grau, Horrocks, Motik, Parsia, Patel-Schneider,\ \BBA\
  Sattler}{Grau et~al.}{2008}]{Grau2008}
Grau, B., Horrocks, I., Motik, B., Parsia, B., Patel-Schneider, P., \BBA\
  Sattler, U. \BBOP2008\BBCP.
\newblock \BBOQ {OWL 2}: The next step for {OWL}\BBCQ\
\newblock {\Bem Web Semantics}, {\Bem 6\/}(4), 309--322.

\bibitem[\protect\BCAY{Gupta\ \BBA\ Manning}{Gupta\ \BBA\
  Manning}{2014}]{gupta2014}
Gupta, S.\BBACOMMA\  \BBA\ Manning, C.~D. \BBOP2014\BBCP.
\newblock \BBOQ Improved pattern learning for bootstrapped entity
  extraction\BBCQ\
\newblock In {\Bem 18th Conference on Computational Natural Language Learning},
  \BPGS\ 98--108, Ann Arbor, Michigan.

\bibitem[\protect\BCAY{Halaschek-Wiener, Golbeck, Parsia, Kolovski,\ \BBA\
  Hendler}{Halaschek-Wiener et~al.}{2008}]{HalaschekWiener2008}
Halaschek-Wiener, C., Golbeck, J., Parsia, B., Kolovski, V., \BBA\ Hendler, J.
  \BBOP2008\BBCP.
\newblock \BBOQ Image browsing and natural language paraphrases of semantic web
  annotations\BBCQ\
\newblock In {\Bem 1st International Workshop on Semantic Web Annotations for
  Multimedia}, Tenerife, Spain.

\bibitem[\protect\BCAY{Horrocks, Patel-Schneider,\ \BBA\ van Harmelen}{Horrocks
  et~al.}{2003}]{Horrocks2003}
Horrocks, I., Patel-Schneider, P., \BBA\ van Harmelen, F. \BBOP2003\BBCP.
\newblock \BBOQ From {SHIQ} and {RDF} to {OWL}: the making of a {W}eb
  {O}ntology {L}anguage\BBCQ\
\newblock {\Bem Web Semantics}, {\Bem 1\/}(1), 7--26.

\bibitem[\protect\BCAY{Huang\ \BBA\ Chiang}{Huang\ \BBA\
  Chiang}{2007}]{Huang2007}
Huang, L.\BBACOMMA\  \BBA\ Chiang, D. \BBOP2007\BBCP.
\newblock \BBOQ Forest rescoring: Faster decoding with integrated language
  models\BBCQ\
\newblock In {\Bem 45th Annual Meeting of {ACL}}, \BPGS\ 144--151, Prague,
  Czech Republic.

\bibitem[\protect\BCAY{Isard, Oberlander, Androutsopoulos,\ \BBA\
  Matheson}{Isard et~al.}{2003}]{Isard2003}
Isard, A., Oberlander, J., Androutsopoulos, I., \BBA\ Matheson, C.
  \BBOP2003\BBCP.
\newblock \BBOQ Speaking the users' languages\BBCQ\
\newblock {\Bem IEEE Intelligent Systems}, {\Bem 18\/}(1), 40--45.

\bibitem[\protect\BCAY{Jurafsky\ \BBA\ Martin}{Jurafsky\ \BBA\
  Martin}{2008}]{JurafskyMartin2008}
Jurafsky, D.\BBACOMMA\  \BBA\ Martin, J. \BBOP2008\BBCP.
\newblock {\Bem Speech and {L}anguage {P}rocessing}.
\newblock Prentice Hall.

\bibitem[\protect\BCAY{Kaljurand\ \BBA\ Fuchs}{Kaljurand\ \BBA\
  Fuchs}{2007}]{Kaljurand2007}
Kaljurand, K.\BBACOMMA\  \BBA\ Fuchs, N. \BBOP2007\BBCP.
\newblock \BBOQ Verbalizing {OWL} in {A}ttempto {C}ontrolled {E}nglish\BBCQ\
\newblock In {\Bem 3rd International Workshop on {OWL}: Experiences and
  Directions}, Innsbruck, Austria.

\bibitem[\protect\BCAY{Klein\ \BBA\ Manning}{Klein\ \BBA\
  Manning}{2002}]{Klein2004}
Klein, D.\BBACOMMA\  \BBA\ Manning, C. \BBOP2002\BBCP.
\newblock \BBOQ Parsing and hypergraphs\BBCQ\
\newblock In Bunt, H., Carroll, J., \BBA\ Satta, G.\BEDS, {\Bem New
  Developments in Parsing Technology}, \BPGS\ 351--372. Kluwer Academic
  Publishers.

\bibitem[\protect\BCAY{Konstas\ \BBA\ Lapata}{Konstas\ \BBA\
  Lapata}{2012a}]{Konstas2012}
Konstas, I.\BBACOMMA\  \BBA\ Lapata, M. \BBOP2012a\BBCP.
\newblock \BBOQ Concept-to-text generation via discriminative reranking\BBCQ\
\newblock In {\Bem 50th Annual Meeting of {ACL}}, \BPGS\ 369--378, Jeju Island,
  Korea.

\bibitem[\protect\BCAY{Konstas\ \BBA\ Lapata}{Konstas\ \BBA\
  Lapata}{2012b}]{Konstas2012b}
Konstas, I.\BBACOMMA\  \BBA\ Lapata, M. \BBOP2012b\BBCP.
\newblock \BBOQ Unsupervised concept-to-text generation with hypergraphs\BBCQ\
\newblock In {\Bem Conference on Human Language Technology of Annual Conference
  of the North American Chapter of {ACL}}, \BPGS\ 752--761, Montr\'{e}al,
  Canada.

\bibitem[\protect\BCAY{Li\ \BBA\ Eisner}{Li\ \BBA\ Eisner}{2009}]{Li2009}
Li, Z.\BBACOMMA\  \BBA\ Eisner, J. \BBOP2009\BBCP.
\newblock \BBOQ First- and second-order expectation semirings with applications
  to minimum-risk training on translation forests\BBCQ\
\newblock In {\Bem Conference on Empirical Methods in Natural Language
  Processing}, \BPGS\ 40--51, Singapore.

\bibitem[\protect\BCAY{Liang, Jordan,\ \BBA\ Klein}{Liang
  et~al.}{2009}]{Liang2009}
Liang, P., Jordan, M., \BBA\ Klein, D. \BBOP2009\BBCP.
\newblock \BBOQ Learning semantic correspondences with less supervision\BBCQ\
\newblock In {\Bem 47th Meeting of {ACL} and 4th International Joint Conference
  on Natural Language Processing of the Asian Federation of Natural Language
  Processing}, \BPGS\ 91--99, Suntec, Singapore.

\bibitem[\protect\BCAY{Liang, Stevens, Scott,\ \BBA\ Rector}{Liang
  et~al.}{2011}]{Liang2011b}
Liang, S., Stevens, R., Scott, D., \BBA\ Rector, A. \BBOP2011\BBCP.
\newblock \BBOQ Automatic verbalisation of {SNOMED} classes using
  {OntoVerbal}\BBCQ\
\newblock In {\Bem 13th Conference on Artificial Intelligence in Medicine},
  \BPGS\ 338--342, Bled, Slovenia.

\bibitem[\protect\BCAY{Lu\ \BBA\ Ng}{Lu\ \BBA\ Ng}{2011}]{Lu2011}
Lu, W.\BBACOMMA\  \BBA\ Ng, H.~T. \BBOP2011\BBCP.
\newblock \BBOQ A probabilistic forest-to-string model for language generation
  from typed lambda calculus expressions\BBCQ\
\newblock In {\Bem Conference on Empirical Methods in Natural Language
  Processing}, \BPGS\ 1611--1622, Edinburgh, {UK}.

\bibitem[\protect\BCAY{Lu, Ng,\ \BBA\ Lee}{Lu et~al.}{2009}]{Lu2009}
Lu, W., Ng, H.~T., \BBA\ Lee, W.~S. \BBOP2009\BBCP.
\newblock \BBOQ Natural language generation with tree conditional random
  fields\BBCQ\
\newblock In {\Bem Conference on Empirical Methods in Natural Language
  Processing}, \BPGS\ 400--409, Singapore.

\bibitem[\protect\BCAY{Lu, Ng, Lee,\ \BBA\ Zettlemoyer}{Lu
  et~al.}{2008}]{Lu2008}
Lu, W., Ng, H.~T., Lee, W.~S., \BBA\ Zettlemoyer, L.~S. \BBOP2008\BBCP.
\newblock \BBOQ A generative model for parsing natural language to meaning
  representations\BBCQ\
\newblock In {\Bem Conference on Empirical Methods in Natural Language
  Processing}, \BPGS\ 783--792, Honolulu, Hawaii.

\bibitem[\protect\BCAY{Manning\ \BBA\ Schutze}{Manning\ \BBA\
  Schutze}{2000}]{Manning2000}
Manning, C.\BBACOMMA\  \BBA\ Schutze, H. \BBOP2000\BBCP.
\newblock {\Bem Foundations of Statistical Natural Language Processing}.
\newblock {MIT} Press.

\bibitem[\protect\BCAY{Mellish\ \BBA\ Pan}{Mellish\ \BBA\
  Pan}{2008}]{Mellish2008}
Mellish, C.\BBACOMMA\  \BBA\ Pan, J. \BBOP2008\BBCP.
\newblock \BBOQ Natural language directed inference from ontologies\BBCQ\
\newblock {\Bem Artificial Intelligence}, {\Bem 172\/}(10), 1285--1315.

\bibitem[\protect\BCAY{Mellish, Scott, Cahill, Paiva, Evans,\ \BBA\
  Reape}{Mellish et~al.}{2006}]{Mellish2006c}
Mellish, C., Scott, D., Cahill, L., Paiva, D., Evans, R., \BBA\ Reape, M.
  \BBOP2006\BBCP.
\newblock \BBOQ A reference architecture for natural language generation
  systems\BBCQ\
\newblock {\Bem Natural Language Engineering}, {\Bem 12\/}(1), 1--34.

\bibitem[\protect\BCAY{Mellish\ \BBA\ Sun}{Mellish\ \BBA\
  Sun}{2006}]{Mellish2006}
Mellish, C.\BBACOMMA\  \BBA\ Sun, X. \BBOP2006\BBCP.
\newblock \BBOQ The {S}emantic {W}eb as a linguistic resource: opportunities
  for natural language generation\BBCQ\
\newblock {\Bem Knowledge Based Systems}, {\Bem 19\/}(5), 298--303.

\bibitem[\protect\BCAY{Muslea}{Muslea}{1999}]{Muslea1999}
Muslea, I. \BBOP1999\BBCP.
\newblock \BBOQ Extraction patterns for information extraction tasks\BBCQ\
\newblock In {\Bem AAAI Workshop on Machine Learning for Information
  Extraction}, \BPGS\ 1--6, Orlando, Florida.

\bibitem[\protect\BCAY{Ngonga~Ngomo, B\"{u}hmann, Unger, Lehmann,\ \BBA\
  Gerber}{Ngonga~Ngomo et~al.}{2013}]{NgongaNgomo2013}
Ngonga~Ngomo, A.-C., B\"{u}hmann, L., Unger, C., Lehmann, J., \BBA\ Gerber, D.
  \BBOP2013\BBCP.
\newblock \BBOQ Sorry, {I} don't speak {SPARQL}: Translating {SPARQL} queries
  into natural language\BBCQ\
\newblock In {\Bem 22nd International Conference on World Wide Web}, \BPGS\
  977--988, Rio de Janeiro, Brazil.

\bibitem[\protect\BCAY{O'Donnell, Mellish, Oberlander,\ \BBA\ Knott}{O'Donnell
  et~al.}{2001}]{ODonnell2001}
O'Donnell, M., Mellish, C., Oberlander, J., \BBA\ Knott, A. \BBOP2001\BBCP.
\newblock \BBOQ {ILEX}: an architecture for a dynamic hypertext generation
  system\BBCQ\
\newblock {\Bem Natural Language Engineering}, {\Bem 7\/}(3), 225--250.

\bibitem[\protect\BCAY{Patwardhan\ \BBA\ Riloff}{Patwardhan\ \BBA\
  Riloff}{2006}]{Patwardhan2006}
Patwardhan, S.\BBACOMMA\  \BBA\ Riloff, E. \BBOP2006\BBCP.
\newblock \BBOQ Learning domain-specific information extraction patterns from
  the {W}eb\BBCQ\
\newblock In {\Bem {ACL} Workshop on Information Extraction Beyond the
  Document}, \BPGS\ 66--73, Sydney, Australia.

\bibitem[\protect\BCAY{Power}{Power}{2010}]{Power2010}
Power, R. \BBOP2010\BBCP.
\newblock \BBOQ Complexity assumptions in ontology verbalisation\BBCQ\
\newblock In {\Bem 48th Annual Meeting of {ACL} (short papers)}, \BPGS\
  132--136, Uppsala, Sweden.

\bibitem[\protect\BCAY{Power\ \BBA\ Third}{Power\ \BBA\
  Third}{2010}]{Power2010b}
Power, R.\BBACOMMA\  \BBA\ Third, A. \BBOP2010\BBCP.
\newblock \BBOQ Expressing {OWL} axioms by {E}nglish sentences: Dubious in
  theory, feasible in practice\BBCQ\
\newblock In {\Bem 23rd International Conf.\ on Computional Linguistics},
  \BPGS\ 1006--1013, Beijing, China.

\bibitem[\protect\BCAY{Ratnaparkhi}{Ratnaparkhi}{2000}]{Ratnaparkhi2000}
Ratnaparkhi, A. \BBOP2000\BBCP.
\newblock \BBOQ Trainable methods for surface natural language generation\BBCQ\
\newblock In {\Bem 1st Annual Conference of the North American Chapter of
  {ACL}}, \BPGS\ 194--201, Seattle, WA.

\bibitem[\protect\BCAY{Ratnaparkhi}{Ratnaparkhi}{2002}]{Ratnaparkhi2002}
Ratnaparkhi, A. \BBOP2002\BBCP.
\newblock \BBOQ Trainable approaches to surface natural language generation and
  their application to conversational dialog systems\BBCQ\
\newblock {\Bem Computer Speech {\&} Language}, {\Bem 16\/}(3-4), 435--455.

\bibitem[\protect\BCAY{Rector, Drummond, Horridge, Rogers, Knublauch, Stevens,
  Wang,\ \BBA\ Wroe}{Rector et~al.}{2004}]{Rector2004}
Rector, A., Drummond, N., Horridge, M., Rogers, J., Knublauch, H., Stevens, R.,
  Wang, H., \BBA\ Wroe, C. \BBOP2004\BBCP.
\newblock \BBOQ {OWL} pizzas: Practical experience of teaching {OWL-DL}: Common
  errors \& common patterns\BBCQ\
\newblock In {\Bem 14th Int.\ Conf.\ on Knowledge Eng.\ \& Knowledge
  Management}, \BPGS\ 63--81, Northamptonshire, UK.

\bibitem[\protect\BCAY{Reiter\ \BBA\ Dale}{Reiter\ \BBA\
  Dale}{2000}]{ReiterDale2000}
Reiter, E.\BBACOMMA\  \BBA\ Dale, R. \BBOP2000\BBCP.
\newblock {\Bem Building Natural Language Generation systems}.
\newblock Cambridge University Press.

\bibitem[\protect\BCAY{Reiter, Sripada,\ \BBA\ Robertson}{Reiter
  et~al.}{2003}]{Reiter2003}
Reiter, E., Sripada, S., \BBA\ Robertson, R. \BBOP2003\BBCP.
\newblock \BBOQ Acquiring correct knowledge for natural language
  generation\BBCQ\
\newblock {\Bem Journal of Artificial Intelligence Research}, {\Bem 18},
  491--516.

\bibitem[\protect\BCAY{Riloff}{Riloff}{1996}]{Riloff1996}
Riloff, E. \BBOP1996\BBCP.
\newblock \BBOQ Automatically generating extraction patterns from untagged
  text\BBCQ\
\newblock In {\Bem 13th National Conference on Artificial Intelligence - Volume
  2}, \BPGS\ 1044--1049, Portland, Oregon.

\bibitem[\protect\BCAY{Riloff\ \BBA\ Jones}{Riloff\ \BBA\
  Jones}{1999}]{Riloff1999}
Riloff, E.\BBACOMMA\  \BBA\ Jones, R. \BBOP1999\BBCP.
\newblock \BBOQ Learning dictionaries for information extraction by multi-level
  bootstrapping\BBCQ\
\newblock In {\Bem 16th National Conference on Artificial Intelligence and the
  11th Innovative Applications of Artificial Intelligence Conference}, \BPGS\
  474--479, Orlando, Florida, USA.

\bibitem[\protect\BCAY{Schutte}{Schutte}{2009}]{Schutte2009}
Schutte, N. \BBOP2009\BBCP.
\newblock \BBOQ Generating natural language descriptions of ontology
  concepts\BBCQ\
\newblock In {\Bem 12th European Workshop on Natural Language Generation},
  \BPGS\ 106--109, Athens, Greece.

\bibitem[\protect\BCAY{Schwitter}{Schwitter}{2010}]{Schwitter2010}
Schwitter, R. \BBOP2010\BBCP.
\newblock \BBOQ Controlled natural languages for knowledge representation\BBCQ\
\newblock In {\Bem 23rd International Conference on Computational Linguistics
  (posters)}, \BPGS\ 1113--1121, Beijing, China.

\bibitem[\protect\BCAY{Schwitter, Kaljurand, Cregan, Dolbear,\ \BBA\
  Hart}{Schwitter et~al.}{2008}]{Schwitter2008}
Schwitter, R., Kaljurand, K., Cregan, A., Dolbear, C., \BBA\ Hart, G.
  \BBOP2008\BBCP.
\newblock \BBOQ A comparison of three controlled natural languages for {OWL}
  1.1\BBCQ\
\newblock In {\Bem 4th {OWL}: Experiences and Directions Workshop}, Washington
  DC.

\bibitem[\protect\BCAY{Shadbolt, Berners-Lee,\ \BBA\ Hall}{Shadbolt
  et~al.}{2006}]{Shadbolt2006}
Shadbolt, N., Berners-Lee, T., \BBA\ Hall, W. \BBOP2006\BBCP.
\newblock \BBOQ The {S}emantic {W}eb revisited\BBCQ\
\newblock {\Bem {IEEE} Intelligent Systems}, {\Bem 21\/}(3), 96--101.

\bibitem[\protect\BCAY{Stevens, Malone, Williams, Power,\ \BBA\ Third}{Stevens
  et~al.}{2011}]{Stevens2011}
Stevens, R., Malone, J., Williams, S., Power, R., \BBA\ Third, A.
  \BBOP2011\BBCP.
\newblock \BBOQ Automatic generation of textual class definitions from {OWL} to
  {E}nglish\BBCQ\
\newblock {\Bem Biomedical Semantics}, {\Bem 2\/}(S 2:S5).

\bibitem[\protect\BCAY{Szpektor, Dagan,\ \BBA\ Coppola}{Szpektor
  et~al.}{2004}]{Szpektor2004}
Szpektor, I.and~Tanev, H., Dagan, I., \BBA\ Coppola, B. \BBOP2004\BBCP.
\newblock \BBOQ Scaling web-based acquisition of entailment relations\BBCQ\
\newblock In {\Bem Conf.\ on Empirical Methods in Natural Language Processing},
  \BPGS\ 41--48, Barcelona, Spain.

\bibitem[\protect\BCAY{Williams, Third,\ \BBA\ Power}{Williams
  et~al.}{2011}]{Williams2011}
Williams, S., Third, A., \BBA\ Power, R. \BBOP2011\BBCP.
\newblock \BBOQ Levels of organization in ontology verbalization\BBCQ\
\newblock In {\Bem 13th European Workshop on Natural Language Generation},
  \BPGS\ 158--163, Nancy, France.

\bibitem[\protect\BCAY{Wong\ \BBA\ Mooney}{Wong\ \BBA\ Mooney}{2006}]{Wong2006}
Wong, Y.~W.\BBACOMMA\  \BBA\ Mooney, R.~J. \BBOP2006\BBCP.
\newblock \BBOQ Learning for semantic parsing with statistical machine
  translation\BBCQ\
\newblock In {\Bem Conf.\ on Human Lang.\ Technology of the Annual Conf.\ of
  {ACL}}, \BPGS\ 439--446, New York, New York.

\bibitem[\protect\BCAY{Wong\ \BBA\ Mooney}{Wong\ \BBA\ Mooney}{2007}]{wong2007}
Wong, Y.~W.\BBACOMMA\  \BBA\ Mooney, R.~J. \BBOP2007\BBCP.
\newblock \BBOQ Learning synchronous grammars for semantic parsing with lambda
  calculus\BBCQ\
\newblock In {\Bem 45th Annual Meeting of {ACL}}, \BPGS\ 960--967, Prague,
  Czech Republic.

\bibitem[\protect\BCAY{Xu, Uszkoreit,\ \BBA\ Li}{Xu et~al.}{2007}]{Xu2007}
Xu, F., Uszkoreit, H., \BBA\ Li, H. \BBOP2007\BBCP.
\newblock \BBOQ A seed-driven bottom-up machine learning framework for
  extracting relations of various complexity\BBCQ\
\newblock In {\Bem 45th Annual Meeting of {ACL}}, \BPGS\ 584--591, Prague,
  Czech Republic.

\end{thebibliography}
\bibliographystyle{theapa}
}

\end{document}